%% file: main.tex
\documentclass[11pt]{article}

\usepackage[utf8]{inputenc}
\usepackage[T1]{fontenc}
\usepackage{authblk}
\usepackage[
a4paper,
top=2.2cm,
bottom=2.3cm,
left=2.3cm,
right=2.3cm,
headheight=15pt
]{geometry}

\usepackage{newtxtext}

\usepackage{amsmath}
\usepackage{amssymb}

\usepackage{newtxmath}

\usepackage{microtype}

\usepackage{setspace}
\usepackage{titlesec}

\titleformat{\section}
{\Large\bfseries}
{\thesection}
{0.75em}
{}

\titleformat{\subsection}
{\large\bfseries}
{\thesubsection}
{0.75em}
{}

\titleformat{\subsubsection}
{\normalsize\bfseries}
{\thesubsubsection}
{0.75em}
{}

\titlespacing*{\section}
{0pt}
{2.4ex plus 0.5ex minus 0.2ex}
{1.1ex}

\titlespacing*{\subsection}
{0pt}
{2.0ex plus 0.4ex minus 0.2ex}
{0.8ex}

\titlespacing*{\subsubsection}
{0pt}
{1.6ex plus 0.3ex minus 0.2ex}
{0.6ex}

\usepackage{float}
\usepackage{graphicx}
\usepackage{caption}
\usepackage{subcaption}

\usepackage{booktabs}
\usepackage{tabularx}
\usepackage{array}
\usepackage{multirow}
\usepackage{threeparttable}
\usepackage{rotating}

\renewcommand{\arraystretch}{1.15}
\newcolumntype{Y}{>{\centering\arraybackslash}X}
\newcolumntype{L}[1]{>{\raggedright\arraybackslash}p{#1}}
\newcolumntype{C}[1]{>{\centering\arraybackslash}p{#1}}
\newcolumntype{R}[1]{>{\raggedleft\arraybackslash}p{#1}}

\usepackage{enumitem}

\setlist{
	topsep=4pt,
	itemsep=2pt,
	parsep=0pt,
	leftmargin=1.6em
}

\usepackage[numbers,sort&compress]{natbib}

\usepackage{fancyhdr}

\usepackage{hyperref}

\title{A$^3$PreCore: Robust Rigid Registration for Native FLAIR-Space Brain-Mask Transfer}

\author[1]{Elmira Mirzabeigi\thanks{Corresponding author: \href{mailto:elmira.mirzabeigi@umontreal.ca}{elmira.mirzabeigi@umontreal.ca}}}

\author[1]{Zahra Khalilzad}

\author[1]{Taylor Snowden-Richardson}

\author[1]{Jesus Eduardo Lugo Arce}

\author[3]{Jodie R. Gawryluk}

\author[2]{Brian R. Christie}

\author[1]{Jocelyn Faubert\thanks{Corresponding author: \href{mailto:jocelyn.faubert@umontreal.ca}{jocelyn.faubert@umontreal.ca}}}

\affil[1]{%
	Faubert Laboratory, School of Optometry,
	Université de Montréal,
	3744 Jean-Brillant,
	Montréal, Québec H3T 1P1, Canada
}

\affil[2]{%
	Division of Medical Sciences,
	University of Victoria,
	3800 Finnerty Road,
	Victoria, British Columbia V8P 5C2, Canada
}

\affil[3]{%
	Department of Psychology,
	University of Victoria,
	3800 Finnerty Road,
	Victoria, British Columbia V8P 5C2, Canada
}

\date{}

\begin{document}
	
	\maketitle
	
	% Reduce the blank space beneath the title
	\vspace{-2em}
	
	% ============================================================
	% ABSTRACT
	% ============================================================
	
	\begin{abstract}
		Reliable brain-mask transfer into native fluid-attenuated inversion recovery
		(FLAIR) space can fail when T$_1$-to-FLAIR rigid registration is unstable.
		We developed A\textsuperscript{3}PreCore, an optimization-based rigid-registration strategy that estimates a subject-specific six-degree-of-freedom rigid transformation between the full bias-corrected T$_1$-weighted image and native FLAIR using normalized mutual information, jointly optimizing rotational and translational parameters while preserving the established T$_1$-space anatomical mask.
		We also introduced extended quantitative quality control (E-QC), a
		multidimensional framework that retains method-specific descriptors of mask
		extent, field-of-view-relative position, FLAIR intensity support, boundary
		support, and inferior morphology together with pairwise descriptors of
		volumetric, centroid, and principal-axis disagreement, without collapsing
		them into a composite score. In 78 participants with paired T$_1$-weighted
		and FLAIR imaging, A\textsuperscript{3}PreCore was compared with
		\texttt{fsl\_anat} and the Brain Extraction Tool (BET), which generated masks
		directly in native FLAIR space. Gross rigid-body misregistration occurred in
		21/78 \texttt{fsl\_anat} outputs (26.9\%) and 0/78
		A\textsuperscript{3}PreCore outputs (exact two-sided McNemar
		$p=9.54\times10^{-7}$). Among 57 participants without gross
		\texttt{fsl\_anat} misregistration, A\textsuperscript{3}PreCore preserved
		close agreement (median Dice = 0.993; median paired $\Delta$NMI =
		$-0.00007$, $p=0.429$), indicating that robustness was not achieved through
		indiscriminate mask alteration. A\textsuperscript{3}PreCore remained broadly
		concordant with BET (median Dice = 0.955), with representative comparisons
		showing less inferior and extracranial extension. Method-blinded
		neuroscientists ranked A\textsuperscript{3}PreCore first in 35/40 participants
		(87.5\%). E-QC additionally exposed severe orientation-related disagreement
		not evident from global descriptors or centroid displacement alone. These
		findings support A\textsuperscript{3}PreCore as a robust native-FLAIR
		brain-mask transfer strategy and E-QC as a complementary framework for
		multidimensional mask characterization and participant-specific geometric
		quality control.
	\end{abstract}
	
	% ============================================================
	% KEYWORDS
	% ============================================================
	
	
	\input{sections/Sec1}
	\input{sections/Sec2}
	\input{sections/Sec3}
	\input{sections/Sec4}
	\input{sections/Sec5}
	\input{sections/Sec6}
	\input{sections/Sec7}
	\input{sections/Sec8}
	
	\section*{Data and Code Availability}
	
	The individual-level MRI data analyzed in this study are not publicly
	available because they contain human-participant imaging data and are
	subject to institutional ethics approval, informed consent, and privacy
	restrictions. Requests for access to de-identified data may be considered
	through the corresponding study investigators and are subject to the
	applicable institutional data-governance requirements.
	
	The source code implementing A\textsuperscript{3}PreCore and the analysis
	pipelines used in this study are publicly available at:
	
	\url{https://github.com/ElmiraMirzabeigi/A3PreCore.git}
	
	The repository includes the implementation, analysis scripts, and
	documentation required to reproduce the reported computational workflow.
	The version corresponding to this manuscript will be archived with a
	permanent release identifier upon publication.
	
	\section*{Acknowledgments}
	
	We want to thank the Brain-Canada Foundation and NeuroTracker, which partly funded this research. This work was funded by CQDM SynergiQc Grant \#FAubert-SYN-347\_NRFA\_2024-03-01.

	\section*{CRediT authorship contribution statement}
	
	\textbf{Elmira Mirzabeigi:}
	Conceptualization, Data curation, Formal analysis, Investigation,
	Methodology, Validation, Visualization, Writing -- original draft,
	Writing -- review \& editing.
	
	\textbf{Zahra Khalilzad:}
	Formal analysis, Investigation, Methodology, Validation,
	Writing -- review \& editing.
	
	\textbf{Taylor Snowden-Richardson:}
	Formal analysis, Investigation, Methodology, Validation,
	Writing -- review \& editing.
	
	\textbf{Jesus Eduardo Lugo Arce:}
	Conceptualization, Formal analysis, Project administration,
	Resources, Software, Supervision, Writing -- original draft,
	Writing -- review \& editing.
	
	\textbf{Jodie R. Gawryluk:}
	Funding acquisition, Validation, Resources, Software,
	Writing -- review \& editing.
	
	\textbf{Brian R. Christie:}
	Funding acquisition, Resources, Software,
	Writing -- review \& editing.
	
	\textbf{Jocelyn Faubert:}
	Conceptualization, Formal analysis, Funding acquisition,
	Project administration, Resources, Software, Supervision,
	Writing -- original draft, Writing -- review \& editing.
	
\bibliographystyle{unsrtnat}
\bibliography{references}
	
	\newpage
	\input{sections/Sec9}
	
\end{document}

%% file: sections/Sec1.tex
\section{Introduction}

% ============================================================
\subsection{Clinical and Methodological Importance of Native FLAIR-Space Brain Masking}
% ============================================================

Fluid-attenuated inversion recovery (FLAIR) magnetic resonance imaging is
central to lesion-oriented neuroimaging because suppression of
cerebrospinal-fluid signal increases the conspicuity of abnormalities in
periventricular, juxtacortical, and deep white matter
\citep{hajnal1992flair,decoene1992flair}. FLAIR is therefore widely used to characterize white matter hyperintensities
(WMHs), lacunar and periventricular abnormalities, edema, gliosis, and
demyelinating lesions across aging, cerebral small-vessel disease,
multiple sclerosis (MS), neurodegeneration, and other neurological disorders
\citep{fazekas1993pathologic,longstreth1996clinical,deleeuw2001prevalence,
	vermeer2003silent,wardlaw2013neuroimaging,duering2023strive2}. WMHs, in
particular, are associated with stroke, cognitive impairment, dementia, and
disruption of white-matter integrity, making reliable FLAIR-based
quantification important in both clinical research and population
neuroimaging
\citep{debette2010clinical,prins2015white,maniega2015white,
	debette2019clinical,darbeloff2019wmh,botz2023spatial}.

Because lesion segmentation and regional lesion-burden estimation are
commonly performed directly in native FLAIR space, many analysis workflows
require or benefit from restricting the image to a reliable intracranial
region. Numerous automated
lesion-segmentation approaches have been developed, including probabilistic,
conventional machine-learning, and deep-learning methods
\citep{schmidt2012automated,griffanti2016bianca,sudre2015bayesian,
	dadar2017performance,valverde2017automated,kuijf2019standardized,
	tran2022automatic,rahmani2024wmh}. Irrespective of the downstream algorithm,
however, errors in the native FLAIR-space brain mask may propagate into
lesion volume, regional distribution, boundary measurements, and subsequent
statistical analyses. These effects are particularly relevant near cortical,
juxtacortical, periventricular, cerebellar, and inferior brain boundaries,
where lesion signal and mask errors may occur in close spatial proximity.
Reliable native-space brain masking is therefore an important methodological
requirement for reproducible lesion-focused FLAIR analysis.

% ============================================================
\subsection{Challenges of Brain-Mask Generation in Native FLAIR Space}
% ============================================================

Generating a reliable brain mask directly from FLAIR remains challenging
because the sequence combines substantial cross-modal tissue-contrast
differences with heterogeneous acquisition geometry and
pathology-dependent signal variation. Raw FLAIR images may contain skull,
scalp, neck, spinal-cord signal, background artifacts, intensity
inhomogeneity, incomplete head coverage, or motion-related distortion.
These characteristics vary across scanners, acquisition protocols, head
positions, and fields of view and can alter the apparent separation between
intracranial and extracranial tissue.

Previous work has shown that brain extraction is sensitive to image
modality, acquisition quality, anatomical coverage, and pathology
\citep{smith2002fast,shattuck2002brainsuite,segonne2004hybrid,
	iglesias2011robust,eskildsen2012beast,kleesiek2016deep,
	isensee2019automated,hoopes2022synthstrip}. In native FLAIR space, direct
extraction may therefore produce inferior extracranial extension, inclusion
of low-signal non-brain tissue, or exclusion of cortical, temporal,
cerebellar, or inferior brain regions. Importantly, these defects can remain
spatially localized: a mask may retain a plausible total volume and centroid
while containing boundary abnormalities that are consequential for regional
lesion analysis. Evaluation based only on global mask characteristics may
therefore fail to detect clinically or methodologically relevant local
errors. Robust assessment should consequently consider both global mask
characteristics and localized boundary behaviour.

% ============================================================
\subsection{Limitations of T$_1$-to-FLAIR Brain-Mask Transfer}
% ============================================================

An alternative strategy is to define a brain mask from the T$_1$-weighted
image and propagate it into native FLAIR space. T$_1$-weighted images provide
strong anatomical contrast and are well suited to brain-mask definition using
established anatomical-processing tools. Among these, FSL's Brain Extraction
Tool (BET) is one of the most widely used approaches for brain extraction from
T$_1$-weighted MRI and also provides a relevant reference method for the
comparisons performed in the present study
\citep{smith2002fast,smith2004advances,jenkinson2012fsl}.
This strategy separates anatomical mask definition from the modality in which
lesions are ultimately quantified and can preserve a previously established
anatomical mask across modalities.

Its reliability nevertheless depends on two distinct operations: the
anatomical validity of the T$_1$-space source mask and the geometric accuracy
of its placement in native FLAIR space. A plausible source mask can yield an
unusable FLAIR-space output when transformation estimation fails.
Multimodal T$_1$-to-FLAIR alignment may be influenced by the transformation
model, similarity objective, image intensity non-uniformity, interpolation,
head position, field-of-view differences, and optimization behaviour
\citep{wells1996multi,maes1996multi,jenkinson2001global,
	jenkinson2002improved,avants2008symmetric,greve2009accurate,
	klein2010evaluation,tustison2010n4itk}.

In the \texttt{fsl\_anat}-based comparison evaluated in this study (\texttt{fsl\_anat}), the anatomical-processing workflow of the
FMRIB Software Library (FSL), provided the skull-stripped T$_1$
image and anatomical brain mask used for subsequent T$_1$-to-FLAIR
transfer. Transformation estimation was then performed between the
skull-stripped T$_1$ image and the raw full-head FLAIR image.
The two images therefore differ not only in modality-specific intensity
contrast but also in the anatomical support presented to the registration
objective: the moving image contains primarily brain tissue, whereas the
fixed image may additionally contain skull, scalp, neck, and other surrounding
structures. Although this configuration can produce satisfactory alignment,
it may be vulnerable in scans with unusual head position, strong inferior
extracranial signal, variable coverage, residual bias-field effects, or other
challenging image characteristics.

A resulting mask may retain a plausible overall size or centroid while being
substantially translated, rotated, or otherwise displaced in native FLAIR
space. Conversely, a registration method designed to increase robustness
should not achieve that robustness merely by introducing broad changes into
masks that were already successfully aligned. These considerations motivate
separate assessment of source-mask plausibility, transformation quality, and
the degree to which a modified registration strategy preserves successful
mask placement while avoiding severe participant-specific failures.

% ============================================================
\subsection{Rationale for Robust T$_1$-to-FLAIR Rigid Registration}
% ============================================================

A\textsuperscript{3}PreCore was developed to improve the robustness of the
transformation-estimation stage while preserving the anatomical brain mask
established in T$_1$ space. Rather than relying on a skull-stripped
T$_1$ image during transformation estimation, A\textsuperscript{3}PreCore
uses the full bias-corrected T$_1$-weighted image as the moving image and
the raw native FLAIR image as the fixed image. The full T$_1$ image retains
skull, scalp, head shape, and surrounding anatomical structures that are
also represented in the raw FLAIR image, thereby providing more comparable
spatial support across the two modalities.

A\textsuperscript{3}PreCore further performs six-degree-of-freedom rigid registration using normalized mutual information. Translation and rotation are estimated jointly as components of the subject-specific T$_1$-to-FLAIR transformation, allowing the optimization procedure to determine the relative pose that best aligns the full T$_1$-weighted image with native FLAIR. This optimization-based strategy is intended to reduce convergence to grossly displaced translational or rotational solutions under challenging imaging conditions. The resulting participant-specific transformation is then applied to the same T$_1$-derived anatomical brain mask used in the \texttt{fsl\_anat} comparison. A\textsuperscript{3}PreCore therefore modifies transformation estimation rather than redefining the underlying anatomical source mask.

This design leads to two complementary methodological questions. First, can
the proposed registration strategy reduce gross rigid-body misregistration
when conventional T$_1$-to-FLAIR transfer fails? Second, does it preserve
close mask geometry when \texttt{fsl\_anat} already produces a successful
registration? The second question is essential for distinguishing targeted
registration robustness from indiscriminate modification of the transferred
mask. A robust method should preferentially avoid severe participant-specific
failures while leaving already satisfactory registrations largely unchanged.

A further comparison with a mask generated directly from native FLAIR can
provide complementary information because such a comparator does not depend
on T$_1$-to-FLAIR transformation estimation. In the present study, BET was
therefore used as a secondary registration-independent comparator. This
comparison was designed to determine whether A\textsuperscript{3}PreCore
remains broadly concordant with an independently generated native-FLAIR mask
and to characterize the location and direction of residual disagreement,
including peripheral, inferior, and extracranial mask behaviour. BET was not
treated as anatomical ground truth.

% ============================================================
\subsection{Need for Multidimensional Quantitative Quality Control}
\label{subsec:need_anatomical_qc}
% ============================================================

Visual inspection remains important for neuroimaging quality control, but it
is time-consuming, potentially subjective, and difficult to standardize in
large or multisite datasets. Automated QC methods can provide reproducible
indicators and reduce manual-review burden
\citep{gedamu2008automated,esteban2017mriqc}. Their interpretation,
however, depends on the type of processing output being evaluated.
Acquisition-level tools such as MRIQC were not designed to validate derived
brain-mask geometry, and global morphological outlier strategies may miss
localized processing defects or exclude biologically plausible observations
\citep{esteban2017mriqc,monereoSanchez2021quality}. Measures such as mask
volume, centroid position, compactness, or mean within-mask intensity can
identify extreme deviations, but they may remain insensitive to localized
cortical exclusion, inferior extracranial extension, or spatially coherent
boundary displacement.

Pairwise overlap measures provide complementary but incomplete information.
Dice and Jaccard coefficients summarize spatial agreement but do not
explicitly encode the anatomical location, spatial arrangement, or physical
distance of disagreement
\citep{taha2015metrics,yeghiazaryan2018boundary}. Boundary- and
distance-based measures are therefore important when local surface
displacement is relevant. Moreover, in the absence of an independently
delineated manual reference, overlap between two automatically generated
masks cannot determine which mask is anatomically more accurate. Low overlap
establishes disagreement, whereas high overlap does not exclude systematic
errors shared by both methods. Pairwise overlap should therefore be
interpreted as inter-method agreement rather than as ground-truth accuracy.

Transfer-based workflows require an additional distinction between intrinsic
mask plausibility and registration geometry. Dedicated registration-QC
studies have used standardized visual protocols, anatomically specific
overlap measures, and learned classifiers to identify failed alignments
\citep{benhajali2020standardized,dubost2020multiatlas,fonov2022darq}.
Mutual information and normalized mutual information are established
multimodal registration criteria
\citep{pluim2003mutual,studholme1999overlap}, but they do not directly encode
anatomical landmarks or localized boundary correspondence. When the same or
a related similarity criterion is used during optimization, it should not be
interpreted as independent evidence of anatomical correctness.

Accordingly, this study developed and applied a multidimensional extended
quantitative quality-control framework (E-QC) for native FLAIR-space brain
masks. E-QC separates method-specific mask characterization from pairwise
geometric characterization. The method-specific component retains
complementary descriptors of physical mask extent, field-of-view-relative
position, FLAIR intensity support, boundary-gradient support, and inferior
morphology. The pairwise component characterizes participant-specific
disagreement using relative volume difference, physical centroid
displacement, PCA-derived principal-axis orientation difference, and
PCA-orientation-induced boundary displacement.

The purpose of this multidimensional design is not to replace mask quality
with a single aggregate score, but to preserve complementary characteristics
that may reveal different forms of disagreement. In particular, a mask can
maintain apparently plausible global characteristics while exhibiting
substantial orientation-related or localized boundary disagreement. E-QC
therefore retains method-specific mask descriptors and pairwise geometric
descriptors as complementary dimensions and does not assume that either
\texttt{fsl\_anat} or BET constitutes anatomical ground truth.

\begin{figure}[h]
	\centering
	\includegraphics[width=\textwidth]
	{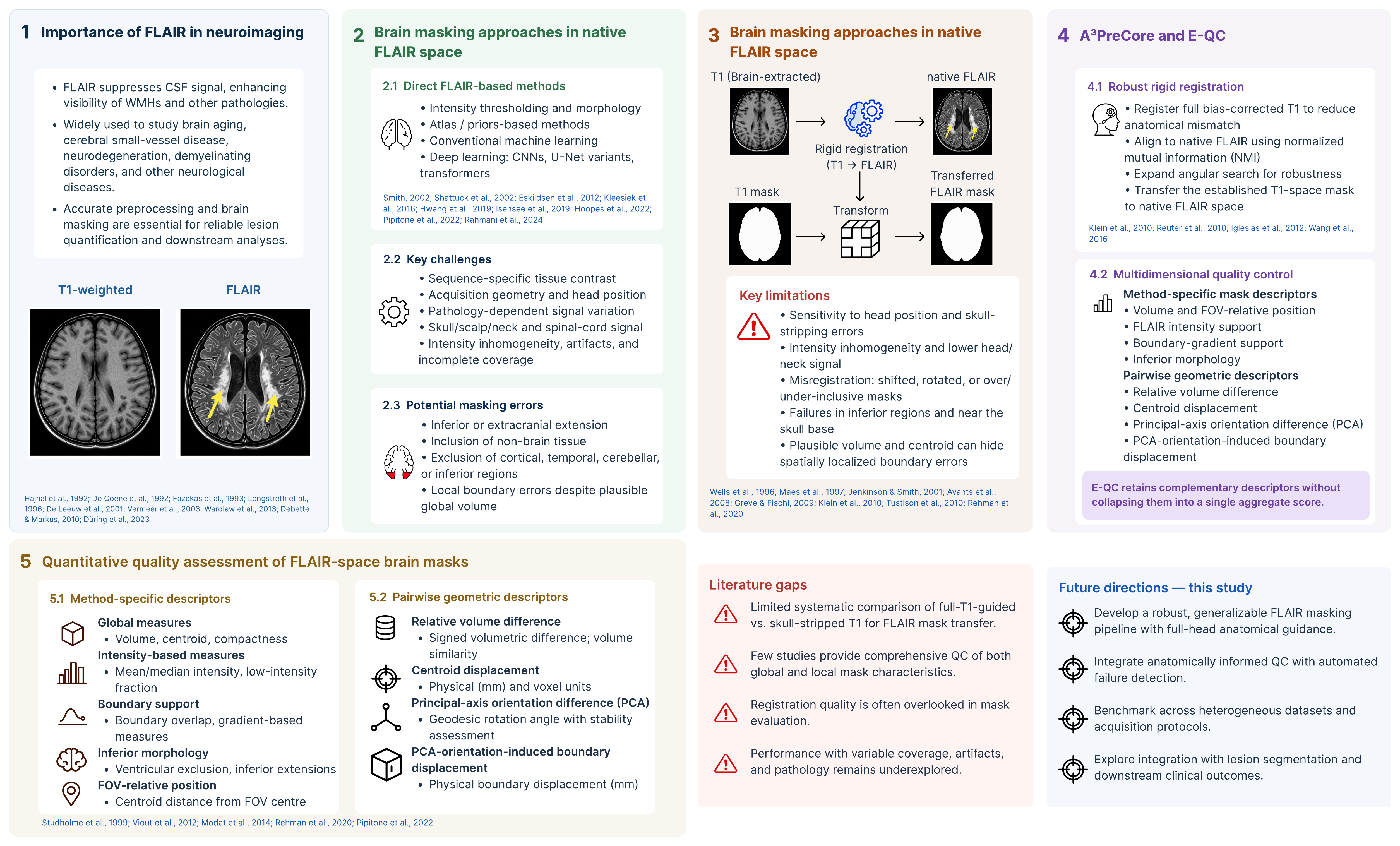}
	
	\caption{
		\textbf{Conceptual synthesis of native FLAIR-space brain masking,
			T$_1$-to-FLAIR mask transfer, and multidimensional quality assessment.}
		The schematic summarizes the clinical relevance of native FLAIR-space
		brain masking, major classes of brain-masking approaches, potential
		failure modes of T$_1$-to-FLAIR mask transfer, and design principles for
		improving rigid-registration robustness. The A\textsuperscript{3}PreCore
		strategy is illustrated as a full-T$_1$-guided, optimization-based rigid
		registration approach in which a subject-specific six-degree-of-freedom
		transformation is estimated between the full bias-corrected T$_1$ image
		and native FLAIR using normalized mutual information, with translational
		and rotational parameters optimized jointly before transfer of the
		established T$_1$-space anatomical mask. The schematic further
		illustrates the proposed extended quantitative quality-control framework
		(E-QC), which retains method-specific mask descriptors and pairwise
		geometric disagreement as complementary dimensions rather than combining
		them into a single aggregate score. Global, FLAIR-support,
		boundary-sensitive, inferior-morphology, and geometric descriptors are
		used to characterize mask behaviour and participant-specific inter-method
		disagreement. The final panel summarizes the principal literature gaps and
		the evaluation strategy used in the present study, including comparison with
		\texttt{fsl\_anat}, registration-independent
		comparison with BET, preservation analysis in successful registrations,
		gross-failure characterization, E-QC analysis, and blinded neuroscientist
		ranking.
	}
	
	\label{fig:flair_literature_synthesis}
\end{figure}

The conceptual relationship between native FLAIR masking, geometric mask
transfer, and multidimensional quality assessment is summarized in
Figure~\ref{fig:flair_literature_synthesis}.

% ============================================================
\subsection{Study Contributions}
% ============================================================
This study makes four principal contributions. 
First, it introduces A\textsuperscript{3}PreCore, an optimization-based rigid-registration strategy for native FLAIR-space brain-mask transfer that combines a full bias-corrected T$_1$ moving image with normalized-mutual-information-based six-degree-of-freedom transformation optimization while preserving the established T$_1$-space anatomical source mask.

Second, the evaluation is explicitly designed to distinguish selective
registration robustness from broad or indiscriminate alteration of mask
geometry. In addition to determining whether A\textsuperscript{3}PreCore
reduces gross rigid-body misregistration relative to \texttt{fsl\_anat},
the analysis examines whether close registration and spatial agreement are
preserved in participants for whom \texttt{fsl\_anat} already produces a
successful transfer. This distinction tests whether improved robustness is
concentrated in severe participant-specific failure cases rather than
reflecting a systematic departure from otherwise successful registrations.

Third, A\textsuperscript{3}PreCore is compared with BET as a
registration-independent method that generates a brain mask directly in
native FLAIR space. This comparison evaluates overall spatial concordance
together with regional, boundary, directional, inferior, and extracranial
patterns of inter-method disagreement, including behaviour near the skull
base and upper cervical/neck extent. Because no independently delineated
native-FLAIR brain-mask reference was available, these comparisons are
interpreted as complementary evidence of inter-method agreement and
mask-boundary behaviour rather than as direct measurements of anatomical
accuracy.

Fourth, the study introduces E-QC as a multidimensional framework combining
scalar native-FLAIR mask plausibility with participant-level geometric
characterization. E-QC is evaluated together with controlled registration
analysis, volumetric and boundary-based agreement measures, regional and
slice-wise analyses, tri-planar qualitative assessment, and method-blinded
neuroscientist ranking. These complementary evaluation layers are intended
to determine whether scalar plausibility, spatial agreement, geometric
behaviour, and expert assessment converge on the same pattern of
registration robustness while identifying failure modes that may remain
hidden when any single quality measure is considered in isolation.

%% file: sections/Sec2.tex
\section{Materials and Data}

% ============================================================
\subsection{Study Cohort and Ethics}
\label{subsec:study_dataset}
% ============================================================

This cross-sectional methodological study used baseline MRI data from the
CQDM concussion--aging cohort, a multimodal investigation of long-term
outcomes associated with concussion history in community-dwelling older
adults. Participants were recruited from the community through posters,
word of mouth, and the REACH BC participant-recruitment platform.

The baseline imaging cohort comprised 80 participants with a mean age of
65.6 $\pm$ 7.8 years (median, 64.5 years; IQR, 60.0--71.0 years; range,
50--84 years). Forty-nine participants were female (61.3\%) and 31 were
male (38.8\%). Fifty-five participants (68.8\%) had a history of
concussion and 25 (31.3\%) comprised the control group. One baseline MRI
examination per participant was included in the present study.

A readable native fluid-attenuated inversion recovery (FLAIR) image was
available for all 80 participants. Seventy-eight participants additionally
had the paired T$_1$-weighted and FLAIR images required for
registration-based analyses. The two participants without complete paired
inputs were classified as input-ineligible rather than as processing
failures.

Participants were not excluded because of subsequently observed
registration failure, spatial disagreement, or abnormal mask-boundary
behaviour. Such cases were retained because participant-specific failure
modes constituted an explicit component of the methodological evaluation.
The dataset therefore captured participant- and scan-level variability in
anatomy, concussion history, age-related and vascular imaging findings,
head position, image coverage, and image quality within a single-site
acquisition setting.

The study was approved by the University of Victoria Human Research Ethics
Board (protocol 21-0591) and the University of Montreal Comit\'e
d'\'ethique de la recherche clinique (protocol 2024-5858). All
participants provided written informed consent before participation.

% ============================================================
\subsection{MRI Acquisition and Image Inputs}
\label{subsec:mri_acquisition}
% ============================================================

MRI examinations were acquired at West Coast Medical Imaging using a
3~T GE SIGNA Pioneer system by a trained MRI technologist, with a research
team member present during each scanning session. The present analysis used
the T$_1$-weighted anatomical and FLAIR sequences required for anatomical
brain-mask definition, T$_1$-to-FLAIR registration, and evaluation in
native FLAIR space. Diffusion MRI data collected as part of the parent
study were not used in the present analysis.

The T$_1$-weighted anatomical acquisition used a three-dimensional sagittal
BRAVO sequence, whereas the FLAIR acquisition used a three-dimensional
sagittal T$_2$-weighted CUBE FLAIR sequence. Both sequences were acquired
using the GE ``Head 32'' receive coil. The scanner software reported in the
DICOM metadata was MR software release PX26.1\_R03\_2128.b.
Sequence-specific acquisition and reconstruction parameters are summarized
in Table~\ref{tab:mri_acquisition} and were extracted directly from the
original DICOM metadata.

For each participant, the native FLAIR image, denoted $F$, served as the
fixed image and defined the target coordinate system for registration,
brain-mask evaluation, and all inter-method comparisons. The corresponding
T$_1$-weighted image provided the anatomical information used to define
the source brain mask and estimate its placement in native FLAIR space.

T$_1$ preprocessing provided a full bias-corrected image, $T$, a
skull-stripped representation, $T_B$, and an anatomical brain mask,
$M_T^{\mathrm{anat}}$. The same anatomical source mask was used by both
registration-based approaches, allowing differences in the resulting
native-FLAIR masks to reflect differences in T$_1$-to-FLAIR transformation
estimation rather than differences in source-mask definition.

All resulting masks were represented and evaluated directly on the native
FLAIR grid. Native-space evaluation was used to avoid additional spatial
normalization or resampling that could obscure errors specifically related
to T$_1$-to-FLAIR mask placement.

\begin{table}[htbp]
	\centering
	
	\caption{MRI acquisition and reconstruction parameters for the
		T$_1$-weighted and FLAIR sequences.}
	\label{tab:mri_acquisition}
	
	{
		\small
		\setlength{\tabcolsep}{4pt}
		\renewcommand{\arraystretch}{1.15}
		
		\begin{tabular}{
				@{}
				p{0.34\linewidth}
				p{0.27\linewidth}
				p{0.27\linewidth}
				@{}
			}
			
			\hline
			
			\textbf{Parameter}
			& \textbf{T$_1$-weighted}
			& \textbf{FLAIR}
			\\
			
			\hline
			
			Scanner
			& GE SIGNA Pioneer
			& GE SIGNA Pioneer
			\\
			
			Field strength
			& 3~T
			& 3~T
			\\
			
			Receiver coil
			& Head 32
			& Head 32
			\\
			
			Sequence
			& 3D sagittal BRAVO
			& 3D sagittal T$_2$ CUBE FLAIR
			\\
			
			Acquisition type
			& 3D
			& 3D
			\\
			
			Orientation
			& Sagittal
			& Sagittal
			\\
			
			Repetition time, TR (ms)
			& 8.564
			& 6802
			\\
			
			Echo time, TE (ms)
			& 3.296
			& 95.19
			\\
			
			Inversion time, TI (ms)
			& 450
			& 1899
			\\
			
			Flip angle ($^\circ$)
			& 10
			& 90
			\\
			
			Reconstructed voxel spacing (mm$^3$)
			& $1.0 \times 1.0 \times 0.5$
			& $0.4883 \times 0.4883 \times 0.7$
			\\
			
			Slice thickness (mm)
			& 1.0
			& 1.4
			\\
			
			Reconstructed slice spacing (mm)
			& 0.5
			& 0.7
			\\
			
			Acquisition matrix
			& $256 \times 256$
			& $240 \times 220$
			\\
			
			Reconstruction matrix
			& $256 \times 256$
			& $512 \times 512$
			\\
			
			In-plane field of view (mm$^2$)
			& $256 \times 256$
			& $250 \times 250$
			\\
			
			Number of reconstructed slices
			& 380
			& 252
			\\
			
			Echo-train length
			& 1
			& 185
			\\
			
			Parallel imaging
			& ARC, $2.0 \times 1.0$
			& ARC, $2.5 \times 1.5$
			\\
			
			Acquisition duration
			& 296.7~s (4:56.7)
			& 309.4~s (5:09.4)
			\\
			
			\hline
			
		\end{tabular}
	}
	
	\vspace{2mm}
	
	\begin{minipage}{0.94\linewidth}
		\footnotesize
		Voxel spacing and slice counts refer to the reconstructed DICOM image
		series. Parallel-imaging factors are reported as in-plane $\times$
		slice-encoding acceleration. The T$_1$-weighted and FLAIR acquisitions
		contained 380 and 252 reconstructed sagittal images, respectively.
	\end{minipage}
	
\end{table}

% ============================================================
\subsection{Analysis Population and Mask Data}
\label{subsec:analysis_population}
% ============================================================

Direct FLAIR brain extraction using BET required only a readable native
FLAIR image and was therefore available for all 80 participants.
Registration-based processing required paired T$_1$-weighted and FLAIR
images and was available for 78 participants. To ensure participant-level
comparability, all primary inter-method analyses were performed in this
common 78-participant cohort. Participants were retained regardless of
subsequent registration or mask-quality classification, including cases of
gross rigid-body misregistration, because exclusion of such cases would
remove a principal failure mode under investigation.

We evaluated three candidate brain masks in native FLAIR space. BET
generated a mask directly from native FLAIR,
$M_F^{\mathrm{BET}}$, and therefore served as a
registration-independent comparator. The \texttt{fsl\_anat} procedure propagated $M_T^{\mathrm{anat}}$ to native FLAIR
space after estimating alignment from the skull-stripped T$_1$ image
$T_B$, producing $M_F^{\mathrm{FSL}}$. A\textsuperscript{3}PreCore
estimated rigid alignment using the full bias-corrected T$_1$ image $T$
and propagated the same anatomical source mask, producing
$M_F^{\mathrm{A3P}}$.

No independently delineated manual brain masks were available in native
FLAIR space. Consequently, none of the automatically generated masks,
including BET, was treated as anatomical ground truth. Pairwise analyses
were therefore interpreted as assessments of inter-method agreement rather
than direct measures of anatomical accuracy. Detailed registration
procedures, quantitative agreement measures, quality-control analyses, and
visual and expert evaluations are described in the Methods section.

%% file: sections/Sec3.tex
\section{A\textsuperscript{3}PreCore: Full-T1-Guided Registration for Native-FLAIR Brain-Mask Transfer}
\label{sec:a3precore_method}

% ============================================================
\subsection{Methodological Rationale}
\label{subsec:a3precore_rationale}
% ============================================================

A\textsuperscript{3}PreCore was developed to address a geometric failure
mode in registration-based native-FLAIR preprocessing. A plausible anatomical
brain mask may be obtained in T$_1$ space, yet its final placement in native
FLAIR space depends on the transformation used for T$_1$-to-FLAIR
registration. Anatomical mask definition and geometric mask placement were
therefore treated as distinct operations: the former determines which
T$_1$-space voxels belong to the brain, whereas the latter determines where
that fixed anatomical mask is positioned after transformation to native
FLAIR space.

Let $F$ denote the raw native FLAIR image, $T$ the full bias-corrected
T$_1$-weighted image, $T_B$ the skull-stripped bias-corrected T$_1$ image,
and $M_T^{\mathrm{anat}}$ the anatomical brain mask generated during
\texttt{fsl\_anat} preprocessing. The two registration-based workflows used
the same source anatomical mask and the same native FLAIR reference, but
estimated the T$_1$-to-FLAIR transformation from different moving-image
representations:
\begin{align}
\text{\texttt{fsl\_anat}:}\qquad & T_B \rightarrow F, \\
\text{A\textsuperscript{3}PreCore:}\qquad & T \rightarrow F.
\end{align}

A\textsuperscript{3}PreCore therefore does not redefine
$M_T^{\mathrm{anat}}$ and is not a direct FLAIR skull-stripping method.
Instead, it modifies the transformation-estimation strategy used to place
the same T$_1$-derived anatomical mask in native FLAIR space. The full
bias-corrected T$_1$ image retains head contour, skull, scalp, and inferior
extracranial structures that are removed from $T_B$ but remain represented
in the raw FLAIR acquisition. We hypothesized that retaining this broader
spatial context, together with normalized mutual information and an extended
rotational search, would reduce severe translational and rotational placement
errors in challenging registrations.

\begin{figure}[h]
    \centering
    \includegraphics[width=\textwidth]
    {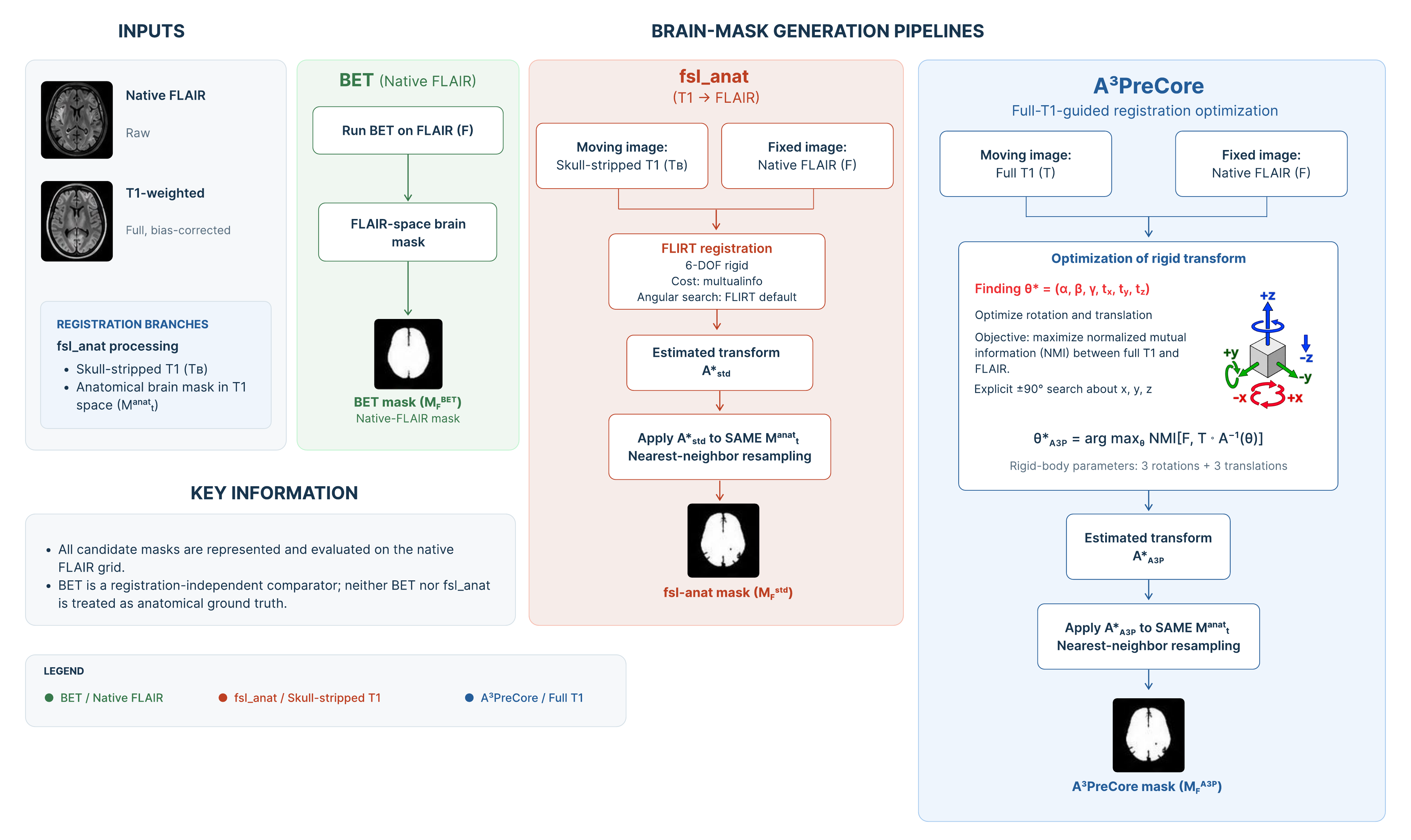}
    \caption{
    \textbf{Native-FLAIR brain-mask transfer and comparison workflows.}
    BET generated a brain mask directly from the native FLAIR image, $F$.
The \texttt{fsl\_anat} and A\textsuperscript{3}PreCore workflows
propagated the same T$_1$-space anatomical mask,
$M_T^{\mathrm{anat}}$, but differed in transformation estimation.
The \texttt{fsl\_anat} registered the skull-stripped
T$_1$ image, $T_B$, to $F$ using FLIRT with \texttt{mutualinfo},
whereas A\textsuperscript{3}PreCore registered the full bias-corrected
T$_1$ image, $T$, to $F$ using \texttt{normmi}.
Both workflows estimated subject-specific six-degree-of-freedom rigid
transformations comprising rotational and translational parameters and
applied the resulting transformation to the same anatomical source mask
using nearest-neighbour resampling. All candidate masks were evaluated
on the native FLAIR grid.
    }
    \label{fig:a3precore_methodology}
\end{figure}

% ============================================================
\subsection{\texttt{fsl\_anat} T$_1$-to-FLAIR Transfer Comparator}
\label{subsec:standard_transfer}
% ============================================================

The \texttt{fsl\_anat} served as the registration-based methodological
comparator. It used the skull-stripped T$_1$ image and anatomical brain mask
produced during \texttt{fsl\_anat} preprocessing; the subsequent
T$_1$-to-FLAIR registration and mask-transfer operations were performed as
study-specific processing steps.

The transformation was estimated by registering the skull-stripped
bias-corrected T$_1$ image, $T_B$, to the raw native FLAIR image, $F$:
\begin{equation}
T_B \rightarrow F.
\end{equation}
The resulting transformation was then applied to
$M_T^{\mathrm{anat}}$ to obtain the corresponding native-FLAIR-space mask.

This workflow provides a controlled registration-based comparator because
the anatomical source mask is identical to that used by
A\textsuperscript{3}PreCore. Consequently, differences between the resulting
native-FLAIR masks can be attributed to differences in transformation
estimation rather than source-mask anatomy.

A potential limitation of the \texttt{fsl\_anat} is the difference in
anatomical support between the moving and fixed images. The moving image
$T_B$ contains predominantly intracranial tissue, whereas the raw FLAIR
reference may retain skull, scalp, inferior neck structures, spinal-canal
signal, and additional non-brain foreground. This mismatch may reduce the
spatial constraints available to multimodal registration in participants
with atypical head position, variable inferior coverage, intensity
inhomogeneity, motion-related artifact, or prominent extracranial FLAIR
signal.

% ============================================================
\subsection{A\textsuperscript{3}PreCore Full-T$_1$ Registration Strategy}
\label{subsec:a3precore_strategy}
% ============================================================

A\textsuperscript{3}PreCore replaced the skull-stripped moving image with
the full bias-corrected T$_1$ image, \texttt{T1\_biascorr.nii.gz}.
FLIRT estimated a subject-specific six-degree-of-freedom rigid
transformation from $T$ to $F$ using the \texttt{normmi} cost function,
with three rotational parameters $(\alpha,\beta,\gamma)$ and three
translational parameters $(t_x,t_y,t_z)$ optimized jointly as part of the
rigid transformation.
Separately, the implemented FLIRT command explicitly specified angular
search bounds of $-90^{\circ}$ to $+90^{\circ}$ for each image axis.
These bounds define the angular domain explored during the initial search
stage and are distinct from the subject-specific rotational parameters
$(\alpha,\beta,\gamma)$ estimated by the registration optimization; they
do not represent fixed rotations applied to the FLAIR image.
The workflow otherwise retained FLIRT's default initialization and
optimization schedule. No input masks, voxel-weighting images, cropping
operations, or additional reorientation options were supplied during
transformation estimation.

The workflow retained FLIRT's default initialization and optimization
schedule. No input masks, voxel-weighting images, cropping operations, or
additional reorientation options were supplied during transformation
estimation. Because no explicit intensity-image interpolation option was
specified, FLIRT used its default trilinear interpolation when generating
the registered full-T$_1$ image.

Let the transformation estimated by A\textsuperscript{3}PreCore be
$A_{\mathrm{A3P}}^{*}$. After registration, this transformation was applied
to the same $M_T^{\mathrm{anat}}$ used by the \texttt{fsl\_anat} comparator.
Accordingly, the resulting mask,
$M_F^{\mathrm{A3P}}$, represents the placement of a fixed anatomical source
mask under the A\textsuperscript{3}PreCore registration strategy rather than
a newly estimated anatomical segmentation.

% ============================================================
\subsection{Rigid Registration Model and Multimodal Similarity Objectives}
\label{subsec:rigid_registration}
% ============================================================

Both registration-based workflows modelled the T$_1$-to-FLAIR mapping as a
six-degree-of-freedom rigid-body transformation. Because the T$_1$ and
FLAIR acquisitions were obtained from the same participant, the transformation
was restricted to translations and rotations and deliberately excluded
scaling, shearing, and nonlinear deformation.
Let
\begin{equation}
\theta =
(\alpha,\beta,\gamma,t_x,t_y,t_z)^{\top}
\in \Theta \subset \mathbb{R}^{6}
\end{equation}
denote the rigid-transformation parameters, where
$\alpha$, $\beta$, and $\gamma$ are rotations and
$t_x$, $t_y$, and $t_z$ are translations. For a physical coordinate
$x\in\Omega_T$, the mapping from T$_1$ space to native FLAIR space was
represented as
\begin{equation}
\widetilde{y}
=
A_{T\rightarrow F}(\theta)\widetilde{x},
\qquad
A_{T\rightarrow F}(\theta)\in\mathrm{SE}(3),
\end{equation}
with
\begin{equation}
A_{T\rightarrow F}(\theta)
=
\begin{bmatrix}
R(\alpha,\beta,\gamma) & t \\
0^{\top} & 1
\end{bmatrix},
\qquad
t=
\begin{bmatrix}
t_x\\
t_y\\
t_z
\end{bmatrix},
\end{equation}
and
\begin{equation}
R(\alpha,\beta,\gamma)
=
R_z(\gamma)R_y(\beta)R_x(\alpha).
\end{equation}

Because T$_1$-weighted and FLAIR images exhibit different tissue-contrast
relationships, entropy-based multimodal similarity criteria were used rather
than direct voxel-intensity differences
\citep{wells1996multi,studholme1999overlap}. For a candidate transformation
$\theta$, let
\begin{equation}
I_{\theta}(y)
=
I\left(A_{T\rightarrow F}^{-1}(\theta)y\right),
\qquad y\in\Omega_F,
\end{equation}
denote the moving image sampled on the native FLAIR grid.

The \texttt{fsl\_anat} used FLIRT's mutual-information objective, represented
conceptually as
\begin{equation}
\theta_{\mathrm{fsl\_anat}}^{*}
=
\arg\max_{\theta\in\Theta}
\operatorname{MI}
\left[
F,\,
T_B\circ A_{T\rightarrow F}^{-1}(\theta)
\right],
\end{equation}
where
\begin{equation}
\operatorname{MI}(F,I_{\theta})
=
H(F)+H(I_{\theta})-H(F,I_{\theta}).
\end{equation}

A\textsuperscript{3}PreCore used normalized mutual information,
conceptually expressed as
\begin{equation}
\operatorname{NMI}(F,I_{\theta})
=
\frac{H(F)+H(I_{\theta})}
{H(F,I_{\theta})},
\end{equation}
with
\begin{equation}
\theta_{\mathrm{A3P}}^{*}
=
\arg\max_{\theta\in\Theta}
\operatorname{NMI}
\left[
F,\,
T\circ A_{T\rightarrow F}^{-1}(\theta)
\right].
\end{equation}

Here, $\alpha$, $\beta$, and $\gamma$ are subject-specific rotational
parameters estimated jointly with translation during rigid optimization.
These optimized rotational parameters are distinct from the angular-search
bounds used by FLIRT during its initial search procedure. The corresponding rigid transformation matrices were
\begin{equation}
A_{\mathrm{fsl\_anat}}^{*}
=
A_{T\rightarrow F}(\theta_{\mathrm{fsl\_anat}}^{*}),
\qquad
A_{\mathrm{A3P}}^{*}
=
A_{T\rightarrow F}(\theta_{\mathrm{A3P}}^{*}).
\end{equation}

These expressions describe the conceptual optimization objectives; numerical
estimation used FLIRT with its internal optimization schedule.
Because normalized mutual information formed part of the
A\textsuperscript{3}PreCore optimization objective, subsequent controlled
NMI analyses were interpreted as registration-similarity characterization
rather than as independent evidence of anatomical accuracy.

% ============================================================
\subsection{Anatomical Mask Propagation and Native-FLAIR Outputs}
\label{subsec:mask_transfer}
% ============================================================

Each estimated transformation was applied to the same binary anatomical
source mask,
$M_T^{\mathrm{anat}}:\Omega_T\rightarrow\{0,1\}$.
The raw native FLAIR image defined the reference geometry, and binary masks
were resampled using nearest-neighbour interpolation.

For $y\in\Omega_F$, the transferred masks were written as
\begin{align}
M_F^{\mathrm{fsl\_anat}}(y)
&=
\operatorname{NN}
\left[
M_T^{\mathrm{anat}}
\left(
(A_{\mathrm{fsl\_anat}}^{*})^{-1}y
\right)
\right],
\\
M_F^{\mathrm{A3P}}(y)
&=
\operatorname{NN}
\left[
M_T^{\mathrm{anat}}
\left(
(A_{\mathrm{A3P}}^{*})^{-1}y
\right)
\right].
\end{align}

Nearest-neighbour interpolation preserved the discrete values of the binary
source mask. No additional thresholding, connected-component filtering,
hole filling, or other mask-cleanup operation was applied immediately after
resampling.

The corresponding non-dilated brain-extracted native FLAIR images were
generated as
\begin{equation}
F_{\mathrm{brain}}^{(m)}(y)
=
F(y)M_F^{(m)}(y),
\qquad
m\in\{\mathrm{fsl\_anat},\mathrm{A3P}\}.
\end{equation}

For the configured permissive output, one
\texttt{fslmaths -dilM} operation was subsequently applied to each
registration-derived mask:
\begin{equation}
M_{F,\mathrm{dil}}^{(m)}
=
\operatorname{Dilate}_{\mathrm{dilM}}
\left(M_F^{(m)}\right),
\end{equation}
followed by
\begin{equation}
F_{\mathrm{brain,dil}}^{(m)}(y)
=
F(y)M_{F,\mathrm{dil}}^{(m)}(y).
\end{equation}

The primary mask analyses described in
Section~\ref{sec:experimental_evaluation} used the configured dilated mask
variants. Dilation was therefore part of the reported analysis pipeline
rather than an unreported downstream visualization step.

% ============================================================
\subsection{Methodological Contrast and Technical Output Validation}
\label{subsec:methodological_contrast}
% ============================================================

The \texttt{fsl\_anat} and A\textsuperscript{3}PreCore workflows shared the same
native FLAIR reference, T$_1$-derived anatomical source mask, six-degree-
of-freedom transformation model, nearest-neighbour mask resampling,
configured dilation step, and native-FLAIR output geometry. The implemented
workflows differed in three registration components: the moving T$_1$
representation, multimodal similarity objective, and rotational-search
specification.

The \texttt{fsl\_anat} used the skull-stripped image $T_B$,
\texttt{mutualinfo}, and FLIRT's default angular-search configuration,
whereas A\textsuperscript{3}PreCore used the full bias-corrected image $T$,
\texttt{normmi}, and an explicit $\pm90^{\circ}$ search around all three
axes. The principal methodological comparison therefore evaluates the two
registration strategies as implemented and should not be interpreted as
isolating the effect of the moving-image representation alone.

The contrast can be summarized as
\begin{align}
M_T^{\mathrm{anat}}
&\xrightarrow[
\substack{\texttt{mutualinfo}\\
\text{default angular search}}
]{T_B\rightarrow F}
M_F^{\mathrm{fsl\_anat}},
\\
M_T^{\mathrm{anat}}
&\xrightarrow[
\substack{\texttt{normmi}\\
\pm90^{\circ}\text{ angular search}}
]{T\rightarrow F}
M_F^{\mathrm{A3P}}.
\end{align}

For every participant, generated registered images and masks were checked
for three-dimensional readability and compatibility with the corresponding
native FLAIR image. Output dimensions were required to match exactly, and
affine matrices were required to agree within an absolute element-wise
tolerance of $10^{-4}$.

Missing or unreadable outputs and native-grid incompatibility were recorded
separately from anatomical implausibility or registration failure.
Importantly, outputs exhibiting adverse spatial behaviour or gross
misregistration were retained rather than removed after technical
validation, because such failures constituted outcomes of methodological
interest. Detailed quality-control, pairwise comparison, registration-
assessment, and statistical procedures are described in
Sections~\ref{sec:extended_qc} and
\ref{sec:experimental_evaluation}.

%% file: sections/Sec4.tex
\section{E-QC: Multidimensional Quality Control for Native-FLAIR Brain Masks}
\label{sec:extended_qc}

% ============================================================
\subsection{Rationale and Design Principles}
\label{subsec:eqc_rationale}
% ============================================================

Conventional global quality-control measures may fail to identify
participant-specific abnormalities in automatically generated brain masks
\citep{mortamet2009automatic,esteban2017mriqc,kim2019loni}.
A candidate mask may retain a plausible total volume, remain near the centre
of the acquisition field of view, and exhibit apparently reasonable FLAIR
signal characteristics while still being displaced, differently oriented,
or locally extended into non-brain tissue. Conversely, localized boundary
abnormalities may occur without producing marked changes in global mask
characteristics.

These limitations are particularly relevant when comparing direct
FLAIR-space brain extraction with registration-based mask-transfer workflows.
A transferred mask may preserve a plausible global size and morphology despite
an incorrect T$_1$-to-FLAIR transformation, whereas a directly generated
native-FLAIR mask avoids transformation-estimation failure but may remain
susceptible to modality-specific extraction errors.

We therefore developed extended quantitative quality control (E-QC) as a
multidimensional framework that preserves complementary information rather
than collapsing heterogeneous descriptors into a single quality score.
E-QC comprises two analytically distinct levels:

\begin{enumerate}

    \item \textbf{Method-specific mask characterization.}
    Each candidate mask is described independently using measures of physical
    extent, field-of-view-relative position, FLAIR intensity support,
    boundary-gradient support, peripheral foreground occupation, and
    inferior morphology.

    \item \textbf{Pairwise geometric characterization.}
    For each predefined pair of candidate masks, disagreement is described
    using relative volume difference, physical centroid displacement,
    principal-axis orientation difference, and the physical-scale boundary
    consequence of that orientation difference.

\end{enumerate}

The method-specific descriptors characterize how an individual candidate
mask relates to its own native FLAIR image and acquisition geometry. The
pairwise descriptors instead characterize disagreement between two candidate
masks. We did not interpret either level as anatomical ground truth.

Importantly, no arithmetic mean, geometric mean, minimum operator, learned
weighting scheme, or other scalarization was used to combine E-QC
descriptors. This design preserves potentially discordant dimensions of mask
behaviour that would otherwise be obscured by compensation within an
aggregate score.
For a candidate mask \(M_a\), native FLAIR image \(F\), and predefined
comparator \(M_b\), E-QC therefore retains
\begin{equation}
\mathcal{P}(M_a,F)
=
\left[
V_a,\;
C_a,\;
R_{I,a},\;
B_{G,a},\;
P_{\mathrm{peripheral},a},\;
P_{\mathrm{tail},a}
\right]^{\mathsf T},
\label{eq:eqc_method_profile}
\end{equation}
together with the pairwise vector
\begin{equation}
\mathbf{Q}_{\mathrm{E\text{-}QC}}^{(a,b)}
=
\left[
\delta_V^{(a,b)},\;
D_C^{(a,b)},\;
\theta_{\mathrm{PCA}}^{(a,b)},\;
D_{\mathrm{PCA}}^{(a,b)}
\right]^{\mathsf T}.
\label{eq:eqc_pairwise_profile}
\end{equation}

\begin{figure}[h]
    \centering
    \includegraphics[
        width=.9\textwidth
    ]{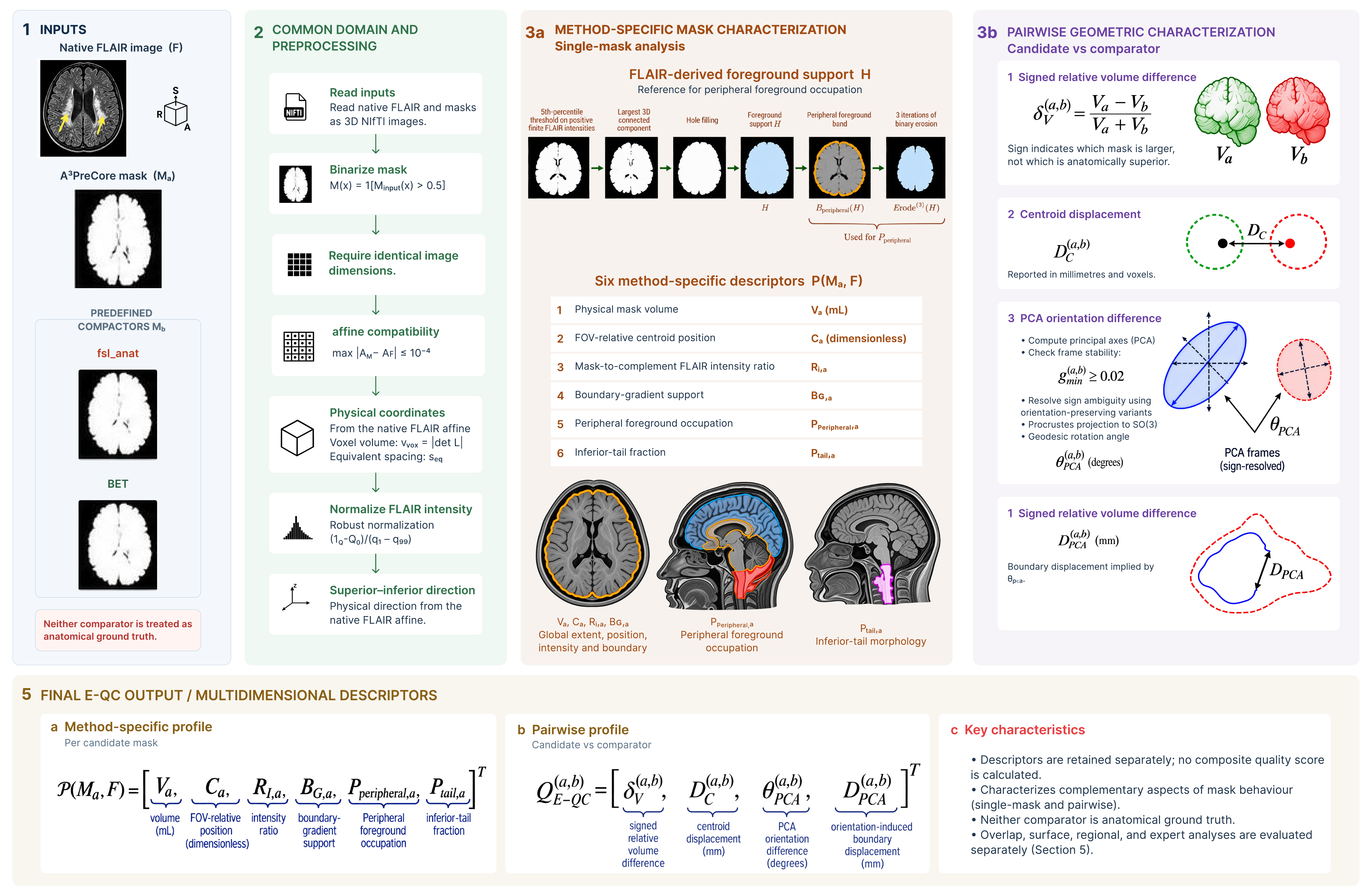}
\caption{
\textbf{Overview of the extended quantitative quality-control (E-QC)
framework for native-FLAIR brain-mask assessment.}
E-QC receives the native FLAIR image, \(F\), a candidate mask, \(M_a\),
and a prespecified comparator, \(M_b\), after common input validation and
native-space preprocessing. The framework separates into two complementary
analytical branches. Method-specific mask characterization independently
retains physical volume, field-of-view-relative position, FLAIR intensity
support, boundary-gradient support, peripheral foreground occupation,
and inferior-tail morphology. Pairwise geometric
characterization quantifies signed relative volume difference, physical
centroid displacement, principal-axis orientation difference, and the
corresponding PCA-orientation-induced boundary displacement, with PCA-based
descriptors reported only when the predefined eigengap stability criterion
is satisfied. The resulting method-specific and pairwise profiles are
retained as separate multidimensional descriptors and are not reduced to a
composite quality score. The \texttt{fsl\_anat} serves as the
primary methodological comparator and BET as the registration-independent
native-FLAIR comparator; neither is treated as anatomical ground truth.
}
    \label{fig:eqc_framework}
\end{figure}

The \texttt{fsl\_anat} served as the primary methodological
comparator, whereas BET served as the registration-independent native-FLAIR
comparator. Comparator assignment was specified a priori at the analysis
level and was not influenced by E-QC results. Neither comparator was treated
as anatomical ground truth. Figure~\ref{fig:eqc_framework} summarizes the
complete E-QC workflow, including input validation, method-specific mask
characterization, pairwise geometric characterization, and retention of the
resulting multidimensional descriptor profiles without reduction to a
composite quality score.

% ============================================================
\subsection{Common Image Domain and Input Validation}
\label{subsec:eqc_notation}
% ============================================================

Let \(F(x)\) denote native FLAIR intensity at voxel
\(x\in\Omega\), where \(\Omega\) represents the complete three-dimensional
image domain. Each input mask was converted to a binary representation,
\begin{equation}
M(x)
=
\mathbb{1}
\left[
M_{\mathrm{input}}(x)>0.5
\right],
\qquad
M(x)\in\{0,1\},
\label{eq:eqc_binary_mask}
\end{equation}
where \(\mathbb{1}[\cdot]\) denotes the indicator function.
The number of foreground voxels was
\begin{equation}
|M|
=
\sum_{x\in\Omega}M(x).
\label{eq:eqc_mask_size}
\end{equation}

E-QC proceeded only when the candidate mask and corresponding FLAIR image
were readable three-dimensional NIfTI images with identical matrix
dimensions. Affine matrices were considered compatible when
\begin{equation}
\max_{i,j}
\left|
A_{M,ij}-A_{F,ij}
\right|
\leq10^{-4},
\label{eq:eqc_affine_compatibility}
\end{equation}
using an absolute element-wise tolerance and zero relative tolerance.
Voxel indices were mapped to physical coordinates using the native FLAIR
voxel-to-world affine
\citep{cox2004nifti},
\begin{equation}
\tilde p(x)
=
A_F
\begin{bmatrix}
x_1 & x_2 & x_3 & 1
\end{bmatrix}^{\mathsf T},
\qquad
p(x)=\tilde p_{1:3}(x),
\label{eq:eqc_physical_coordinates}
\end{equation}
where \(p(x)\in\mathbb{R}^{3}\) is expressed in millimetres.
Let \(L\in\mathbb{R}^{3\times3}\) denote the linear spatial component of
\(A_F\). The physical volume represented by one voxel was
\begin{equation}
v_{\mathrm{vox}}
=
|\det(L)|.
\label{eq:eqc_voxel_volume}
\end{equation}

The corresponding physical voxel dimensions were denoted
\(s_1,s_2,s_3\). Their geometric mean,
\begin{equation}
s_{\mathrm{eq}}
=
\left(
s_1s_2s_3
\right)^{1/3},
\label{eq:eqc_equivalent_spacing}
\end{equation}
was used when a single physical length scale was required.

% ============================================================
\subsection{FLAIR Intensity and Spatial Preprocessing}
\label{subsec:eqc_common_preprocessing}
% ============================================================

For descriptors requiring image-intensity or gradient information, positive
finite FLAIR intensities were robustly normalized
\citep{esteban2017mriqc}.
Let \(q_1\) and \(q_{99}\) denote the 1st and 99th percentiles of positive
finite FLAIR intensities. For \(q_{99}>q_1\),
\begin{equation}
F_{\mathrm{norm}}(x)
=
\operatorname{clip}
\left(
\frac{F(x)-q_1}
     {q_{99}-q_1},
0,1
\right).
\label{eq:eqc_flair_normalization}
\end{equation}

Degenerate intensity distributions for which this normalization could not
be evaluated were treated as invalid for the affected descriptor rather than
assigned an artificial score.

The anatomical inferior--superior direction was determined from the native
FLAIR affine. Let \(\mathbf{u}_{\mathrm{SI}}\) denote the unit vector pointing
in the physical superior direction. The corresponding physical
inferior--superior coordinate was
\begin{equation}
z_{\mathrm{SI}}(x)
=
\mathbf{u}_{\mathrm{SI}}^{\mathsf T}p(x).
\label{eq:eqc_si_coordinate}
\end{equation}

The anatomical orientation encoded by the affine, rather than raw
voxel-index ordering alone, was used to establish the inferior direction
for morphological characterization.

% ============================================================
\subsection{Method-Specific Mask Characterization}
\label{subsec:eqc_method_specific}
% ============================================================

The six components of
Equation~\ref{eq:eqc_method_profile} characterize complementary aspects of
individual candidate-mask behaviour. They were retained and analyzed
separately because they differ in units, physical interpretation, and
sensitivity to acquisition geometry.

% ------------------------------------------------------------
\subsubsection{Physical Mask Volume}
\label{subsubsec:eqc_volume}
% ------------------------------------------------------------

The physical volume of candidate mask \(M\) was
\begin{equation}
V_M
=
v_{\mathrm{vox}}
\sum_{x\in\Omega}M(x),
\label{eq:eqc_mask_volume}
\end{equation}
and was reported in millilitres after division by \(1000\).
Mask volume provides a global measure of spatial extent but does not identify
the anatomical location of disagreement
\citep{taha2015metrics}. Accordingly, \(V_M\) was interpreted
descriptively rather than as evidence that a larger or smaller candidate mask
was anatomically superior.

% ------------------------------------------------------------
\subsubsection{Geometric Centroid and FOV-Relative Position}
\label{subsubsec:eqc_center}
% ------------------------------------------------------------

For a non-empty candidate mask, its physical geometric centroid was
\begin{equation}
c_M
=
\frac{1}{|M|}
\sum_{x:M(x)=1}p(x).
\label{eq:eqc_mask_centroid}
\end{equation}

The physical centre of the complete native FLAIR field of view was
\begin{equation}
c_{\Omega}
=
\frac{1}{|\Omega|}
\sum_{x\in\Omega}p(x),
\label{eq:eqc_fov_center}
\end{equation}
and the physical field-of-view radius was
\begin{equation}
r_{\Omega}
=
\max_{x\in\Omega}
\left\|
p(x)-c_{\Omega}
\right\|_2.
\label{eq:eqc_fov_radius}
\end{equation}

The normalized FOV-relative centroid offset was
\begin{equation}
C_M
=
\frac{
\left\|
c_M-c_{\Omega}
\right\|_2
}{
r_{\Omega}
}.
\label{eq:eqc_fov_relative_center}
\end{equation}
\(C_M\) is dimensionless and describes mask position relative to the physical
extent of the acquisition field of view. Because this reference is determined
by acquisition geometry rather than by an anatomical landmark, valid
differences in head position, cropping, or inferior image coverage can alter
\(C_M\). It was therefore interpreted as a positional descriptor rather than
as an anatomical-centring score
\citep{kim2019loni}.

% ------------------------------------------------------------
\subsubsection{Mask-to-Complement FLAIR Intensity Ratio}
\label{subsubsec:eqc_intensity}
% ------------------------------------------------------------

To characterize relative FLAIR intensity support, the positive-intensity
complement of a candidate mask was defined as
\begin{equation}
\overline{M}_{+}
=
\left\{
x\in\Omega:
M(x)=0,\;
F(x)>0,\;
F(x)\ \mathrm{finite}
\right\}.
\label{eq:eqc_positive_complement}
\end{equation}

The robust mask-to-complement squared-intensity ratio was
\begin{equation}
R_I(M,F)
=
\frac{
\operatorname{median}
\left\{
|F(x)|^2:
M(x)=1,\;
F(x)\ \mathrm{finite}
\right\}
}{
\operatorname{median}
\left\{
|F(x)|^2:
x\in\overline{M}_{+}
\right\}
}.
\label{eq:eqc_intensity_ratio}
\end{equation}

The quantity was treated as undefined if either sampling region was empty or
if the denominator was zero.
Because the complement changes with the candidate mask,
\(R_I\) is a method-specific mask-to-image descriptor rather than an
invariant tissue-contrast measure. Its magnitude was not interpreted as a
direct measure of anatomical inclusion or exclusion
\citep{shehzad2015qap,esteban2017mriqc}.

% ------------------------------------------------------------
\subsubsection{Spacing-Aware Boundary-Gradient Support}
\label{subsubsec:eqc_boundary_gradient}
% ------------------------------------------------------------

A spacing-aware three-dimensional gradient magnitude was calculated from
\(F_{\mathrm{norm}}\),
\begin{equation}
G(x)
=
\left\|
\nabla F_{\mathrm{norm}}(x)
\right\|_2,
\label{eq:eqc_gradient}
\end{equation}
with finite differences evaluated using the corresponding physical voxel
spacing
\citep{kass1988snakes}.
To reduce dependence on voxel anisotropy, the internal mask boundary was
represented as a physical boundary band. Let \(d_M(x)\) denote the
spacing-aware Euclidean distance from an interior mask voxel \(x\in M\) to
the complement of \(M\). The boundary band was
\begin{equation}
\partial_{\rho}M
=
\left\{
x\in M:
d_M(x)\leq\rho
\right\},
\qquad
\rho=s_{\mathrm{eq}}.
\label{eq:eqc_physical_boundary}
\end{equation}

The boundary-gradient support descriptor was
\begin{equation}
B_G(M,F)
=
\frac{
\operatorname{median}_{x\in\partial_{\rho}M}G(x)
}{
P_{75}
\left(
\left\{
G(x):G(x)>0
\right\}
\right)
},
\label{eq:eqc_boundary_gradient}
\end{equation}
provided that the denominator was non-zero.
\(B_G\) describes the amount of local FLAIR-gradient support associated with
the candidate-mask boundary relative to the whole-image gradient
distribution. Strong gradients may occur at valid intracranial boundaries
but may also occur at skull, scalp, sinus, or neck interfaces. Consequently,
\(B_G\) was interpreted as image-edge support rather than as a direct measure
of anatomical boundary accuracy.

% ------------------------------------------------------------
\subsubsection{FLAIR-Derived Foreground Support}
\label{subsubsec:eqc_foreground}
% ------------------------------------------------------------

Peripheral occupation was evaluated relative to a coarse foreground support
obtained independently of the candidate mask from the native FLAIR image.
Let $t_{\mathrm{FG}}$ denote the fifth percentile of positive, finite native
FLAIR intensities. An initial foreground was obtained by thresholding the
native image at $t_{\mathrm{FG}}$; its largest three-dimensional connected
component was retained, and internal holes were filled to obtain $H$
\citep{haralick1987morphology}:
\begin{equation}
H
=
\operatorname{FillHoles}
\left[
\operatorname{LCC}
\left(
\left\{x\in\Omega:
F(x)>t_{\mathrm{FG}},\;
F(x)\ \mathrm{finite}
\right\}
\right)
\right].
\label{eq:eqc_foreground_support}
\end{equation}

The resulting support provides an image-derived reference for peripheral
mask occupation and was not treated as a brain segmentation or anatomical
ground truth. A peripheral band was constructed by subtracting a
three-iteration binary erosion of $H$ from the original support:
\begin{equation}
B_{\mathrm{per}}(H)
=
H\setminus\operatorname{Erode}^{(3)}(H).
\label{eq:eqc_foreground_band}
\end{equation}
Here, $\operatorname{Erode}^{(3)}$ denotes three successive binary-erosion
iterations with a face-connected three-dimensional structuring element on
the native image grid; voxels outside the image were treated as background.
The band therefore follows the operational voxel-space
construction used for the reported peripheral descriptor; its thickness
should not be interpreted as a fixed physical distance in millimetres.
The same foreground construction and erosion settings were applied to all
candidate methods for a given participant.

% ------------------------------------------------------------
\subsubsection{Peripheral Foreground Occupation}
\label{subsubsec:eqc_peripheral}
% ------------------------------------------------------------

Peripheral foreground occupation characterizes the fraction of a candidate
mask that overlaps the outer band of the independently derived FLAIR
foreground support. For a non-empty candidate mask $M$, it was defined as
\begin{equation}
P_{\mathrm{peripheral}}(M,F)
=
\frac{
\left|M\cap B_{\mathrm{per}}(H)\right|
}{
|M|
}.
\label{eq:eqc_peripheral_fraction}
\end{equation}
Thus, $0\leq P_{\mathrm{peripheral}}\leq1$; larger values indicate that
a greater fraction of the candidate mask occupies the peripheral band of
the image-derived foreground. This descriptor may reveal peripheral mask
extensions that are not apparent from global volume or FOV-relative
centroid position. However, voxels lying entirely outside $H$ do not
contribute to its numerator. Consequently, $P_{\mathrm{peripheral}}$
characterizes foreground-band occupation, not total extracranial inclusion
or all possible mask overextensions. As $H$ is an intensity-derived
approximation rather than an anatomical reference, the descriptor was
interpreted descriptively rather than as direct evidence of anatomical
accuracy.

% ------------------------------------------------------------
\subsubsection{Inferior-Tail Fraction}
\label{subsubsec:eqc_tail}
% ------------------------------------------------------------

A complementary descriptor characterized narrow inferior mask extensions
that global volume or centroid measures may not capture.
The candidate mask was evaluated along the physical inferior--superior
direction. Let \(S_k\) denote contiguous physical slabs of thickness
\begin{equation}
\Delta z=s_{\mathrm{eq}}.
\end{equation}

The mask volume within slab \(k\) was
\begin{equation}
V_M(k)
=
v_{\mathrm{vox}}
\sum_{x\in S_k}M(x),
\label{eq:eqc_slab_volume}
\end{equation}
and its cross-sectional-equivalent area was
\begin{equation}
A_M(k)
=
\frac{V_M(k)}{\Delta z}.
\label{eq:eqc_slab_area}
\end{equation}

Let
\begin{equation}
A_{\max}
=
\max_k A_M(k),
\qquad
a(k)
=
\frac{A_M(k)}{A_{\max}}.
\label{eq:eqc_tail_normalized_area}
\end{equation}

Beginning at the inferior end of the occupied mask extent,
\(T_{\mathrm{inf}}\) was defined as the maximal contiguous sequence of slabs
satisfying
\begin{equation}
a(k)\leq\tau,
\qquad
\tau=0.25,
\label{eq:eqc_tail_threshold}
\end{equation}
with accumulation terminating at the first slab for which
\(a(k)>\tau\).
The inferior-tail fraction was
\begin{equation}
P_{\mathrm{tail}}(M)
=
\frac{
\sum_{k\in T_{\mathrm{inf}}}V_M(k)
}{
V_M
}.
\label{eq:eqc_tail_fraction}
\end{equation}

Thus,
\begin{equation}
0\leq P_{\mathrm{tail}}\leq1.
\end{equation}

\(P_{\mathrm{tail}}\) specifically characterizes a narrow
low-cross-sectional-area extension originating from the inferior end of the
mask. It is not intended to detect every form of inferior abnormality;
broad inferior extensions may produce little change in this descriptor.
\(P_{\mathrm{peripheral}}\) provides complementary information when a
mask extension intersects the outer band of the FLAIR-derived foreground,
but neither measure captures every possible form of mask overextension.

Valid brainstem, cerebellar, or other inferior anatomy may also contribute
to the measured tail fraction. \(P_{\mathrm{tail}}\) was therefore
interpreted as a morphological descriptor rather than as a definitive
contamination label.

% ============================================================
\subsection{Pairwise Geometric Characterization}
\label{subsec:eqc_pairwise_geometry}
% ============================================================

Method-specific descriptors do not directly describe how two candidate masks
differ from one another. For each predefined pair \(M_a\) and \(M_b\), E-QC
therefore retained complementary measures of global extent, position, and
principal-axis orientation.

% ------------------------------------------------------------
\subsubsection{Signed Relative Volume Difference}
\label{subsubsec:eqc_relative_volume}
% ------------------------------------------------------------

For physical mask volumes \(V_a>0\) and \(V_b>0\),
\begin{equation}
\delta_V^{(a,b)}
=
\frac{
V_a-V_b
}{
V_a+V_b
}.
\label{eq:eqc_relative_volume}
\end{equation}

The quantity lies in \((-1,1)\) and changes sign when the order of the methods
is reversed,
\begin{equation}
\delta_V^{(b,a)}
=
-\delta_V^{(a,b)}.
\end{equation}

Its magnitude is symmetric and is directly related to volumetric similarity
\citep{taha2015metrics},
\begin{equation}
\mathrm{VS}(a,b)
=
1-
\frac{
|V_a-V_b|
}{
V_a+V_b
},
\qquad
\left|
\delta_V^{(a,b)}
\right|
=
1-\mathrm{VS}(a,b).
\label{eq:eqc_volume_similarity_relation}
\end{equation}

The sign identifies only which candidate has the larger physical volume and
does not indicate anatomical superiority.

% ------------------------------------------------------------
\subsubsection{Physical Centroid Displacement}
\label{subsubsec:eqc_centroid_displacement}
% ------------------------------------------------------------

For candidate masks \(M_a\) and \(M_b\), the signed centroid-difference vector
was
\begin{equation}
\Delta c_{a-b}
=
c_a-c_b.
\label{eq:eqc_centroid_vector}
\end{equation}

Its Euclidean magnitude defines the physical centroid displacement,
\begin{equation}
D_C^{(a,b)}
=
\left\|
c_a-c_b
\right\|_2,
\label{eq:eqc_centroid_mm}
\end{equation}
reported in millimetres
\citep{kim2019loni}.
For compatibility with the native voxel grid, the corresponding
voxel-coordinate displacement was additionally retained as
\begin{equation}
D_C^{(\mathrm{vox})}
=
\left\|
L^{-1}
\left(
c_a-c_b
\right)
\right\|_2.
\label{eq:eqc_centroid_vox}
\end{equation}

Centroid displacement may result from rigid translation, inferior extension,
unilateral inclusion, localized exclusion, or other differences in mask
shape. It was therefore not interpreted as a direct estimate of registration
translation.

% ------------------------------------------------------------
\subsubsection{Principal-Axis Frames and Stability}
\label{subsubsec:eqc_pca_stability}
% ------------------------------------------------------------

For mask \(M_A\), the physical-coordinate covariance matrix was
\begin{equation}
\Sigma_A
=
\frac{1}{|M_A|}
\sum_{x:M_A(x)=1}
\left[
p(x)-c_A
\right]
\left[
p(x)-c_A
\right]^{\mathsf T}.
\label{eq:eqc_covariance}
\end{equation}

When a mask contained more than \(250{,}000\) voxels, deterministic
subsampling was used for computational efficiency.
Let
\begin{equation}
\lambda_{A,1}
\geq
\lambda_{A,2}
\geq
\lambda_{A,3}
\geq0
\end{equation}
denote the ordered eigenvalues of \(\Sigma_A\), with corresponding ordered
unit eigenvectors \(q_{A,1}\), \(q_{A,2}\), and \(q_{A,3}\)
\citep{pearson1901closestfit}.
The principal-axis frame was
\begin{equation}
Q_A
=
\begin{bmatrix}
q_{A,1} &
q_{A,2} &
q_{A,3}
\end{bmatrix}.
\label{eq:eqc_pca_frame}
\end{equation}

If
\begin{equation}
\det(Q_A)<0,
\end{equation}
the sign of the final eigenvector was reversed so that \(Q_A\) formed a
right-handed orthonormal frame.
Because individual PCA axes become unstable when adjacent eigenvalues are
nearly equal, relative eigengaps were calculated as
\citep{davis1970rotation}
\begin{equation}
g_{A,12}
=
\frac{
\lambda_{A,1}-\lambda_{A,2}
}{
\lambda_{A,1}
},
\qquad
g_{A,23}
=
\frac{
\lambda_{A,2}-\lambda_{A,3}
}{
\lambda_{A,2}
}.
\label{eq:eqc_eigengaps}
\end{equation}

For a pair \(M_a,M_b\), joint frame stability was summarized by
\begin{equation}
g_{\min}^{(a,b)}
=
\min
\left(
g_{a,12},
g_{a,23},
g_{b,12},
g_{b,23}
\right).
\label{eq:eqc_joint_eigengap}
\end{equation}

PCA-based orientation descriptors were considered interpretable only when
\begin{equation}
g_{\min}^{(a,b)}
\geq0.02.
\label{eq:eqc_eigengap_threshold}
\end{equation}

When either mask failed this criterion,
\(\theta_{\mathrm{PCA}}\) and \(D_{\mathrm{PCA}}\) were recorded as
unavailable rather than forcing an orientation comparison between unstable
principal-axis frames.

The eigengap threshold was used as an operational numerical-stability
criterion and was not interpreted as an anatomical or diagnostic cutoff.

% ------------------------------------------------------------
\subsubsection{Principal-Axis Orientation Difference}
\label{subsubsec:eqc_pca_orientation}
% ------------------------------------------------------------

Each PCA eigenvector is defined only up to sign. The four
orientation-preserving sign configurations were therefore
\begin{equation}
\mathcal{S}
=
\left\{
s\in\{-1,+1\}^{3}:
s_1s_2s_3=1
\right\}.
\label{eq:eqc_sign_set}
\end{equation}

For each \(s\in\mathcal{S}\), the sign-adjusted candidate frame was
\begin{equation}
Q_a(s)
=
Q_a
\operatorname{diag}(s).
\label{eq:eqc_sign_adjustment}
\end{equation}

The corresponding relative frame matrix was
\begin{equation}
H_s
=
Q_bQ_a(s)^{\mathsf T}.
\label{eq:eqc_relative_frame}
\end{equation}

To remove small numerical deviations from exact orthogonality,
\(H_s\) was projected onto the nearest proper rotation matrix using the
orthogonal Procrustes solution
\citep{schonemann1966procrustes}. If
\begin{equation}
H_s
=
U_s\Sigma_sV_s^{\mathsf T},
\end{equation}
then
\begin{equation}
R_s
=
U_s
\operatorname{diag}
\left(
1,\,
1,\,
\det(U_sV_s^{\mathsf T})
\right)
V_s^{\mathsf T}.
\label{eq:eqc_procrustes_rotation}
\end{equation}

The geodesic rotation angle on \(\mathrm{SO}(3)\) was
\citep{huynh2009rotations}
\begin{equation}
\theta_s
=
\cos^{-1}
\left[
\operatorname{clip}
\left(
\frac{
\operatorname{tr}(R_s)-1
}{2},
-1,
1
\right)
\right].
\label{eq:eqc_geodesic_angle}
\end{equation}

The sign-resolved principal-axis orientation difference was
\begin{equation}
\theta_{\mathrm{PCA}}^{(a,b)}
=
\min_{s\in\mathcal{S}}
\theta_s.
\label{eq:eqc_theta_pca}
\end{equation}

We calculated angles in radians and reported them in degrees.
\(\theta_{\mathrm{PCA}}\) describes disagreement in the dominant geometric
axes of the candidate masks. It does not estimate the actual rotational
parameters of the T$_1$-to-FLAIR registration. Localized inclusion,
asymmetric extraction, inferior extension, dilation, or other mask-shape
differences may alter the PCA frame.

% ------------------------------------------------------------
\subsubsection{PCA-Orientation-Induced Boundary Displacement}
\label{subsubsec:eqc_pca_displacement}
% ------------------------------------------------------------

To express the PCA-frame orientation difference in physical distance units,
the selected sign-resolved rotation was applied to centred boundary
coordinates.

Let \(R_{a\rightarrow b}\) denote the proper rotation corresponding to
\(\theta_{\mathrm{PCA}}^{(a,b)}\). For a boundary point
\(p\in\partial_{\rho}M_a\),
\begin{equation}
d_{a\rightarrow b}(p)
=
\left\|
R_{a\rightarrow b}
\left[
p-c_a
\right]
-
\left[
p-c_a
\right]
\right\|_2.
\label{eq:eqc_directional_pca_displacement}
\end{equation}

The directional summary was
\begin{equation}
D_{a\rightarrow b}
=
\operatorname{median}
\left\{
d_{a\rightarrow b}(p):
p\in\partial_{\rho}M_a
\right\}.
\label{eq:eqc_directional_pca_median}
\end{equation}

When a boundary contained more than \(250{,}000\) points, deterministic
subsampling was used.
The reverse-direction rotation was
\begin{equation}
R_{b\rightarrow a}
=
R_{a\rightarrow b}^{\mathsf T},
\end{equation}
and the symmetric orientation-induced boundary displacement was
\begin{equation}
D_{\mathrm{PCA}}^{(a,b)}
=
\frac{
D_{a\rightarrow b}
+
D_{b\rightarrow a}
}{2}.
\label{eq:eqc_d_pca}
\end{equation}
\(D_{\mathrm{PCA}}\) was reported in millimetres.
This quantity represents the physical boundary-scale consequence of the same
principal-axis orientation difference measured by
\(\theta_{\mathrm{PCA}}\). The two descriptors are therefore related rather
than statistically independent.

\(D_{\mathrm{PCA}}\) is not equivalent to Hausdorff distance, average
symmetric surface distance, surface Dice, point-correspondence displacement,
or actual registration displacement. It quantifies only the boundary
displacement induced by the relative PCA-frame orientation.

%% file: sections/Sec5.tex
\section{Experimental Evaluation}
\label{sec:experimental_evaluation}

% ============================================================
\subsection{Evaluation Design and Prespecified Comparisons}
\label{subsec:evaluation_design}

The experimental evaluation addressed three complementary dimensions of
native-FLAIR brain-mask assessment: method-specific mask characterization,
fixed pairwise inter-method disagreement, and registration-related multimodal
similarity. Independently delineated manual brain masks in native FLAIR space
were not available. Accordingly, method-specific E-QC descriptors and
pairwise overlap, surface, regional, slice-wise, volumetric, and geometric
measures were interpreted as complementary characterizations of mask behaviour
and inter-method agreement rather than as estimates of anatomical accuracy.

We evaluated three native-FLAIR-space masks: the direct BET mask
\(M_F^{\mathrm{BET}}\), the \texttt{fsl\_anat} output
\(M_F^{\mathrm{fsl\_anat}}\), and the proposed
A\textsuperscript{3}PreCore output \(M_F^{\mathrm{A3P}}\).
For \texttt{fsl\_anat}, the skull-stripped T$_1$ image and anatomical mask
were generated during \texttt{fsl\_anat} preprocessing; the subsequent
T$_1$-to-FLAIR registration and mask-transfer operations were study-specific
processing steps.

Two pairwise comparisons were prespecified:
\begin{align}
M_F^{\mathrm{A3P}}
&\quad\text{versus}\quad
M_F^{\mathrm{fsl\_anat}},
\label{eq:comparison_a3p_fsl_anat}
\\
M_F^{\mathrm{A3P}}
&\quad\text{versus}\quad
M_F^{\mathrm{BET}}.
\label{eq:comparison_a3p_bet}
\end{align}

The first comparison characterized disagreement between two workflows that
propagated the same T$_1$-space anatomical source mask. The second compared
A\textsuperscript{3}PreCore with a mask generated directly from native FLAIR.
Comparator assignment was fixed a priori and was not altered according to
E-QC, NMI, overlap, or visual findings.

BET required only a readable native FLAIR image and was attempted in all
\(80\) participants. The \texttt{fsl\_anat} and
A\textsuperscript{3}PreCore registration-based workflows required paired
T$_1$-weighted and FLAIR inputs and were attempted in \(78\) participants.
To preserve participant-level comparability, all primary inter-method
comparisons used the common \(78\)-participant cohort. Participants remained
in the analysis irrespective of subsequent gross-misregistration,
mask-disagreement, or other QC classifications.

All primary mask analyses used the configured dilated mask variants on the
raw, unmasked native FLAIR grid. Before comparison, each mask was binarized
at \(>0.5\), required to contain a non-empty three-dimensional foreground,
and verified for identical image dimensions and affine compatibility with
native FLAIR using an absolute element-wise tolerance of \(10^{-4}\).

% ============================================================
\subsection{Method-Specific E-QC and Visual Registration Assessment}
\label{subsec:intrinsic_visual_evaluation}

The six method-specific E-QC descriptors defined in
Section~\ref{sec:extended_qc} were calculated independently for BET,
\texttt{fsl\_anat}, and A\textsuperscript{3}PreCore. These descriptors
comprised physical mask volume \(V\), FOV-relative centroid position \(C\),
mask-to-complement FLAIR intensity ratio \(R_I\), boundary-gradient support
\(B_G\), inferior peripheral-or-external occupation \(P_{\mathrm{IE}}\), and
inferior-tail fraction \(P_{\mathrm{tail}}\). They were retained separately
rather than combined into a composite score and were interpreted as
complementary characteristics of mask behaviour rather than as validated
measures of anatomical accuracy.

For the \texttt{fsl\_anat} and A\textsuperscript{3}PreCore
registration-based workflows, tri-planar visual assessment examined the
placement of the registered T$_1$ image and propagated mask relative to
native FLAIR anatomy. Evident whole-mask rigid-body displacement was
classified as translation-dominant, rotation-dominant, or combined
translation and rotation.

To establish the \texttt{fsl\_anat} gross-misregistration set, three
neuroscientists independently reviewed all \(78\) \texttt{fsl\_anat}
outputs. Each reviewer independently identified the same \(21\) outputs as
showing gross rigid-body misregistration. Output existence, native-grid
compatibility, and NMI-threshold attainment did not override a visually
confirmed gross registration failure.

The primary categorical registration endpoint was the presence or absence
of gross rigid-body misregistration. E-QC descriptor values were retained
separately from this visually assessed endpoint and were not treated as
equivalent to neuroscientist-confirmed gross misregistration.

% ============================================================
\subsection{Pairwise Inter-Method Mask Agreement}
\label{subsec:pairwise_agreement}

For each prespecified comparison, let
\(M_A=M_F^{\mathrm{A3P}}\) and let \(M_b\) denote the comparator, where
\(b\in\{\mathrm{fsl\_anat},\mathrm{BET}\}\).
Because neither mask constituted anatomical ground truth, all pairwise
quantities were expressed using neutral measures of agreement or
disagreement.

The four pairwise E-QC geometric descriptors defined in
Section~\ref{sec:extended_qc} were also retained:
signed relative volume difference \(\delta_V^{(a,b)}\), physical centroid
displacement \(D_C^{(a,b)}\), principal-axis orientation difference
\(\theta_{\mathrm{PCA}}^{(a,b)}\), and PCA-orientation-induced boundary
displacement \(D_{\mathrm{PCA}}^{(a,b)}\). PCA-based descriptors were
reported only when the predefined eigengap stability criterion was
satisfied.

% ------------------------------------------------------------
\subsubsection{Global Overlap and Exclusive Extent}
\label{subsec:overlap_volume}

Shared voxels were defined by \(M_A\cap M_b\),
A\textsuperscript{3}PreCore-exclusive voxels by
\(M_A\setminus M_b\), comparator-exclusive voxels by
\(M_b\setminus M_A\), and total occupied support by
\(M_A\cup M_b\).

The Dice similarity coefficient and Jaccard index were defined as
\begin{equation}
\operatorname{Dice}(M_A,M_b)
=
\frac{2|M_A\cap M_b|}
{|M_A|+|M_b|},
\label{eq:dice}
\end{equation}
and
\begin{equation}
\operatorname{Jaccard}(M_A,M_b)
=
\frac{|M_A\cap M_b|}
{|M_A\cup M_b|}.
\label{eq:jaccard}
\end{equation}

Directional exclusive extent was quantified as
\begin{equation}
f_{A\setminus b}
=
\frac{|M_A\setminus M_b|}
{|M_A|},
\qquad
f_{b\setminus A}
=
\frac{|M_b\setminus M_A|}
{|M_b|}.
\label{eq:directional_exclusive_fractions}
\end{equation}

These fractions describe the proportion of each candidate mask that is not
shared with the other method. They were not interpreted as false-positive or
false-negative fractions. Exclusive voxel counts were additionally converted
to millilitres using the physical native-FLAIR voxel volume.

% ------------------------------------------------------------
\subsubsection{Mask-Volume Difference}
\label{subsec:volume_difference}

Physical mask volume was calculated using the definition introduced in
Section~\ref{sec:extended_qc}. For each fixed comparison, signed and absolute
volume differences were calculated as
\begin{align}
\Delta V^{\mathrm{A3P}-b}
&=
V(M_A)-V(M_b),
\\
|\Delta V|^{\mathrm{A3P}-b}
&=
\left|
V(M_A)-V(M_b)
\right|.
\end{align}

We additionally expressed these differences relative to the prespecified
comparator:
\begin{align}
\Delta V_{\%,b}^{\mathrm{A3P}-b}
&=
100
\frac{V(M_A)-V(M_b)}
{V(M_b)},
\\
|\Delta V|_{\%,b}^{\mathrm{A3P}-b}
&=
100
\frac{|V(M_A)-V(M_b)|}
{V(M_b)}.
\end{align}

These percentage measures are directional because the fixed comparator
volume appears in the denominator. They are therefore distinct from the
symmetric-denominator E-QC descriptor \(\delta_V^{(a,b)}\) and were retained
as complementary descriptive measures.

% ------------------------------------------------------------
\subsubsection{Surface and Boundary Disagreement}
\label{subsec:surface_disagreement}

Surface-distance analyses used the Euclidean distance transform implemented
in \texttt{scipy.ndimage} (SciPy \(\geq1.11\)). A one-voxel internal surface
was extracted from each binary mask by subtracting a one-step binary erosion
from the original mask. The default three-dimensional erosion structure
provided six-neighbour connectivity; surface points therefore represented
boundary-voxel centres rather than a triangulated mesh.

All mask components were retained, and small disconnected regions were not
removed before surface calculation. Pairwise evaluation required two
non-empty masks; empty outputs were classified as non-evaluable before
surface analysis. A single-voxel component remained a valid surface
component. Native voxel dimensions were supplied to the Euclidean distance
transform through its \texttt{sampling} argument so that distances were
calculated in physical millimetres and anisotropic voxel spacing was handled
directly.

This voxel-centre surface definition was specific to the experimental
surface-distance analysis and was distinct from the spacing-aware physical
boundary construction used by the E-QC boundary-gradient and
PCA-displacement descriptors.

Let \(\mathcal{S}(M_A)\) and \(\mathcal{S}(M_b)\) denote the two
voxel-centre surfaces. Directional nearest-surface distances were calculated
from every point on \(\mathcal{S}(M_A)\) to \(\mathcal{S}(M_b)\) and in the
reverse direction. Let \(D_{A\rightarrow b}\) and \(D_{b\rightarrow A}\)
denote the resulting distance sets and
\(D=D_{A\rightarrow b}\cup D_{b\rightarrow A}\).

The principal surface-disagreement measures were the 95th-percentile
Hausdorff distance
\begin{equation}
\operatorname{HD}_{95}
=
P_{95}(D),
\end{equation}
and the average symmetric surface distance
\begin{equation}
\operatorname{ASSD}
=
\frac{
\operatorname{mean}(D_{A\rightarrow b})
+
\operatorname{mean}(D_{b\rightarrow A})
}{2}.
\end{equation}

Both directional mean distances, the median combined bidirectional distance,
and the maximum Hausdorff distance were also retained. Maximum Hausdorff
distance was treated as exploratory because it can be dominated by an
isolated voxel.

For a physical tolerance \(\tau\), surface Dice was defined as
\begin{equation}
\operatorname{SurfaceDice}_{\tau}
=
\frac{
\left|
\left\{
x\in\mathcal{S}(M_A):
d(x,\mathcal{S}(M_b))\leq\tau
\right\}
\right|
+
\left|
\left\{
y\in\mathcal{S}(M_b):
d(y,\mathcal{S}(M_A))\leq\tau
\right\}
\right|
}{
|\mathcal{S}(M_A)|+|\mathcal{S}(M_b)|
}.
\end{equation}

Surface Dice was calculated at \(1\), \(2\), and \(3\) mm. The \(2\)-mm
tolerance was prespecified as the central value, whereas the \(1\)- and
\(3\)-mm tolerances provided stricter and more permissive sensitivity
descriptors. These measures quantify boundary agreement and do not establish
which boundary is anatomically correct.

% ------------------------------------------------------------
\subsubsection{Regional and Slice-Wise Disagreement}
\label{subsec:regional_slice}

Regional and slice-wise analyses were used to identify localized disagreement
that could be obscured by whole-mask summaries. The voxel axis most closely
aligned with the physical inferior--superior direction was identified from
the native FLAIR NIfTI affine. Native FLAIR slices were then ordered from
inferior to superior and divided across the complete image extent into
inferior, middle, and superior thirds with approximately equal slice counts.

For each region \(G\), the masks were restricted to
\begin{equation}
M_{A,G}=M_A\cap G,
\qquad
M_{b,G}=M_b\cap G.
\end{equation}

Regional Dice was calculated using the same definition as
Equation~\ref{eq:dice}, applied to \(M_{A,G}\) and \(M_{b,G}\).
Regional comparator-relative volume difference was
\begin{equation}
\Delta V_{G,\%,b}^{\mathrm{A3P}-b}
=
100
\frac{
V(M_{A,G})-V(M_{b,G})
}{
V(M_{b,G})
}.
\end{equation}

When the comparator contained no voxels within a region, the relative
volume difference was recorded as unavailable rather than assigning an
arbitrary value. Because the regional partition was based on the complete
field of view rather than a manually defined anatomical extent, these
analyses were considered exploratory. Differences in image cropping,
empty inferior or superior slices, and neck coverage could influence the
regional partition.

Slice-wise Dice was calculated using the same Dice definition,
Equation~\ref{eq:dice}, after restricting the two masks to each
inferior--superior slice. Slices on which both masks were empty were
excluded. A slice on which only one mask was non-empty received a Dice value
of zero.

The primary slice-wise summaries were the median and 10th percentile among
slices with a non-empty mask union. The mean, minimum, number of evaluable
slices, voxel index of the minimum-Dice slice, and its inferior-to-superior
rank were retained as exploratory descriptors. Because each union-positive
slice contributed equally, results from very small inferior or superior
slices were interpreted cautiously.

% ============================================================
\subsection{Controlled Registration-Related Multimodal Similarity}
\label{subsec:controlled_registration_similarity}

The \texttt{fsl\_anat} and A\textsuperscript{3}PreCore transformations were
evaluated under a controlled common-image design. Both subject-specific
transformations were reapplied to the same full bias-corrected T$_1$ image,
\(T\), using native FLAIR as the reference grid and trilinear interpolation:
\begin{align}
T_{\mathrm{fsl\_anat}}^{F}
&=
T\circ
(A_{\mathrm{fsl\_anat}}^{*})^{-1},
\\
T_{\mathrm{A3P}}^{F}
&=
T\circ
(A_{\mathrm{A3P}}^{*})^{-1}.
\end{align}

This design held the evaluation image, interpolation method, reference
geometry, and voxel support constant, thereby avoiding confounding of the
similarity comparison by differences in the image used to evaluate the two
transformations.

The primary evaluation region contained voxels that were finite and non-zero
in both registered full-T$_1$ images and exceeded the fifth percentile of
positive native FLAIR intensities. At least \(10{,}000\) common voxels were
required. Each intensity distribution was clipped to its
\(0.5\)th--\(99.5\)th percentile range, and normalized mutual information
was estimated using \(64\) histogram bins:
\begin{equation}
\operatorname{NMI}(A,B)
=
\frac{H(A)+H(B)}
{H(A,B)}.
\end{equation}

The paired transformation-specific difference was
\begin{equation}
\Delta\operatorname{NMI}
=
\operatorname{NMI}
\left(
T_{\mathrm{A3P}}^{F},F
\right)
-
\operatorname{NMI}
\left(
T_{\mathrm{fsl\_anat}}^{F},F
\right).
\end{equation}

Raw NMI values and participant-level paired differences were reported.
The configured threshold
\(\operatorname{NMI}\geq1.05\) was retained only as a predefined
operational threshold for paired threshold-attainment analysis and was not
interpreted as a validated anatomical pass criterion. Because NMI was closely
related to the registration objectives, higher controlled NMI was interpreted
as optimization-consistent multimodal-similarity evidence rather than as
independent evidence of anatomical correctness.

Sensitivity analyses repeated the controlled calculation using
\(32\), \(64\), and \(128\) histogram bins and four evaluation regions:
common FLAIR foreground, common non-zero support, the full common finite
field of view, and the BET-mask region when available. These analyses
assessed whether the direction of \(\Delta\operatorname{NMI}\) depended
materially on histogram resolution or voxel support.

% ============================================================
\subsection{Method-Blinded Expert Ranking of Brain-Mask Quality}
\label{subsec:blinded_expert_ranking}

A method-blinded expert assessment was performed in a random subset of
40 participants selected from the common 78-participant cohort. For each
participant, the native FLAIR image together with the three corresponding
native-FLAIR-space masks generated by A\textsuperscript{3}PreCore,
\texttt{fsl\_anat}, and BET were provided separately to three
neuroscientists using neutral identifiers that concealed method identity.
The evaluation therefore comprised 120 candidate masks.

Each neuroscientist independently reviewed the masks in FSLeyes. Reviewers
were not informed which mask had been generated by which method and were
free to choose the order in which the three candidate masks were inspected.
For each participant, the three masks were ranked from 1 to 3 according to
overall mask quality and anatomical plausibility, with rank~1 indicating the
preferred mask and rank~3 the least preferred mask.

All three neuroscientists completed their assessments independently. The
rankings from the first two reviewers were used as the primary assessments.
When these reviewers disagreed on the participant-level ordering, the
independently obtained ranking from the third neuroscientist was used for
adjudication. The final participant-level ordering was determined by majority
agreement, such that the ordering supported by at least two of the three
reviewers was retained for analysis.

% ============================================================
\subsection{Statistical Analysis}
\label{subsec:statistical_analysis}

Continuous outcomes were summarized using the number of evaluable
participants, mean, standard deviation, median, interquartile range, minimum,
and maximum. All method-level contrasts used participant-matched
observations.

Gross visual registration failure and controlled
\(\Delta\operatorname{NMI}\) were designated as the principal
registration-strategy outcomes. Method-specific E-QC descriptors were
treated as key secondary outcomes. Pairwise overlap, volumetric, surface,
geometric, regional, and slice-wise measures were used primarily to
characterize inter-method disagreement. Regional, slice-wise,
maximum-Hausdorff, threshold-sensitivity, subgroup, and association analyses
were considered exploratory. Expert-ranking outcomes were summarized
descriptively using rank counts and percentages.

For continuous paired outcomes, we used the two-sided Wilcoxon signed-rank
test with Pratt handling of zero differences. We reported method-specific
values, median and mean paired differences, a \(95\%\) participant-level
bootstrap confidence interval for the median difference, and rank-biserial
correlation. Bootstrap confidence intervals used \(10{,}000\) resamples and
random seed \(20260717\). Positive, negative, and tied paired differences
were additionally reported, with ties defined using a tolerance of
\(10^{-6}\). Where a direction of benefit was prespecified and scientifically
meaningful, these counts were also summarized directionally and evaluated
using an exact two-sided sign test.

Paired binary outcomes, including gross visual registration failure and
configured NMI-threshold attainment, were compared using the exact two-sided
McNemar test. Subgroup summaries stratified by neuroscientist-confirmed
\texttt{fsl\_anat} gross-misregistration status were descriptive and did not
redefine the primary cohort.

The Benjamini--Hochberg false-discovery-rate procedure was applied separately
within prespecified inferential families, including registration-strategy,
method-specific E-QC, and runtime comparisons. Raw \(P\)-values and adjusted
\(q\)-values were reported using a two-sided significance level of
\(\alpha=0.05\). Pairwise overlap, exclusive extent, volume differences,
and surface-distance measures were interpreted primarily as descriptive
measures of disagreement between automatic masks rather than as hypothesis
tests of anatomical superiority.

Exploratory association analyses used Spearman rank correlation because the
E-QC and distance-based measures were potentially skewed and differed in
scale. These analyses examined relationships among method-specific E-QC
descriptors, pairwise geometric descriptors, controlled
\(\Delta\operatorname{NMI}\), overlap, and surface-disagreement measures.
No scalar E-QC score was defined or analyzed. Unavailable PCA-based measures
were excluded pairwise, and all association analyses were interpreted as
descriptive and exploratory.

%% file: sections/sec6.tex
\section{Results}

The final analysis cohort comprised 78 participants with the imaging inputs
required for all workflows evaluated in the present study. The primary
methodological comparison was between A\textsuperscript{3}PreCore and the
\texttt{fsl\_anat},
whereas BET served as a secondary direct native-FLAIR-space brain-masking
comparator. To maintain a common participation basis in the comparative
analyses, all three workflows were evaluated within the same
cohort of 78 participants.

All three workflows were completed successfully in all 78 participants and
produced readable output masks. The generated masks matched the dimensions
and affine geometry of their corresponding native FLAIR images, and no
attempted workflow failed due to missing, unreadable, or
native-grid-incompatible output. Native-grid compatibility was treated as a
technical processing criterion and did not, by itself, establish correct
anatomical placement.

All 78 participants were retained in subsequent analyses irrespective of
later quality-control or registration-status classifications. In particular,
participants subsequently classified as showing gross misregistration with
\texttt{fsl\_anat} were not excluded from the primary between-workflow
comparisons. The overall processing and analysis structure is summarized in
Figure~\ref{fig:analysis_flow}.

% ============================================================
\subsection{Primary Outcome: Gross Rigid-Body Misregistration}
\label{subsec:primary_misregistration}
% ============================================================

\begin{figure}[h]
	\centering
	
	\begin{subfigure}[t]{0.48\textwidth}
		\centering
		\includegraphics[
		width=\linewidth,
		keepaspectratio
		]{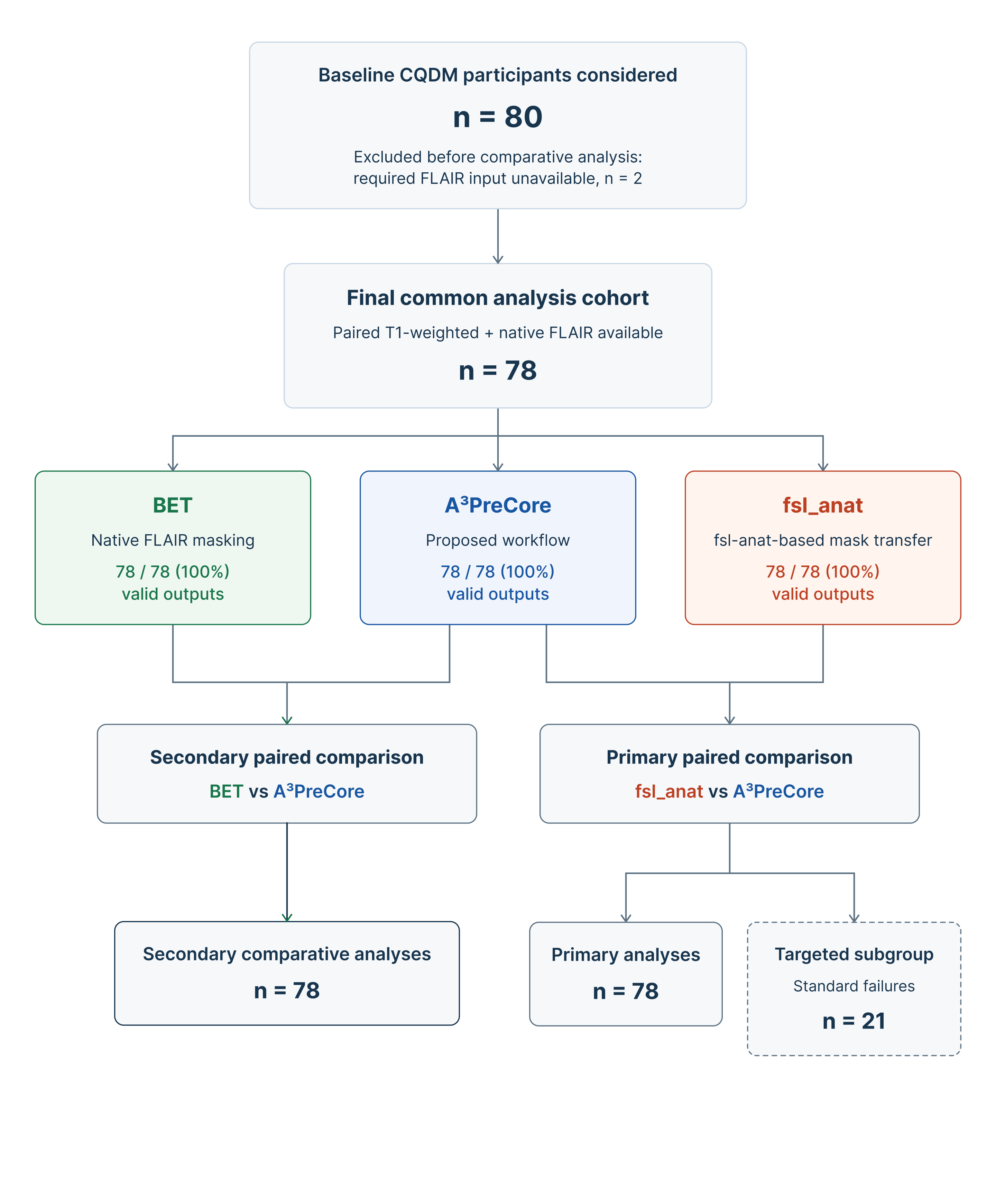}
		\caption{Study cohort and analysis flow.}
		\label{fig:analysis_flow}
	\end{subfigure}
	\hfill
	\begin{subfigure}[t]{0.48\textwidth}
		\centering
		\includegraphics[
		width=\linewidth,
		keepaspectratio
		]{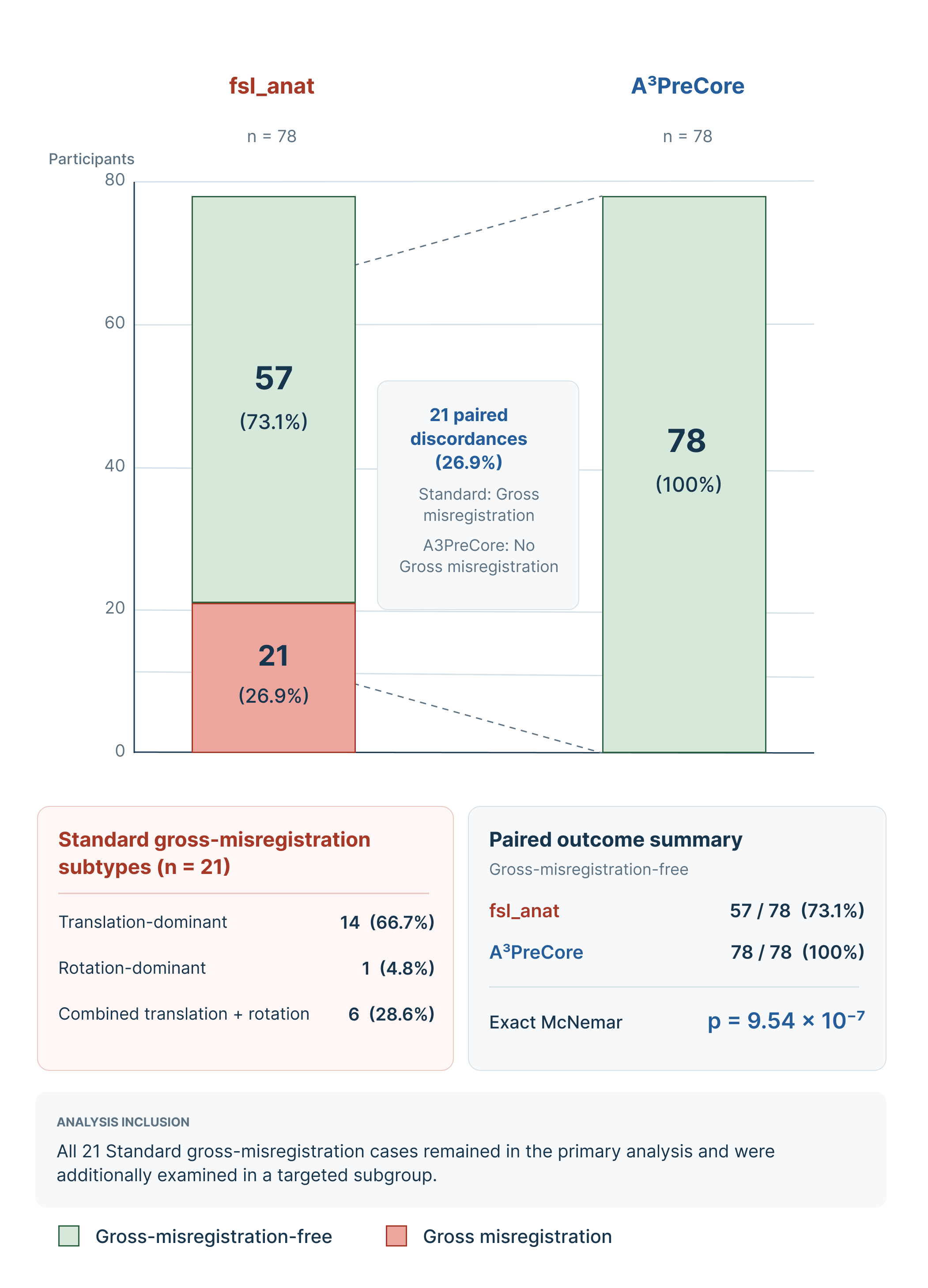}
		\caption{Primary paired gross-misregistration outcome.}
		\label{fig:primary_misregistration}
	\end{subfigure}
	
	\caption{
		\textbf{Study flow and primary paired registration outcome.}
		\textbf{a}, Study cohort, processing completeness, and construction of the
		analysis populations. Of the 80 baseline CQDM participants considered,
		two were excluded from the comparative analysis because the required FLAIR
		input was unavailable, resulting in a final common analysis cohort of
		78 participants. BET, the \texttt{fsl\_anat}, and A\textsuperscript{3}PreCore
		were evaluated in this common cohort, and all three workflows produced
		readable, native-FLAIR-grid-compatible outputs in all 78 participants.
		\textbf{b}, Paired gross rigid-body misregistration outcome for the primary
		comparison between the \texttt{fsl\_anat} and A\textsuperscript{3}PreCore.
		Gross misregistration was identified in 21/78 \texttt{fsl\_anat} outputs (26.9\%)
		and in none of the corresponding A\textsuperscript{3}PreCore outputs
		(0/78, 0.0\%). The 21 paired discordances all consisted of \texttt{fsl\_anat} gross
		misregistration with no corresponding gross misregistration in
		A\textsuperscript{3}PreCore. \texttt{fsl\_anat} failures comprised 14
		translation-dominant, one rotation-dominant, and six combined
		translational and rotational cases. The proportion of
		gross-misregistration-free outputs increased from 57/78 (73.1\%) to
		78/78 (100.0\%), corresponding to an absolute paired increase of
		26.9 percentage points (exact two-sided McNemar test,
		$p = 9.54 \times 10^{-7}$).
	}
	\label{fig:study_flow_primary_outcome}
\end{figure}

The primary registration outcome was the occurrence of gross rigid-body
misregistration of the transferred brain mask in native FLAIR space. Using the participant-level assessment procedure described in the Methods,
gross misregistration was identified in 21 of the 78
\texttt{fsl\_anat} outputs (26.9\%). Independent review by three
neuroscientists was fully concordant: each reviewer independently
identified the same 21 \texttt{fsl\_anat} outputs as showing gross
rigid-body misregistration.  In contrast, none of the corresponding
A\textsuperscript{3}PreCore outputs were classified as showing gross
rigid-body misregistration (0/78, 0.0\%)
(Figure~\ref{fig:study_flow_primary_outcome}\subref{fig:primary_misregistration}).

The 21 \texttt{fsl\_anat} failures comprised distinct rigid-body failure
patterns. Fourteen cases were classified as translation-dominant
(14/78, 17.9\% of the complete cohort; 66.7\% of \texttt{fsl\_anat} failures),
one as rotation-dominant
(1/78, 1.3\% of the complete cohort; 4.8\% of \texttt{fsl\_anat} failures),
and six as showing combined translational and rotational displacement
(6/78, 7.7\% of the complete cohort; 28.6\% of \texttt{fsl\_anat} failures).
The failure subtypes were mutually exclusive and together accounted for all
21 \texttt{fsl\_anat} gross-misregistration cases. The remaining 57 \texttt{fsl\_anat} outputs
(57/78, 73.1\%) showed no evident gross rigid-body misregistration.

The paired comparison consequently showed an increase in the proportion of
gross-misregistration-free outputs from 57/78 (73.1\%) with the \texttt{fsl\_anat}
workflow to 78/78 (100.0\%) with A\textsuperscript{3}PreCore. This
corresponded to an absolute paired increase of 26.9 percentage points.
The binary outcomes were discordant in 21 participants, and all 21
discordances occurred in the same direction: the \texttt{fsl\_anat} output was
classified as grossly misregistered whereas the corresponding
A\textsuperscript{3}PreCore output was not. The paired difference was
significant by the exact two-sided McNemar test
($p = 9.54 \times 10^{-7}$).

Consistent with the paired classification, each of the 21 participants in
the \texttt{fsl\_anat} gross-misregistration subgroup had a corresponding
A\textsuperscript{3}PreCore output that did not exhibit the same gross
rigid-body displacement. This observation is interpreted as a difference in
gross-registration robustness between the two workflows rather than as an
estimate of absolute anatomical accuracy, because no independently delineated
native-FLAIR-space brain-mask reference was available.

The 21 \texttt{fsl\_anat} failure cases were not removed from the primary
quantitative comparisons. Excluding these participants would condition the
analysis on successful \texttt{fsl\_anat} registration and would therefore remove the
principal failure mode under evaluation. All 78 participants were instead
retained in the main between-workflow analyses, while the 21 \texttt{fsl\_anat}
gross-misregistration cases were additionally examined as a targeted subgroup
in subsequent quantitative and qualitative analyses. Because this subgroup
was defined using quantitative mask-disagreement information together with
tri-planar visual assessment, analyses stratified by \texttt{fsl\_anat} registration
status were treated as descriptive characterization rather than as an
independent validation of anatomical accuracy.

% ============================================================
\subsection{Ablation and Registration-Strategy Characterization}
\label{subsec:registration_characterization}
% ============================================================

To characterize the contribution of each component of the
A\textsuperscript{3}PreCore registration strategy to the prevention of
gross rigid-body misregistration, a controlled leave-one-component-out
ablation analysis was performed within the common 78-participant cohort.
The complete strategy combined three modifications relative to the
\texttt{fsl\_anat} reference configuration: registration of the full
bias-corrected T$_1$ image rather than the skull-stripped T$_1$, use of
normalized mutual information (NMI) rather than mutual information (MI),
and extension of the angular search range to $\pm90^{\circ}$. Each
component was removed individually while the remaining two were retained.
Across all configurations, the native FLAIR reference, rigid-body degrees
of freedom, source brain mask, and mask-resampling procedure were held
constant, thereby restricting the comparison to the registration strategy
itself.

\begin{figure}[htbp]
	\centering
	\includegraphics[
	width=0.65\linewidth,
	keepaspectratio
	]{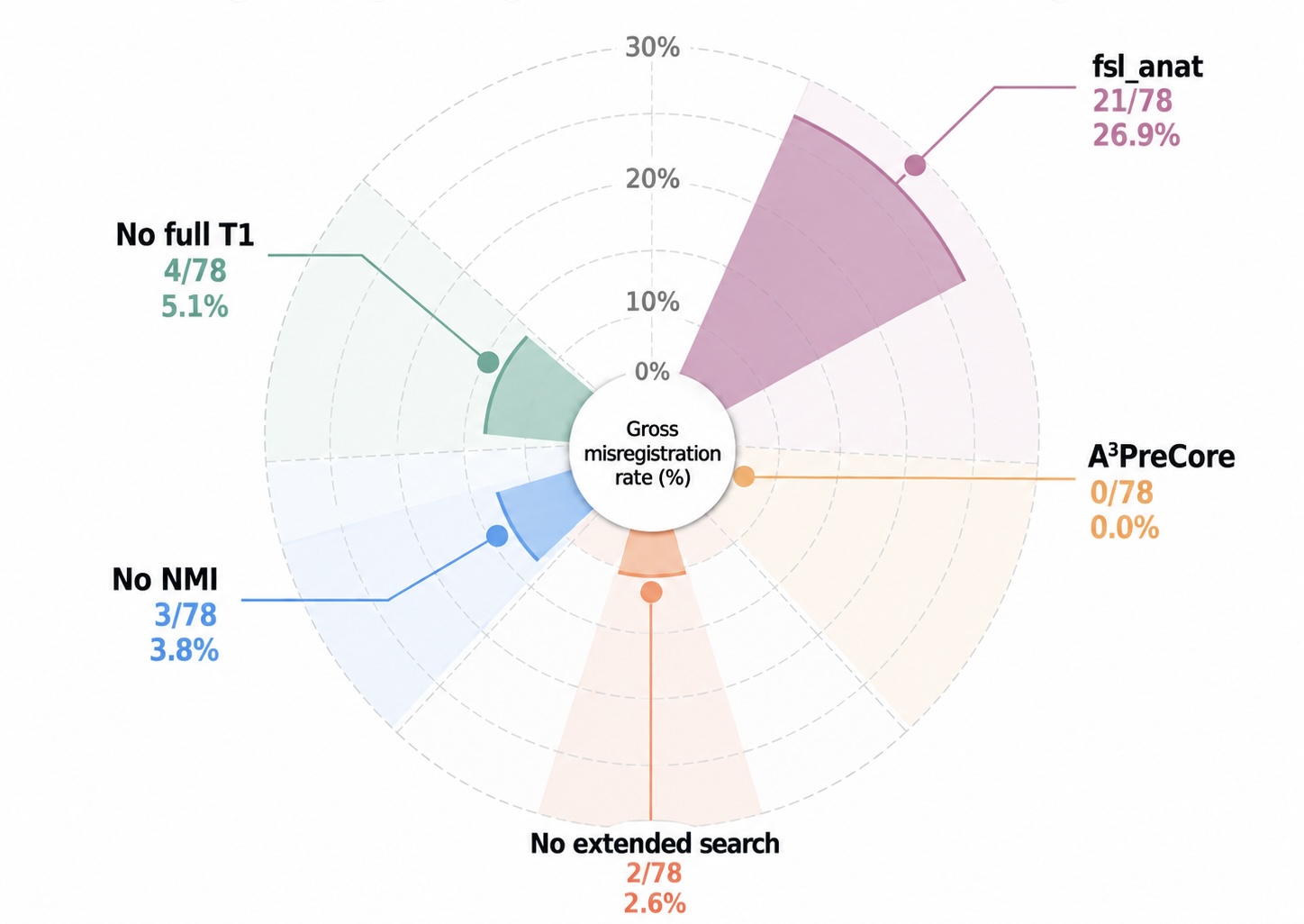}
	
	\caption{
		\textbf{Gross rigid-body misregistration across A\textsuperscript{3}PreCore
			ablation configurations.}
		Rates were 26.9\% for \texttt{fsl\_anat}, 5.1\% for \textit{No full T$_1$},
		3.8\% for \textit{No NMI}, 2.6\% for \textit{No extended search}, and
		0.0\% for the complete A\textsuperscript{3}PreCore configuration, all
		evaluated in the common 78-participant cohort.
	}
	\label{fig:ablation_failure_rate}
\end{figure}

Gross rigid-body misregistration occurred in 21/78 participants (26.9\%)
under the \texttt{fsl\_anat} reference configuration. The gross-failure
rate decreased to 4/78 (5.1\%) in the \textit{No full T$_1$}
configuration, 3/78 (3.8\%) in the \textit{No NMI} configuration, and
2/78 (2.6\%) in the \textit{No extended search} configuration. No gross
misregistration was observed with the complete
A\textsuperscript{3}PreCore configuration (0/78, 0.0\%)
(Table~\ref{tab:ablation_results};
Figure~\ref{fig:ablation_failure_rate}).
Thus, although each two-component configuration substantially reduced the
gross-failure rate relative to \texttt{fsl\_anat}, none reproduced the
failure-free performance of the complete registration strategy.

The paired transition analysis showed that these residual failures arose
from incomplete recovery of the original \texttt{fsl\_anat} failures
rather than from degradation of previously successful registrations.
Among the 21 participants with gross \texttt{fsl\_anat} misregistration,
17/21 were rescued in the \textit{No full T$_1$} configuration, 18/21 in
the \textit{No NMI} configuration, and 19/21 in the
\textit{No extended search} configuration. The complete
A\textsuperscript{3}PreCore configuration rescued all 21/21 baseline
failures. Importantly, none of the ablated configurations introduced a
new gross misregistration among the 57 participants without gross
\texttt{fsl\_anat} misregistration, and the complete configuration
likewise introduced none (0/57 in all cases)
(Table~\ref{tab:ablation_results}).

Paired binary differences relative to \texttt{fsl\_anat} were evaluated
using exact two-sided McNemar tests. The reductions remained significant
in the \textit{No full T$_1$} configuration
($p=1.53\times10^{-5}$), the \textit{No NMI} configuration
($p=7.63\times10^{-6}$), and the \textit{No extended search}
configuration ($p=3.81\times10^{-6}$), with the largest paired
difference observed for the complete A\textsuperscript{3}PreCore
configuration ($p=9.54\times10^{-7}$). These ablation results indicate
that the three registration components have complementary roles:
retaining any two was sufficient to eliminate most baseline failures,
whereas the combination of all three was required to eliminate gross
misregistration throughout the evaluated cohort.

\begin{table}[htbp]
	\centering
	
	\caption{Leave-one-component-out ablation of the
		A\textsuperscript{3}PreCore registration strategy, using gross
		rigid-body misregistration as the primary endpoint. Each ablated
		configuration retained two of the three registration components,
		whereas the complete A\textsuperscript{3}PreCore configuration combined
		full-T$_1$ registration, NMI, and extended angular search.}
	\label{tab:ablation_results}
	
	\small
	\setlength{\tabcolsep}{3pt}
	\renewcommand{\arraystretch}{1.15}
	
	\begin{tabular}{
			p{0.18\linewidth}
			p{0.12\linewidth}
			c
			c
			c
			c
			c
			c
		}
		
		\hline
		
		\textbf{Configuration}
		&
		\textbf{\shortstack{Moving\\T$_1$}}
		&
		\textbf{Cost}
		&
		\textbf{Search}
		&
		\textbf{\shortstack{Gross\\failure}}
		&
		\textbf{Rescued$^{a}$}
		&
		\textbf{\shortstack{New\\failures$^{b}$}}
		&
		\textbf{\shortstack{McNemar\\$p$}}
		\\
		
		\hline
		
		\texttt{fsl\_anat}
		&
		Skull-stripped
		&
		MI
		&
		Default
		&
		21/78 (26.9\%)
		&
		--
		&
		--
		&
		--
		\\
		
		No full T$_1$
		&
		Skull-stripped
		&
		NMI
		&
		$\pm 90^{\circ}$
		&
		4/78 (5.1\%)
		&
		17/21
		&
		0/57
		&
		$1.53 \times 10^{-5}$
		\\
		
		No NMI
		&
		Full
		&
		MI
		&
		$\pm 90^{\circ}$
		&
		3/78 (3.8\%)
		&
		18/21
		&
		0/57
		&
		$7.63 \times 10^{-6}$
		\\
		
		No extended search
		&
		Full
		&
		NMI
		&
		Default
		&
		2/78 (2.6\%)
		&
		19/21
		&
		0/57
		&
		$3.81 \times 10^{-6}$
		\\
		
		A\textsuperscript{3}PreCore
		&
		Full
		&
		NMI
		&
		$\pm 90^{\circ}$
		&
		\textbf{0/78 (0.0\%)}
		&
		\textbf{21/21}
		&
		\textbf{0/57}
		&
		$\mathbf{9.54 \times 10^{-7}}$
		\\
		
		\hline
		
	\end{tabular}
	
	\vspace{2mm}
	
	\begin{minipage}{0.96\linewidth}
		\footnotesize
		MI, mutual information; NMI, normalized mutual information.
		$^{a}$ Number of the 21 participants with gross \texttt{fsl\_anat}
		misregistration who were no longer classified as grossly misregistered
		under the indicated configuration.
		$^{b}$ Number of new gross misregistrations among the 57 participants
		without gross \texttt{fsl\_anat} misregistration.
		McNemar $p$ values correspond to exact two-sided paired comparisons
		against \texttt{fsl\_anat}. All comparisons were performed within the
		same 78-participant cohort.
	\end{minipage}
	
\end{table}

Beyond the binary gross-misregistration endpoint, the resulting
registration transforms were characterized quantitatively using a
controlled NMI analysis. For this analysis, the \texttt{fsl\_anat} and
A\textsuperscript{3}PreCore transforms were reapplied to the same full
bias-corrected T$_1$ image and evaluated against the same native FLAIR
reference using identical preprocessing, histogram binning, and voxel
support. This controlled design removed differences in moving-image content
from the similarity evaluation itself. NMI was therefore interpreted as a
registration-similarity descriptor and not as an independent measure of
anatomical accuracy.

\begin{figure}[htbp]
	\centering
	
	\begin{subfigure}[t]{0.48\textwidth}
		\centering
		\includegraphics[
		width=\linewidth,
		height=6.2cm
		]{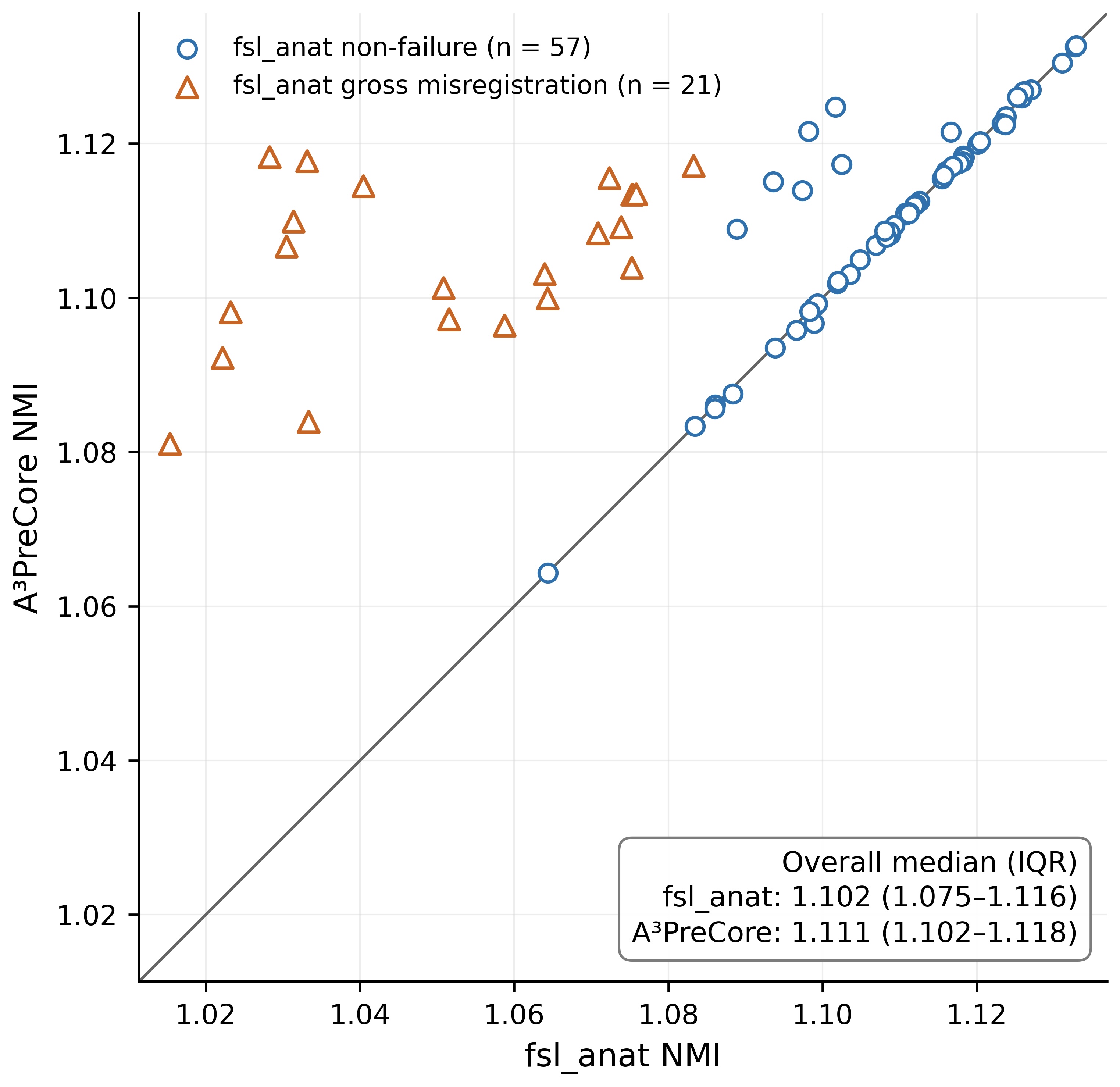}
		\caption{Participant-level controlled NMI.}
		\label{fig:nmi_scatter}
	\end{subfigure}
	\hfill
	\begin{subfigure}[t]{0.48\textwidth}
		\centering
		\includegraphics[
		width=\linewidth,
		height=6.2cm
		]{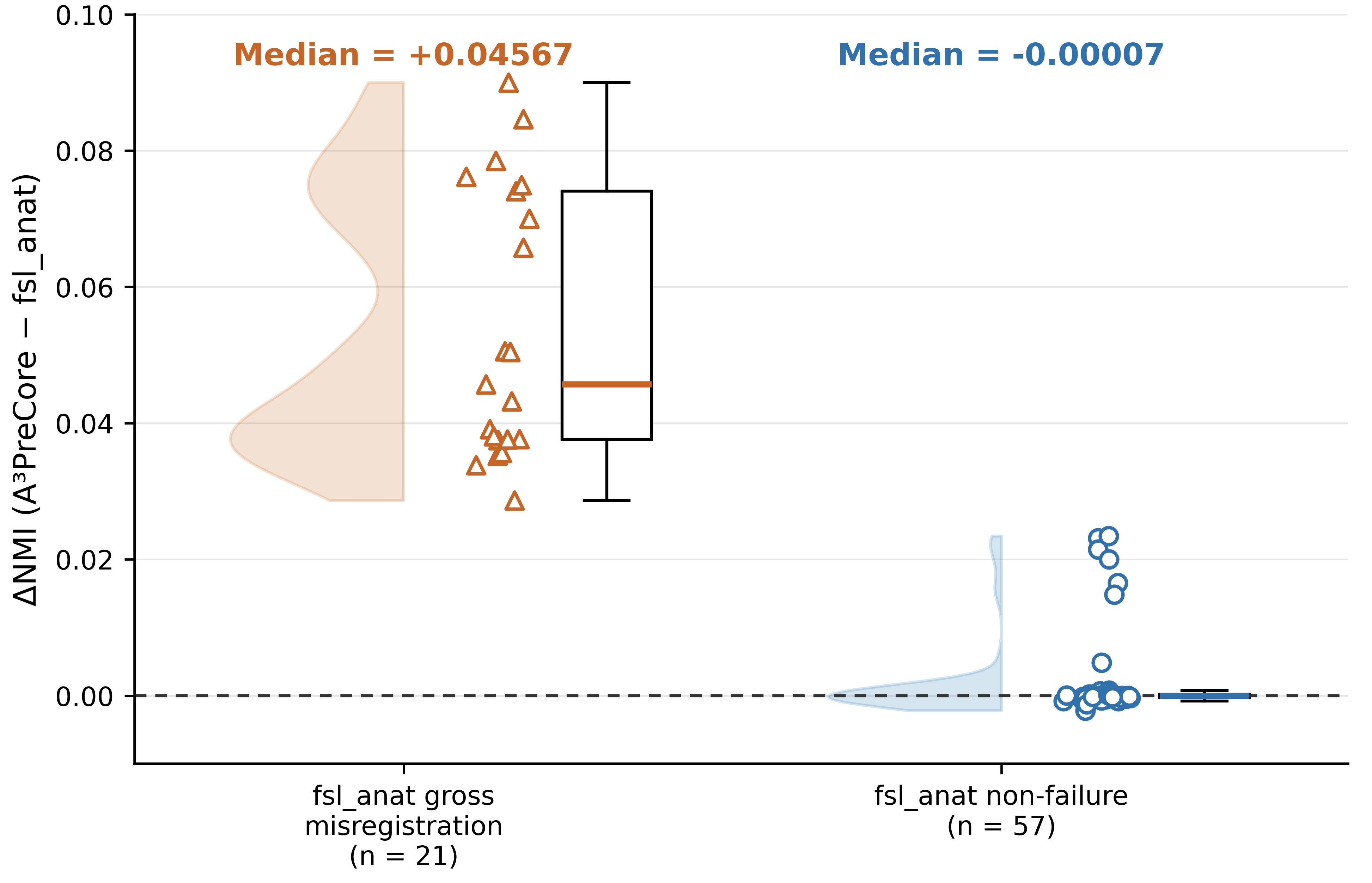}
		\caption{Paired NMI difference by \texttt{fsl\_anat}
			gross-misregistration status.}
		\label{fig:nmi_delta_status}
	\end{subfigure}
	\caption{
		\textbf{Controlled NMI characterization of \texttt{fsl\_anat} and
			A\textsuperscript{3}PreCore registration transforms.}
		\textbf{a}, Participant-level NMI after the \texttt{fsl\_anat} and
		A\textsuperscript{3}PreCore transforms were reapplied to the same full
		bias-corrected T$_1$ image and evaluated against the same native FLAIR
		reference. The identity line denotes equal NMI between the two transforms.
		\textbf{b}, Participant-level paired NMI difference,
		$\Delta\operatorname{NMI}
		=
		\operatorname{NMI}_{\mathrm{A3P}}
		-
		\operatorname{NMI}_{\mathrm{fsl\_anat}}$,
		stratified by \texttt{fsl\_anat} gross-misregistration status.
		Positive differences were concentrated in participants with gross
		\texttt{fsl\_anat} misregistration, whereas paired differences remained
		centred near zero among participants without gross
		\texttt{fsl\_anat} misregistration. NMI is interpreted as a
		registration-similarity descriptor rather than as independent anatomical
		validation.
	}
	\label{fig:controlled_nmi}
\end{figure}

Across all 78 participants, the median NMI was 1.111
(IQR, 1.102--1.118) for A\textsuperscript{3}PreCore and 1.102
(IQR, 1.075--1.116) for \texttt{fsl\_anat}
(Table~\ref{tab:controlled_nmi};
Figure~\ref{fig:controlled_nmi}\subref{fig:nmi_scatter}).
The median participant-level paired difference, calculated as
A\textsuperscript{3}PreCore minus \texttt{fsl\_anat}, was 0.00006
(IQR, $-0.00017$ to 0.03255), with a bootstrap 95\% confidence interval
of $-0.00004$ to 0.00059. The paired distributions differed by the
Wilcoxon signed-rank test
($p=0.0022$; rank-biserial effect size $=0.399$).

The direction of the participant-level difference, however, was not
uniform across the complete cohort. A\textsuperscript{3}PreCore had the
higher NMI in 44 participants and \texttt{fsl\_anat} in 34, with no ties
(exact sign test, $p=0.308$). Thus, the statistically detectable shift in
the paired distribution reflected the magnitude of positive differences in
a subset of participants rather than a consistent increase in NMI across
all participants. In particular, the difference between the marginal
median NMI values of the two workflows should not be interpreted as the
median paired difference.

The positive NMI differences were concentrated in participants with gross
\texttt{fsl\_anat} misregistration
(Figure~\ref{fig:controlled_nmi}\subref{fig:nmi_delta_status}).
Within this 21-participant subgroup, median NMI was 1.108
(IQR, 1.098--1.114) for A\textsuperscript{3}PreCore and 1.052
(IQR, 1.031--1.072) for \texttt{fsl\_anat}. The median paired difference was
0.04567 (IQR, 0.03763--0.07408), with a bootstrap 95\% confidence
interval of 0.03769--0.07003. All 21 paired differences were positive,
with a rank-biserial effect size of 1.0 and a Wilcoxon signed-rank
$p<0.001$.

In contrast, among the 57 participants without gross \texttt{fsl\_anat}
misregistration, the two transforms showed closely matched NMI values.
Median NMI was 1.111 (IQR, 1.104--1.118) for
A\textsuperscript{3}PreCore and 1.109
(IQR, 1.099--1.118) for \texttt{fsl\_anat}. The median paired difference was
$-0.00007$ (IQR, $-0.00025$ to 0.00017), with a bootstrap 95\%
confidence interval of $-0.00016$ to 0.00003. The paired difference was
not significant ($p=0.429$; rank-biserial effect size $=-0.120$).
These results indicate that the NMI distributions were essentially
preserved when \texttt{fsl\_anat} registration did not exhibit gross failure.

Using the prespecified NMI threshold of 1.05, all 78
A\textsuperscript{3}PreCore transforms exceeded the threshold, compared
with 69/78 \texttt{fsl\_anat} transforms. Nine participants were discordantly
classified, and all nine discordances occurred in the same direction:
A\textsuperscript{3}PreCore exceeded the threshold whereas \texttt{fsl\_anat} did
not (exact two-sided McNemar test, $p=0.0039$). All nine discordant cases
belonged to the \texttt{fsl\_anat} gross-misregistration subgroup. By contrast,
both transforms exceeded the threshold in all 57 participants without
gross \texttt{fsl\_anat} misregistration.

Sensitivity analyses across alternative voxel-support definitions and
histogram-bin configurations yielded the same principal pattern: all
21 \texttt{fsl\_anat} gross-misregistration cases retained positive paired NMI
differences across the evaluated configurations, whereas median differences
among \texttt{fsl\_anat} non-failure cases remained close to zero
(Supplementary Figure~\ref{fig:nmi_sensitivity}).

\begin{sidewaystable}[p]
	\caption{
		Controlled NMI characterization of \texttt{fsl\_anat} and
		A\textsuperscript{3}PreCore registration transforms.
	}
	\label{tab:controlled_nmi}
	
	\centering
	\small
	\setlength{\tabcolsep}{5pt}
	\renewcommand{\arraystretch}{1.15}
	
	% ============================================================
	% A. Continuous NMI analysis
	% ============================================================
	
	\begin{tabular*}{0.98\linewidth}{
			@{\extracolsep{\fill}}
			lcccccc
			@{}
		}
		\hline
		
		\multicolumn{7}{l}{\textit{A. Continuous NMI analysis}} \\
		
		\hline
		
		&
		\multicolumn{2}{c}{NMI}
		&
		\multicolumn{2}{c}{Paired $\Delta$NMI}
		&
		&
		\\
		
		\cline{2-3}
		\cline{4-5}
		
		Group ($n$)
		& A\textsuperscript{3}PreCore
		& \texttt{fsl\_anat}
		& Median (IQR)
		& 95\% CI
		& $r_{\mathrm{rb}}$
		& $p$ \\
		
		\hline
		
		All participants (78)
		& 1.111 (1.102--1.118)
		& 1.102 (1.075--1.116)
		& 0.00006 ($-0.00017$--0.03255)
		& $-0.00004$, 0.00059
		& 0.399
		& 0.0022 \\
		
		\texttt{fsl\_anat} gross misregistration (21)
		& 1.108 (1.098--1.114)
		& 1.052 (1.031--1.072)
		& 0.04567 (0.03763--0.07408)
		& 0.03769, 0.07003
		& 1.000
		& $<0.001$ \\
		
		\texttt{fsl\_anat} no gross misregistration (57)
		& 1.111 (1.104--1.118)
		& 1.109 (1.099--1.118)
		& $-0.00007$ ($-0.00025$--0.00017)
		& $-0.00016$, 0.00003
		& $-0.120$
		& 0.429 \\
		
		\hline
	\end{tabular*}
	
	\vspace{7pt}
	
	% ============================================================
	% B. NMI threshold analysis
	% ============================================================
	
	\begin{tabular*}{\textwidth}{
			@{\extracolsep{\fill}}
			lcccc
			@{}
		}
		\hline
		
		\multicolumn{5}{l}{
			\textit{B. NMI threshold analysis ($>1.05$)}
		} \\
		
		\hline
		
		Group ($n$)
		& A\textsuperscript{3}PreCore
		& \texttt{fsl\_anat}
		& Discordant pairs
		& McNemar $p$ \\
		
		\hline
		
		All participants (78)
		& 78/78 (100\%)
		& 69/78 (88.5\%)
		& 9
		& 0.0039 \\
		
		\texttt{fsl\_anat} gross misregistration (21)
		& 21/21 (100\%)
		& 12/21 (57.1\%)
		& 9
		& 0.0039 \\
		
		\texttt{fsl\_anat} no gross misregistration (57)
		& 57/57 (100\%)
		& 57/57 (100\%)
		& 0
		& 1.000 \\
		
		\hline
	\end{tabular*}
	
	\par\smallskip
	\footnotesize
	Continuous values are median (IQR).
	$\Delta$NMI was calculated participant-wise as
	A\textsuperscript{3}PreCore minus \texttt{fsl\_anat}.
	$r_{\mathrm{rb}}$, rank-biserial effect size.
	Confidence intervals for the median paired difference were obtained using
	bootstrap resampling. Continuous paired comparisons used two-sided
	Wilcoxon signed-rank tests; threshold classifications were compared using
	exact two-sided McNemar tests. NMI is interpreted as a
	registration-similarity measure rather than as independent anatomical
	validation.
\end{sidewaystable}

Taken together, the registration-strategy analyses indicate that the
between-workflow differences were concentrated primarily in the subset of
participants in whom \texttt{fsl\_anat} exhibited gross rigid-body
misregistration. In the remaining participants, the \texttt{fsl\_anat} and
A\textsuperscript{3}PreCore transforms showed closely matched controlled
NMI values. The NMI analysis therefore supports the primary
gross-misregistration findings as a mechanistic characterization of the
registration behaviour, while the spatial consequences of these differences
on the transferred brain masks are examined in the following section using
volumetric overlap, boundary-distance, and geometric measures.

% ============================================================
\subsection{Spatial Agreement and Boundary Concordance}
\label{subsec:spatial_agreement}
% ============================================================

Pairwise spatial agreement was examined to determine whether the increased
registration robustness of A\textsuperscript{3}PreCore was accompanied by
substantial changes in the transferred brain mask when the \texttt{fsl\_anat}
did not exhibit gross rigid-body misregistration. Dice and Jaccard
coefficients were used to characterize volumetric overlap, whereas the
95th-percentile Hausdorff distance (HD95) and average symmetric surface
distance (ASSD) were used to characterize boundary disagreement. Directional
coverage and non-overlap measures were additionally used to describe the
spatial distribution of between-method differences. Because no independently
delineated native-FLAIR brain-mask reference was available, all such measures
were interpreted as measures of inter-method agreement rather than anatomical
accuracy.

Across the complete 78-participant cohort, the median Dice coefficient
between A\textsuperscript{3}PreCore and \texttt{fsl\_anat} was 0.990
(IQR, 0.932--0.994), and the corresponding median Jaccard index was 0.981
(IQR, 0.872--0.987). The median normalized symmetric difference was 0.019
(IQR, 0.013--0.128), while the median HD95 and ASSD were 0.777~mm
(IQR, 0.693--5.423~mm) and 0.274~mm
(IQR, 0.186--2.272~mm), respectively
(Tables~\ref{tab:pairwise_agreement} and
\ref{tab:boundary_surface_disagreement}).

The overall agreement distribution was strongly dependent on \texttt{fsl\_anat}
registration status. Among the 57 participants in whom the \texttt{fsl\_anat}
showed no gross rigid-body misregistration, the median Dice coefficient was
0.993 (IQR, 0.989--0.995), and the median Jaccard index was 0.985
(IQR, 0.978--0.989). The median fraction of the \texttt{fsl\_anat} mask covered by
A\textsuperscript{3}PreCore was 0.992
(IQR, 0.989--0.995), while the median fraction of the
A\textsuperscript{3}PreCore mask contained within \texttt{fsl\_anat} was 0.993
(IQR, 0.989--0.995). Directional non-overlap was small and nearly symmetric,
with median A\textsuperscript{3}PreCore-only and \texttt{fsl\_anat}-only volumes of
11.5~mL (IQR, 8.0--15.8~mL) and 11.6~mL
(IQR, 8.2--16.5~mL), respectively.

\begin{figure}[h]
	\centering
	\includegraphics[
	width=0.90\linewidth
	]{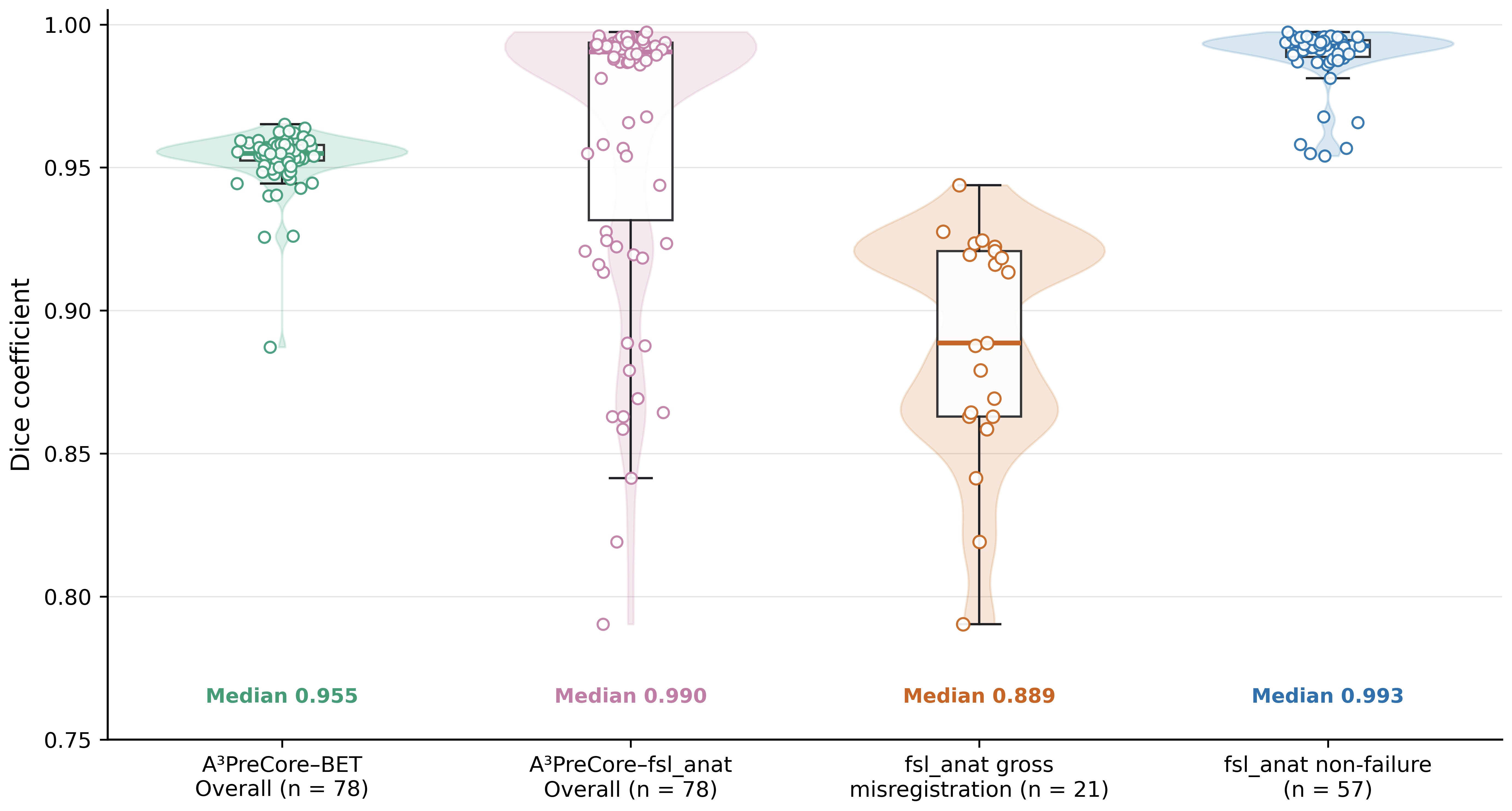}
	
	\caption{
		\textbf{Pairwise Dice agreement according to comparator and \texttt{fsl\_anat}
			registration status.}
		Participant-level Dice coefficients are shown for
		A\textsuperscript{3}PreCore--BET across the complete cohort and for
		A\textsuperscript{3}PreCore--\texttt{fsl\_anat} across the complete cohort and
		separately among participants with and without gross \texttt{fsl\_anat}
		misregistration. Agreement remained high when the \texttt{fsl\_anat}
		showed no gross misregistration, whereas substantially lower overlap was
		concentrated in the \texttt{fsl\_anat} gross-misregistration subgroup.
	}
	\label{fig:pairwise_dice_agreement}
\end{figure}

In contrast, substantially greater spatial disagreement was observed among
the 21 participants classified as having gross \texttt{fsl\_anat}
misregistration. In this subgroup, the median Dice coefficient decreased to
0.889 (IQR, 0.863--0.921), and the median Jaccard index decreased to 0.800
(IQR, 0.759--0.853). The median fraction of the \texttt{fsl\_anat} mask covered by
A\textsuperscript{3}PreCore was 0.884
(IQR, 0.861--0.921), while the median
A\textsuperscript{3}PreCore-contained fraction was 0.894
(IQR, 0.867--0.921). Median directional non-overlap volumes increased to
158.4~mL (IQR, 113.8--180.6~mL) for
A\textsuperscript{3}PreCore-only voxels and 170.8~mL
(IQR, 117.0--188.8~mL) for \texttt{fsl\_anat}-only voxels
(Table~\ref{tab:pairwise_agreement}). Participant-level Dice distributions
for the complete cohort and the \texttt{fsl\_anat}-status subgroups are shown in
Figure~\ref{fig:pairwise_dice_agreement}.

The reduction in overlap within the \texttt{fsl\_anat} gross-misregistration subgroup
does not, by itself, establish which of the two transferred masks was
anatomically more accurate. The direction of the observed disagreement was
therefore interpreted in conjunction with the prespecified participant-level
gross-misregistration assessment described in
Section~\ref{subsec:primary_misregistration}. All 21 participants in this
subgroup were classified as showing gross rigid-body misregistration in the
\texttt{fsl\_anat} output, whereas none of the corresponding
A\textsuperscript{3}PreCore outputs showed the same gross displacement.
Accordingly, the subgroup-specific overlap and non-overlap results were
treated as descriptive characterization of the spatial consequences of the
identified \texttt{fsl\_anat} registration failures rather than as independent
validation of anatomical accuracy.

\begin{table}[htbp]
	
	\caption{
		Pairwise spatial agreement of A\textsuperscript{3}PreCore with BET and the
		\texttt{fsl\_anat}.
	}
	\label{tab:pairwise_agreement}
	
	\centering
	\footnotesize
	\setlength{\tabcolsep}{2pt}
	
	\begin{tabular*}{0.98\linewidth}{
			@{\extracolsep{\fill}}
			lcccc
			@{}
		}
		\hline
		
		&
		\multicolumn{1}{c}{A\textsuperscript{3}PreCore--BET}
		&
		\multicolumn{3}{c}{A\textsuperscript{3}PreCore--\texttt{fsl\_anat}}
		\\
		
		\cline{2-2}
		\cline{3-5}
		
		Metric
		& Overall
		& Overall
		& Gross misregistration
		& No gross misregistration
		\\
		
		& ($n=78$)
		& ($n=78$)
		& ($n=21$)
		& ($n=57$)
		\\
		
		\hline
		
		Dice
		& 0.955 [0.95--0.96]
		& 0.990 [0.93--0.99]
		& 0.889 [0.86--0.92]
		& 0.993 [0.98--0.99]
		\\
		
		Jaccard
		& 0.914 [0.90--0.91]
		& 0.981 [0.87--0.99]
		& 0.800 [0.76--0.85]
		& 0.985 [0.98--0.99]
		\\
		
		Comparator coverage
		& 0.948 [0.94--0.96]
		& 0.990 [0.93--0.99]
		& 0.884 [0.86--0.92]
		& 0.992 [0.98--0.99]
		\\
		
		A\textsuperscript{3}PreCore-contained fraction
		& 0.965 [0.95--0.97]
		& 0.990 [0.93--0.99]
		& 0.894 [0.87--0.92]
		& 0.993 [0.98--0.99]
		\\
		
		A\textsuperscript{3}PreCore-only volume (mL)
		& 52.8 [36.9--71.1]
		& 13.8 [9.7--99.4]
		& 158.4 [113.8--180.6]
		& 11.5 [8.0--15.8]
		\\
		
		Comparator-only volume (mL)
		& 80.7 [64.0--91.7]
		& 14.3 [9.9--100.5]
		& 170.8 [117.0--188.8]
		& 11.6 [8.2--16.5]
		\\
		
		\hline
	\end{tabular*}
	
	\par\smallskip
	\footnotesize
	
	Values are median [interquartile range].
	A\textsuperscript{3}PreCore--BET was evaluated in the complete paired
	cohort only. The gross-misregistration and no-gross-misregistration
	subgroups apply exclusively to the A\textsuperscript{3}PreCore--\texttt{fsl\_anat}
	comparison.
	
	For each comparison, the comparator was BET or \texttt{fsl\_anat}, respectively.
	Comparator coverage was defined as $|A\cap B|/|B|$, and the
	A\textsuperscript{3}PreCore-contained fraction as $|A\cap B|/|A|$,
	where $A$ denotes A\textsuperscript{3}PreCore and $B$ denotes the
	corresponding comparator. These directional measures quantify inter-method
	agreement and do not use an anatomical ground-truth reference.
	
\end{table}

As a secondary registration-independent comparison,
A\textsuperscript{3}PreCore was also compared with BET, which generated its
brain mask directly in native FLAIR space. Across the complete
78-participant cohort, the median Dice coefficient between
A\textsuperscript{3}PreCore and BET was 0.955
(IQR, 0.952--0.958), and the median Jaccard index was 0.914
(IQR, 0.909--0.919). The median fraction of the BET mask covered by
A\textsuperscript{3}PreCore was 0.948
(IQR, 0.938--0.957), while the median
A\textsuperscript{3}PreCore-contained fraction was 0.965
(IQR, 0.953--0.973). Median directional non-overlap volumes were
52.8~mL (IQR, 36.9--71.1~mL) for
A\textsuperscript{3}PreCore-only voxels and 80.7~mL
(IQR, 64.0--91.7~mL) for BET-only voxels
(Table~\ref{tab:pairwise_agreement};
Figure~\ref{fig:pairwise_dice_agreement}). These results were used to
characterize agreement with an independently generated native-FLAIR mask and
not to establish anatomical accuracy.

\begin{figure}[h]
	\centering
	\begin{subfigure}[t]{0.48\textwidth}
		\centering
		\includegraphics[width=\linewidth]
		{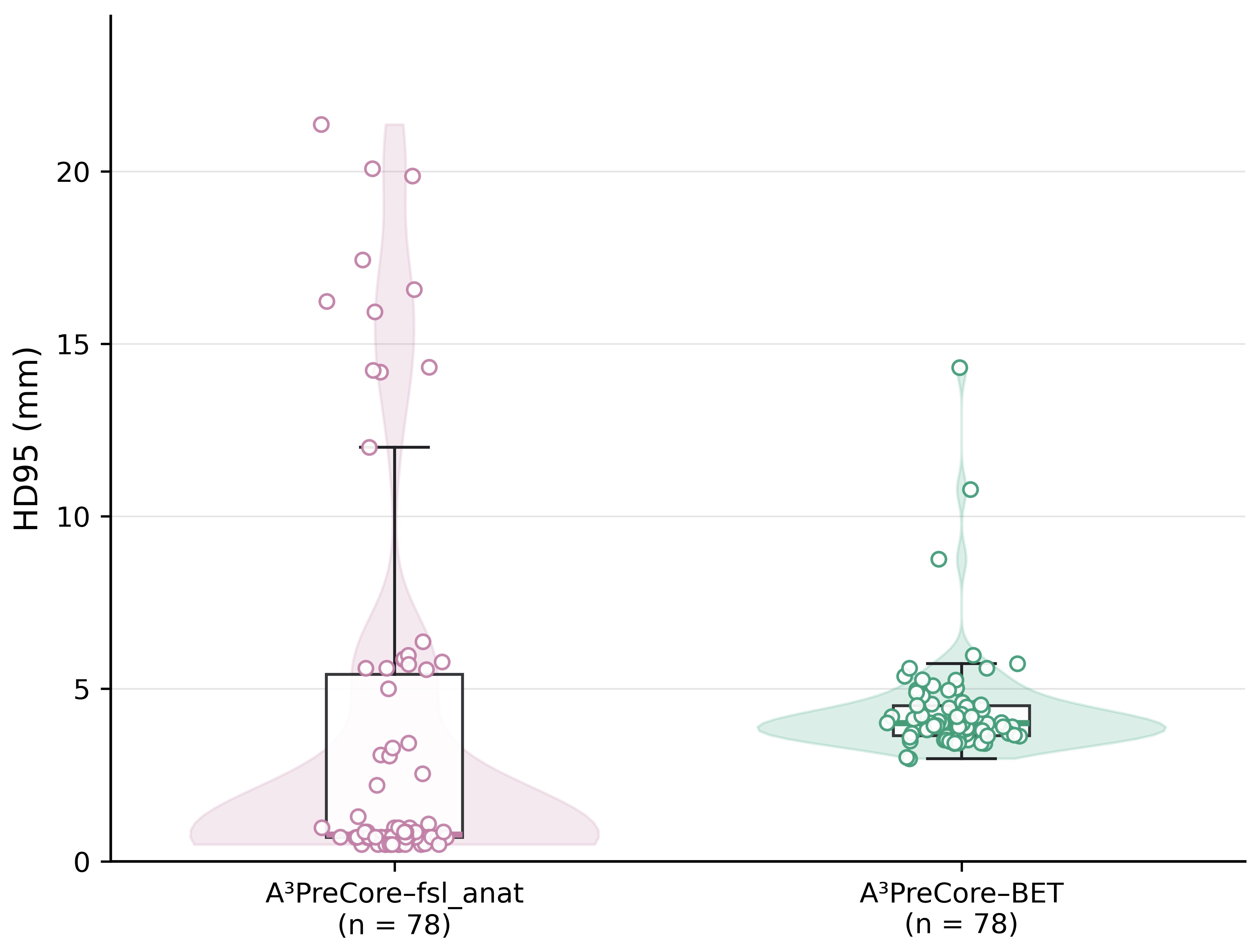}
		\caption{HD95.}
		\label{fig:agreement_hd95}
	\end{subfigure}
	\hfill
	\begin{subfigure}[t]{0.48\textwidth}
		\centering
		\includegraphics[width=\linewidth]
		{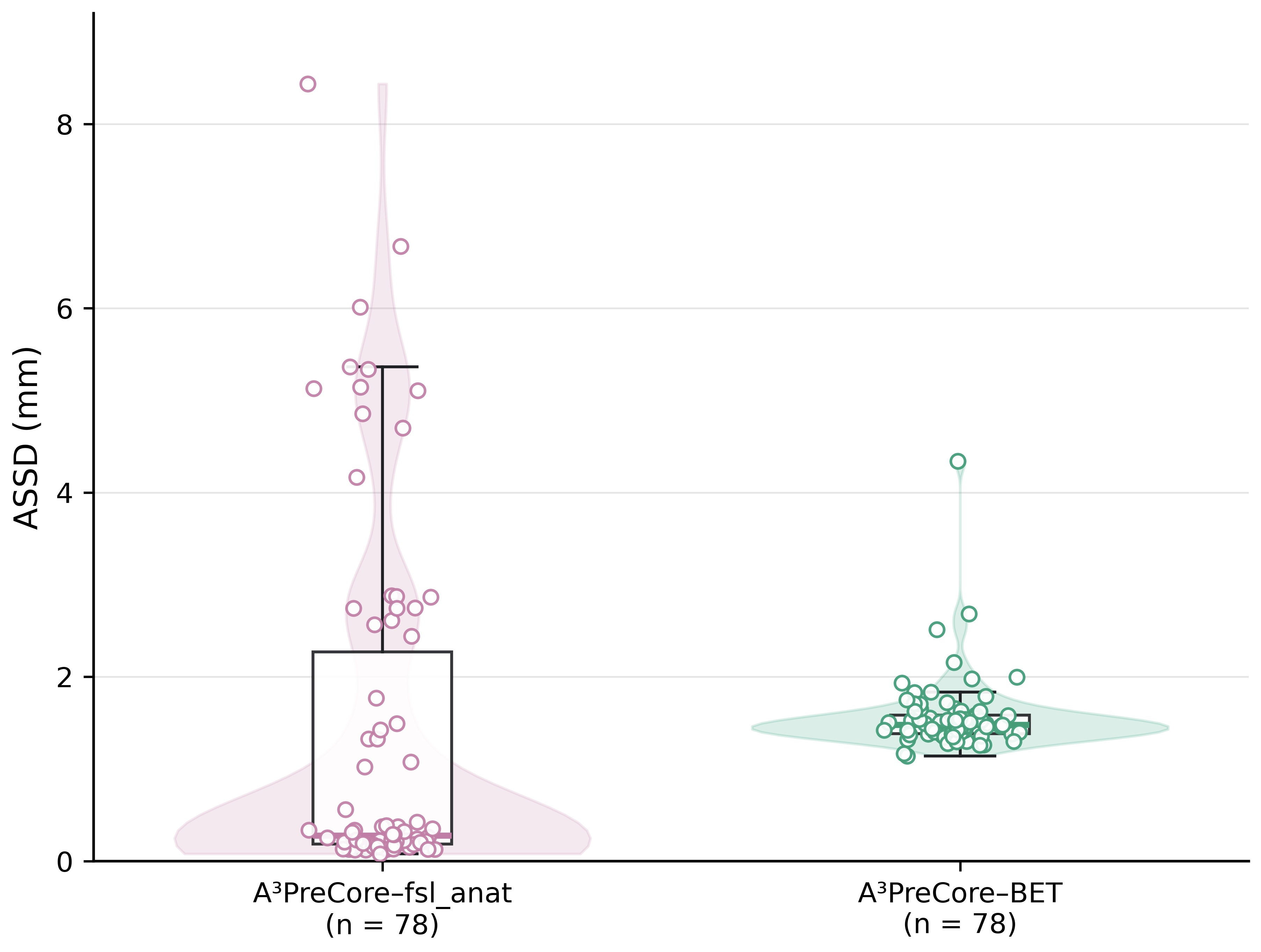}
		\caption{Average symmetric surface distance.}
		\label{fig:agreement_assd}
	\end{subfigure}
	
	\vspace{0.3em}
	
	\begin{subfigure}[t]{0.48\textwidth}
		\centering
		\includegraphics[width=\linewidth]
		{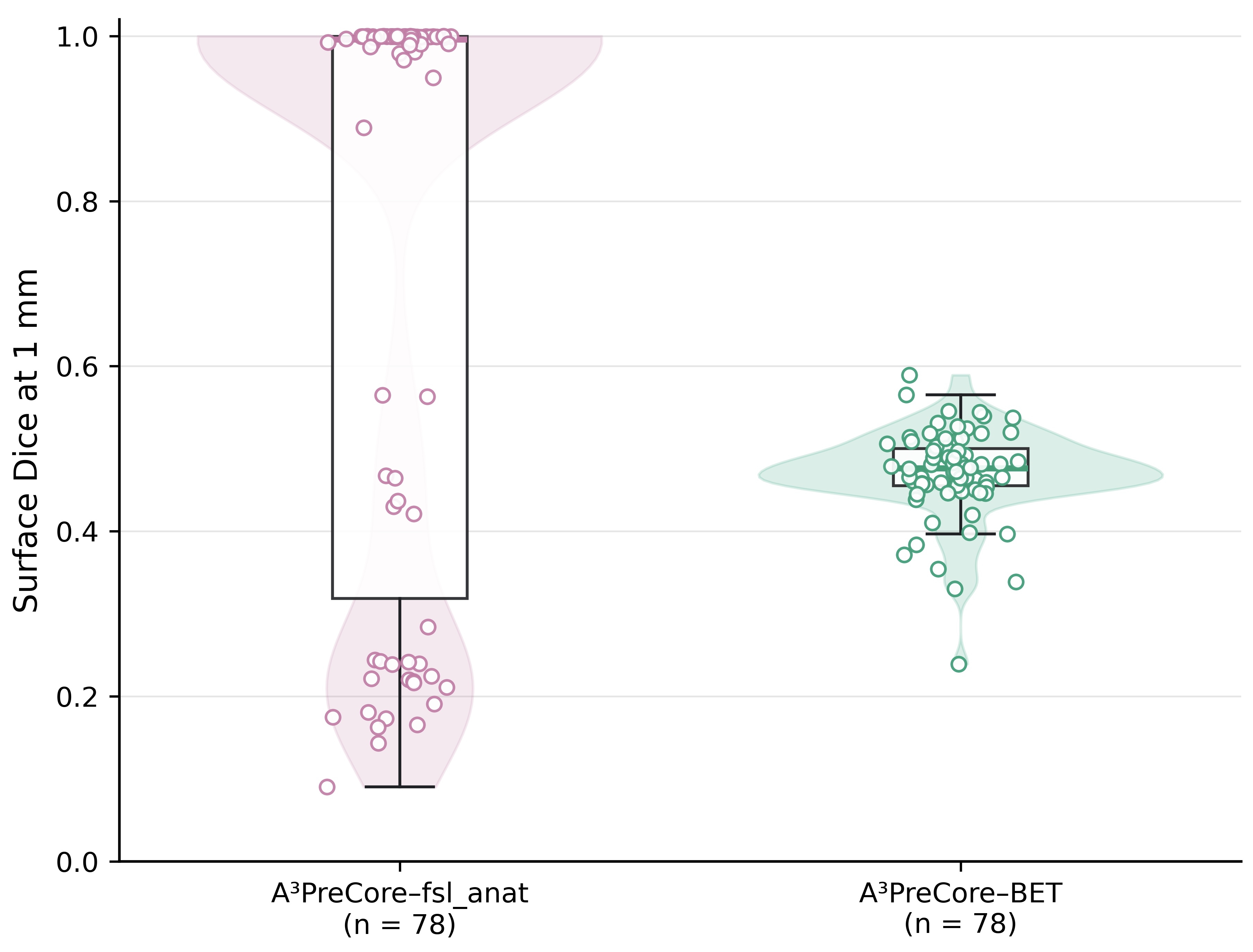}
		\caption{Surface Dice at 1-mm tolerance.}
		\label{fig:agreement_surface_dice}
	\end{subfigure}
	\hfill
	\begin{subfigure}[t]{0.48\textwidth}
		\centering
		\includegraphics[width=\linewidth]
		{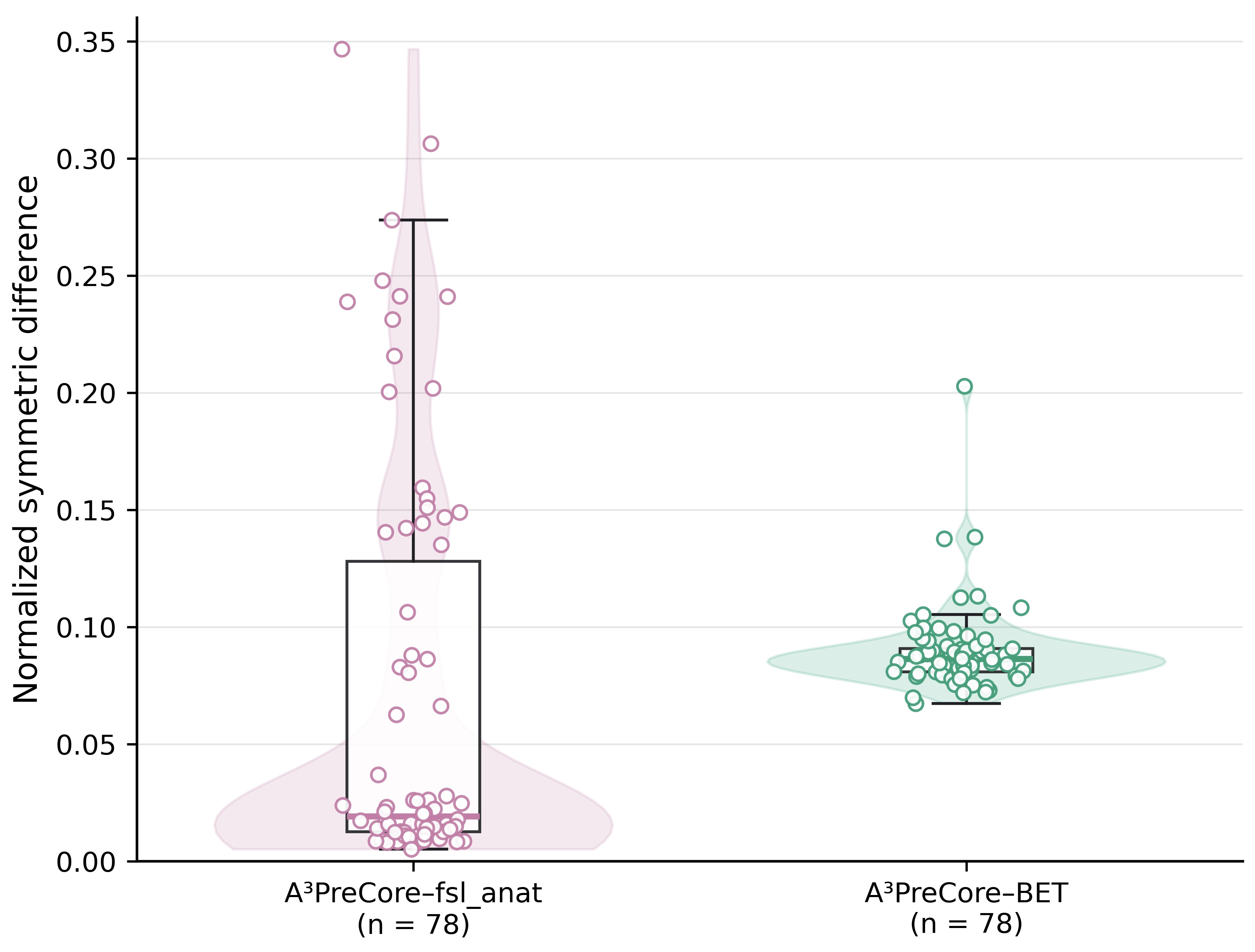}
		\caption{Normalized symmetric difference.}
		\label{fig:agreement_nsd}
	\end{subfigure}
	
	\caption{
		\textbf{Boundary and surface agreement of
			A\textsuperscript{3}PreCore with the \texttt{fsl\_anat} and BET comparators.}
		\textbf{a}, HD95.
		\textbf{b}, Average symmetric surface distance (ASSD).
		\textbf{c}, Surface Dice at a 1-mm tolerance.
		\textbf{d}, Normalized symmetric difference.
		These measures characterize pairwise inter-method agreement and do not
		represent error against an independently delineated anatomical
		ground-truth reference.
	}
	\label{fig:boundary_surface_agreement}
\end{figure}

Boundary- and surface-based measures provided complementary information on
the spatial scale of disagreement with each comparator. For the
A\textsuperscript{3}PreCore--\texttt{fsl\_anat} comparison, the median normalized
symmetric difference was 0.019 (IQR, 0.013--0.128), the median HD95 was
0.777~mm (IQR, 0.693--5.423~mm), and the median ASSD was 0.274~mm
(IQR, 0.186--2.272~mm). For the
A\textsuperscript{3}PreCore--BET comparison, the corresponding values were
0.086 (IQR, 0.081--0.091), 4.005~mm
(IQR, 3.638--4.513~mm), and 1.477~mm
(IQR, 1.384--1.583~mm), respectively. Surface Dice at tolerances of
1, 2, and 3~mm were 0.996, 1.000, and 1.000, respectively, for
A\textsuperscript{3}PreCore--\texttt{fsl\_anat} and 0.476, 0.732, and 0.891 for
A\textsuperscript{3}PreCore--BET
(Table~\ref{tab:boundary_surface_disagreement};
Figure~\ref{fig:boundary_surface_agreement}).
These pairwise values describe agreement with different comparators and were
not interpreted as direct evidence that either comparator constituted a more
accurate anatomical reference.

\begin{table}[htbp]
	
	\caption{
		Boundary and surface disagreement between A\textsuperscript{3}PreCore and
		the \texttt{fsl\_anat} and BET comparators.
	}
	\label{tab:boundary_surface_disagreement}
	
	\centering
	\small
	\renewcommand{\arraystretch}{1.15}
	
	\begin{tabular*}{0.98\linewidth}{
			@{\extracolsep{\fill}}
			lcc
			@{}
		}
		\hline
		
		Metric
		& A\textsuperscript{3}PreCore--\texttt{fsl\_anat}
		& A\textsuperscript{3}PreCore--BET \\
		
		\hline
		
		Normalized symmetric difference
		& 0.019 (0.013--0.128)
		& 0.086 (0.081--0.091) \\
		
		HD95 (mm)
		& 0.777 (0.693--5.423)
		& 4.005 (3.638--4.513) \\
		
		ASSD (mm)
		& 0.274 (0.186--2.272)
		& 1.477 (1.384--1.583) \\
		
		Surface Dice, 1 mm
		& 0.996 (0.318--0.999)
		& 0.476 (0.455--0.500) \\
		
		Surface Dice, 2 mm
		& 1.000 (0.537--1.000)
		& 0.732 (0.707--0.760) \\
		
		Surface Dice, 3 mm
		& 1.000 (0.729--1.000)
		& 0.891 (0.862--0.907) \\
		
		\hline
	\end{tabular*}
	
	\par\smallskip
	\footnotesize
	Values are median (interquartile range). HD95, 95th-percentile Hausdorff
	distance; ASSD, average symmetric surface distance. Lower values indicate
	closer agreement for normalized symmetric difference, HD95, and ASSD,
	whereas higher values indicate closer agreement for surface Dice.
	These measures quantify inter-method disagreement rather than error against
	an independently delineated anatomical reference.
	
\end{table}

Regional and slice-wise analyses further characterized the spatial
distribution of between-method disagreement. For the
A\textsuperscript{3}PreCore--\texttt{fsl\_anat} comparison, median regional
Dice was 0.961 in the inferior third, 0.991 in the middle third, and 0.989
in the superior third. The corresponding values for
A\textsuperscript{3}PreCore--BET were 0.870, 0.962, and 0.942,
respectively.

At the slice level, the participant-level mean slice Dice had a median of
0.975 (IQR, 0.869--0.983) for
A\textsuperscript{3}PreCore--\texttt{fsl\_anat} and 0.883
(IQR, 0.872--0.897) for A\textsuperscript{3}PreCore--BET. The corresponding
participant-level median slice Dice values were 0.990
(IQR, 0.940--0.994) and 0.956 (IQR, 0.952--0.959), respectively.
To characterize lower-tail slice-wise agreement without relying on a single
small terminal slice, the 10th-percentile slice Dice was retained as the
adverse slice-wise summary. Its median was 0.962
(IQR, 0.795--0.977) for
A\textsuperscript{3}PreCore--\texttt{fsl\_anat} and 0.776
(IQR, 0.704--0.828) for A\textsuperscript{3}PreCore--BET
(Supplementary Table~\ref{tab:regional_slice_disagreement}).

Whole-mask volume analysis showed that similar total mask volumes did not
necessarily imply close spatial alignment. For the
A\textsuperscript{3}PreCore--\texttt{fsl\_anat} comparison, the median
signed volume difference, calculated as A\textsuperscript{3}PreCore minus
comparator volume, was $-0.370$~mL
(IQR, $-0.647$ to 0.136~mL). The median absolute difference was
0.464~mL (IQR, 0.217--0.839~mL), corresponding to only 0.030\%
(IQR, 0.015--0.061\%) of mask volume. A\textsuperscript{3}PreCore had the
smaller mask volume in 53 participants and the larger volume in 25.
Nevertheless, 21/78 comparisons had a normalized symmetric difference
greater than 0.10, 21/78 had an HD95 greater than 5~mm, and 11/78 had an
HD95 greater than 10~mm; none had an absolute volume difference greater
than 2\%.

For the A\textsuperscript{3}PreCore--BET comparison, the median signed
volume difference was $-24.013$~mL
(IQR, $-50.048$ to 4.521~mL), while the median absolute difference was
30.277~mL (IQR, 17.690--56.709~mL), corresponding to 2.136\%
(IQR, 1.114--3.910\%). A\textsuperscript{3}PreCore had the smaller mask
volume in 56 participants and the larger volume in 22. Nine of 78
comparisons had a normalized symmetric difference greater than 0.10,
12/78 had an HD95 greater than 5~mm, 2/78 exceeded 10~mm, and 41/78 had
an absolute volume difference greater than 2\%
(Supplementary Table~\ref{tab:volume_disagreement}).

Taken together, the spatial-agreement analyses indicate that
A\textsuperscript{3}PreCore and \texttt{fsl\_anat} were closely concordant
when \texttt{fsl\_anat} did not exhibit gross misregistration, whereas the
largest reductions in overlap and increases in directional non-overlap were
concentrated in the \texttt{fsl\_anat} gross-misregistration subgroup.
The comparison with BET provided a complementary registration-independent
reference for characterizing agreement with a mask generated directly in
native FLAIR space, without implying anatomical ground truth. Notably,
substantial spatial and boundary disagreement could occur despite only
minimal differences in total mask volume, demonstrating that global volume
similarity alone was insufficient to characterize the observed registration
discrepancies. The geometric characteristics of these participant-specific
differences are examined in the following section.

Relative to BET, A\textsuperscript{3}PreCore showed a slightly lower
old-QC score, with a median paired difference of $-0.000857$
(bootstrap 95\% CI, $-0.001277$ to $-0.000553$;
$p<0.001$; rank-biserial effect size $=-0.518$).
In contrast, the paired difference in $S_{\mathrm{mask}}$ was small and
not statistically significant
(median difference, $-0.000275$; bootstrap 95\% CI,
$-0.000655$ to $0.000177$; $p=0.369$;
rank-biserial effect size $=-0.117$).
All paired analyses were performed within the same 78-participant cohort
(Table~\ref{tab:eqc_paired_statistics}).

Taken together, the scalar QC results show that the substantial reduction in
gross registration failure achieved by A\textsuperscript{3}PreCore was not
accompanied by a broad deterioration in native-FLAIR mask plausibility.
At the same time, the close clustering and ceiling behaviour of several
scalar QC components demonstrate why a single global score is insufficient
to characterize the failure modes observed in this study. The principal
advantage of E-QC therefore lies in extending quality assessment beyond
global plausibility to explicitly characterize participant-specific spatial
and geometric disagreement. The geometric component of this profile is
examined in the following section, where centroid displacement,
PCA-derived orientation difference, and orientation-induced boundary
displacement reveal discrepancies that remained largely inconspicuous in the
scalar QC summaries.

% ============================================================
\subsection{Method-Specific Extended Quantitative Quality-Control Characterization}
\label{subsec:eqc_analysis}
% ============================================================

The method-specific E-QC analysis examined whether the improved
registration robustness of A\textsuperscript{3}PreCore was accompanied
by systematic changes in native-FLAIR mask characteristics.
Consistent with the multidimensional formulation introduced in
Section~\ref{sec:extended_qc}, six descriptors were evaluated
independently: physical mask volume ($V$), FOV-relative centroid
position ($C$), mask-to-complement FLAIR intensity ratio ($R_I$),
boundary-gradient support ($B_G$), peripheral foreground occupation
($P_{\mathrm{peripheral}}$), and inferior-tail fraction
($P_{\mathrm{tail}}$). No composite E-QC score was used for
method comparison.

The common analysis cohort comprised 78 participants with valid BET,
\texttt{fsl\_anat}, and A\textsuperscript{3}PreCore masks.
A\textsuperscript{3}PreCore and \texttt{fsl\_anat} showed closely
matched distributions for global mask extent and FOV-relative position.
Median $V$ was 1465.5~mL for A\textsuperscript{3}PreCore,
1465.8~mL for \texttt{fsl\_anat}, and 1503.7~mL for BET,
whereas median $C$ was 0.08075, 0.08076, and 0.07972,
respectively. The corresponding descriptor distributions are summarized
in Table~\ref{tab:eqc_method_summary}.

Greater method-dependent differences were observed for the
image- and boundary-related descriptors. Median $R_I$ was
5.4330 for A\textsuperscript{3}PreCore, 5.2794 for
\texttt{fsl\_anat}, and 5.2696 for BET, while median $B_G$
was 2.7034, 2.4010, and 2.2592, respectively.
Peripheral foreground occupation remained low across all three methods,
with median $P_{\mathrm{peripheral}}$ values of 0.00618,
0.00622, and 0.00705 for A\textsuperscript{3}PreCore,
\texttt{fsl\_anat}, and BET, respectively.
Median $P_{\mathrm{tail}}$ was similarly small in absolute magnitude
(0.02720, 0.02602, and 0.02449, respectively).

Paired comparison with the primary \texttt{fsl\_anat} comparator
showed that the difference in physical mask volume was numerically
small despite a consistent signed shift
(median paired difference, $-0.370$~mL;
95\% CI, $-0.475$ to $-0.080$;
$r_{\mathrm{rb}}=-0.505$; $p<0.001$).
No clear paired difference was detected for $C$
($-0.000326$; 95\% CI, $-0.000640$ to $0.000064$;
$r_{\mathrm{rb}}=-0.248$; $p=0.057$).

For $R_I$, the median paired difference was $+0.001397$
(95\% CI, $-0.000601$ to $0.006260$;
$r_{\mathrm{rb}}=0.420$; $p=0.0013$), whereas $B_G$
showed a positive paired shift of $+0.025187$
(95\% CI, $0.004760$ to $0.058166$;
$r_{\mathrm{rb}}=0.552$; $p<0.001$).
Peripheral foreground occupation was essentially unchanged relative to
\texttt{fsl\_anat}
(median difference, $-5.83\times10^{-6}$;
$r_{\mathrm{rb}}=-0.055$; $p=0.674$), and
$P_{\mathrm{tail}}$ likewise showed no clear paired shift
($+0.000071$; $r_{\mathrm{rb}}=0.175$; $p=0.179$)
(Table~\ref{tab:eqc_paired_statistics}).

Relative to BET, the paired differences were more pronounced and
descriptor dependent. A\textsuperscript{3}PreCore showed a median
$V$ difference of $-24.013$~mL, a $C$ difference of
$+0.001028$, an $R_I$ difference of $+0.132618$, and a
$B_G$ difference of $+0.442448$.
Peripheral foreground occupation was lower
(median difference, $-0.000638$;
$r_{\mathrm{rb}}=-0.864$; $p<0.001$), whereas
$P_{\mathrm{tail}}$ was higher
($+0.002093$; $r_{\mathrm{rb}}=0.548$; $p<0.001$).
The corresponding confidence intervals and paired statistics are
reported in Table~\ref{tab:eqc_paired_statistics}.

\begin{figure}[h]
	\centering
	\includegraphics[width=.72\linewidth]
	{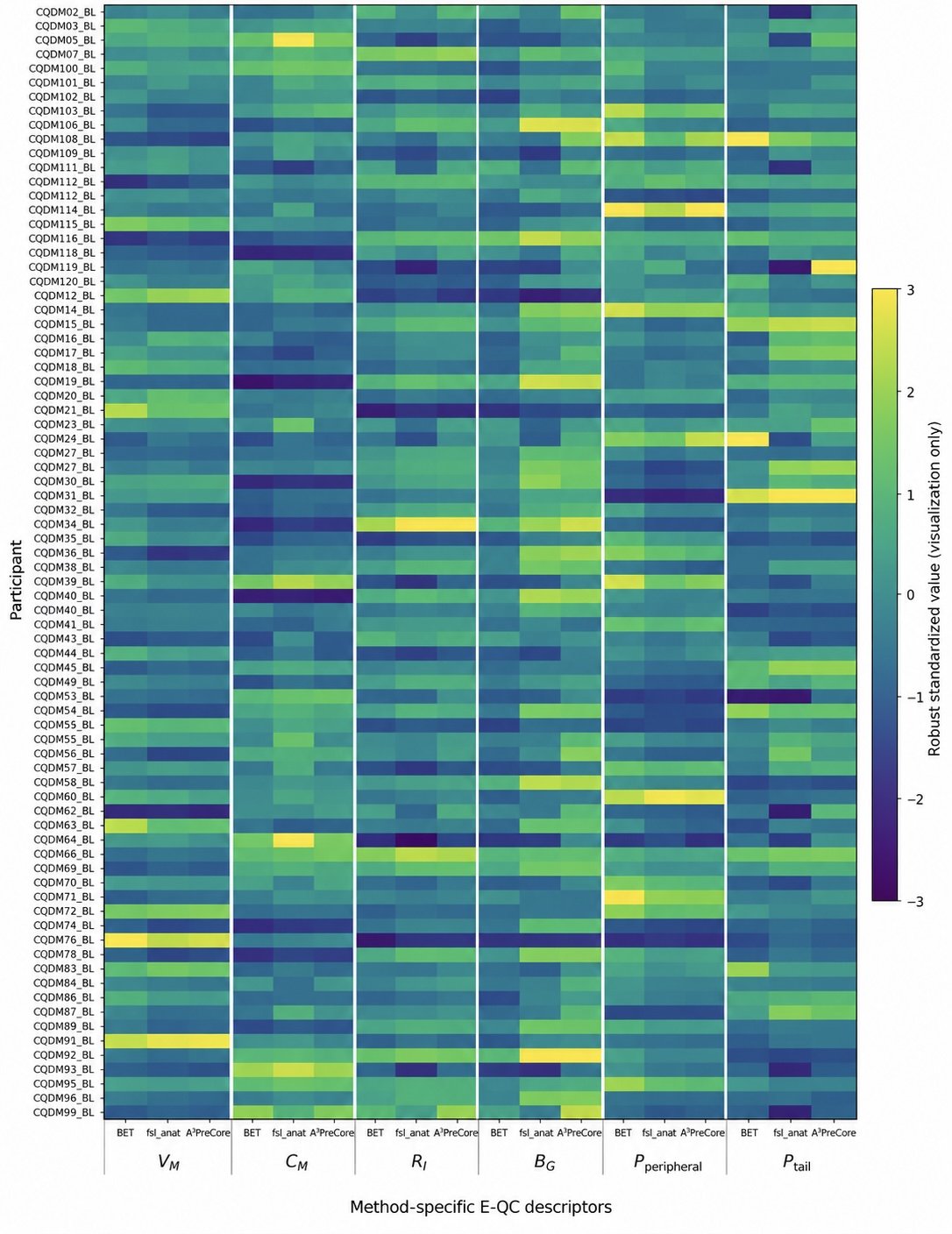}
	\caption{
		\textbf{Participant-level multidimensional E-QC profiles.}
		Rows represent the 78 participants and columns represent the six
		method-specific descriptors ($V$, $C$, $R_I$, $B_G$,
		$P_{\mathrm{peripheral}}$, and $P_{\mathrm{tail}}$) for BET,
		\texttt{fsl\_anat}, and A\textsuperscript{3}PreCore.
		Values were robustly standardized within each descriptor for
		visualization only; all statistical analyses used the original
		descriptor values.
	}
	\label{fig:eqc_heatmap}
\end{figure}

\begin{figure}[htbp]
	\centering
	\includegraphics[width=.7\linewidth]
	{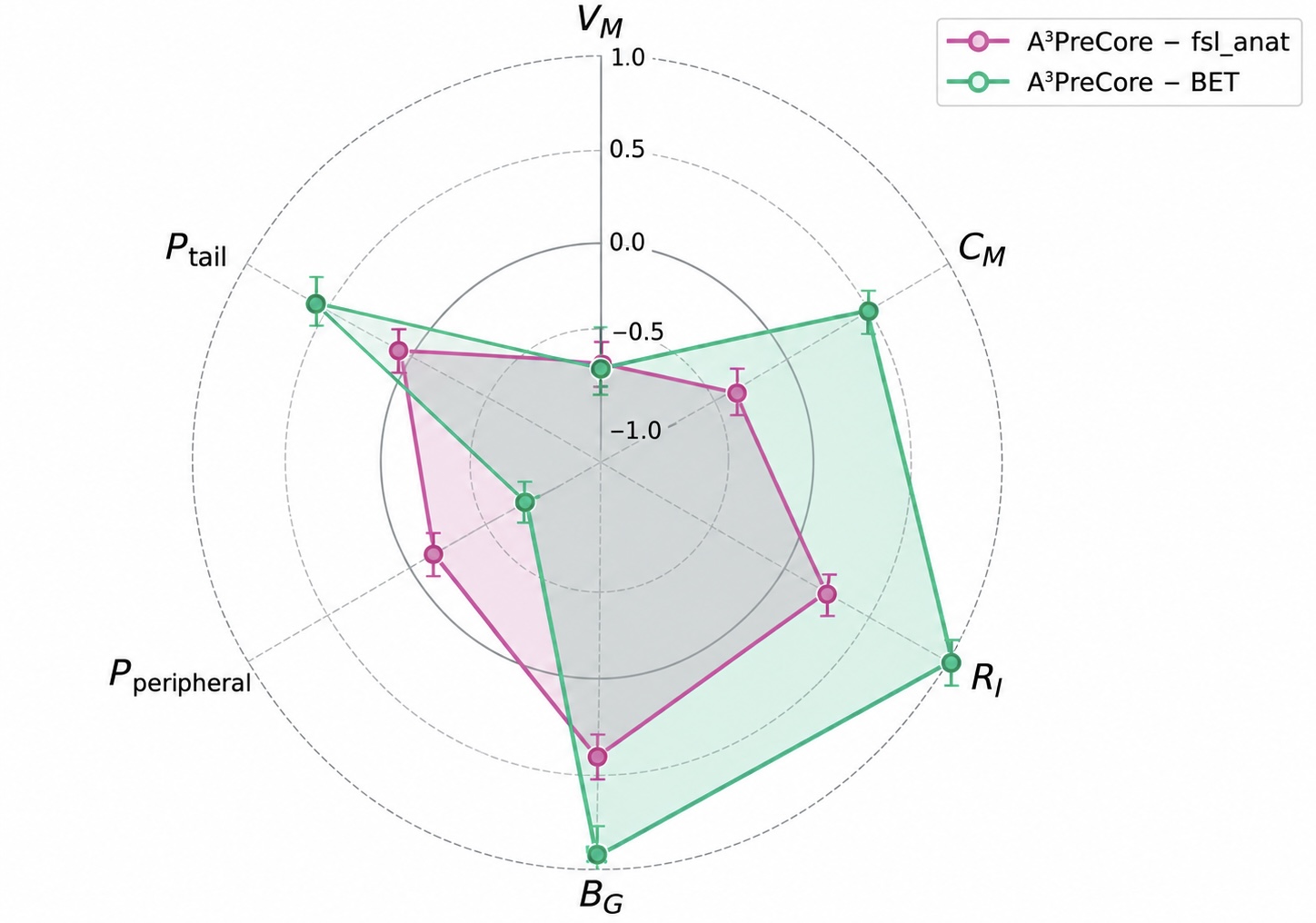}
	\caption{
		\textbf{Paired E-QC descriptor effect profile.}
		Points represent paired rank-biserial effect sizes for
		A\textsuperscript{3}PreCore relative to \texttt{fsl\_anat} and BET
		across the six method-specific E-QC descriptors.
		Positive and negative values indicate the numerical direction of the
		paired difference and do not denote anatomical superiority.
	}
	\label{fig:eqc_effect_profile}
\end{figure}

The participant-level heat map further demonstrated that the
between-method relationships were descriptor specific rather than
uniform across the E-QC profile
(Figure~\ref{fig:eqc_heatmap}). For visualization only, each
descriptor was robustly standardized within its measurement dimension
across the three methods; these standardized values were not used for
inferential analysis. The paired effect profile
(Figure~\ref{fig:eqc_effect_profile}) similarly showed close
A\textsuperscript{3}PreCore--\texttt{fsl\_anat} correspondence for
$C$, $P_{\mathrm{peripheral}}$, and $P_{\mathrm{tail}}$, with
larger directional effects for $R_I$ and $B_G$.
Comparisons with BET showed larger effects across several descriptors,
particularly $R_I$, $B_G$, and $P_{\mathrm{peripheral}}$.
Because the individual coordinates quantify distinct mask and
mask-to-image properties, the direction or magnitude of any single
descriptor was not interpreted as evidence of anatomical superiority.
\begin{table}[htbp]
	\caption{
		Method-specific E-QC descriptor distributions for BET,
		\texttt{fsl\_anat}, and A\textsuperscript{3}PreCore
		in the common 78-participant cohort.
	}
	\label{tab:eqc_method_summary}
	
	\centering
	\small
	\setlength{\tabcolsep}{4pt}
	\renewcommand{\arraystretch}{1.15}
	
	\begin{tabular*}{0.98\linewidth}{
			@{\extracolsep{\fill}}
			lccc
			@{}
		}
		\hline
		
		Metric
		& BET ($n=78$)
		& \texttt{fsl\_anat} ($n=78$)
		& A\textsuperscript{3}PreCore ($n=78$)
		\\
		
		\hline
		
		$V$ (mL)
		& 1503.7 (1391.3--1593.3)
		& 1465.8 (1372.9--1569.2)
		& 1465.5 (1372.8--1566.8)
		\\
		
		$C$
		& 0.07972 (0.07215--0.08637)
		& 0.08076 (0.07386--0.08897)
		& 0.08075 (0.07289--0.08745)
		\\
		
		$R_I$
		& 5.2696 (4.7234--5.8034)
		& 5.2794 (4.7560--5.8867)
		& 5.4330 (4.8589--5.9479)
		\\
		
		$B_G$
		& 2.2592 (1.9709--2.5025)
		& 2.4010 (2.2319--3.0117)
		& 2.7034 (2.3276--3.0739)
		\\
		
		$P_{\mathrm{peripheral}}$
		& 0.00705 (0.00525--0.00875)
		& 0.00622 (0.00472--0.00800)
		& 0.00618 (0.00459--0.00796)
		\\
		
		$P_{\mathrm{tail}}$
		& 0.02449 (0.02107--0.02945)
		& 0.02602 (0.02114--0.03201)
		& 0.02720 (0.02277--0.03260)
		\\
		
		\hline
	\end{tabular*}
	\par\smallskip
	\footnotesize
	Values are median (interquartile range).
	$V$ denotes physical mask volume;
	$C$, FOV-relative centroid offset;
	$R_I$, mask-to-complement FLAIR intensity ratio;
	$B_G$, boundary-gradient support;
	$P_{\mathrm{peripheral}}$, peripheral foreground occupation;
	and $P_{\mathrm{tail}}$, inferior-tail fraction.
	Descriptors were analyzed independently and do not
	directly measure anatomical accuracy.
\end{table}

\begin{table}[htbp]
	\caption{
		Participant-level paired comparisons of six method-specific
		E-QC descriptors for A\textsuperscript{3}PreCore
		relative to \texttt{fsl\_anat} and BET ($n=78$).
	}
	\label{tab:eqc_paired_statistics}
	\centering
	\scriptsize
	\setlength{\tabcolsep}{2.5pt}
	\renewcommand{\arraystretch}{1.12}
	\begin{tabular*}{0.98\linewidth}{@{\extracolsep{\fill}}llcccccc@{}}
		\hline
		Comparator & Descriptor
		& A\textsuperscript{3}PreCore median & Comparator median
		& Paired difference & Bootstrap 95\% CI & $r_{\mathrm{rb}}$ & $p$ \\
		\hline
		\texttt{fsl\_anat}
		& $V$ (mL) & 1465.502 & 1465.765 & $-0.370$ & $-0.475$ to $-0.080$ & $-0.505$ & $<0.001$ \\
		& $C$ & 0.080750 & 0.080761 & $-0.000326$ & $-0.000640$ to $0.000064$ & $-0.248$ & 0.057 \\
		& $R_{I}$ & 5.432968 & 5.279435 & $+0.001397$ & $-0.000601$ to $0.006260$ & 0.420 & 0.0013 \\
		& $B_{G}$ & 2.703394 & 2.401043 & $+0.025187$ & $0.004760$ to $0.058166$ & 0.552 & $<0.001$ \\
		& $P_{\mathrm{peripheral}}$ & 0.006176
		& 0.006222
		& $-5.83\times10^{-6}$
		& $-1.59\times10^{-5}$ to $7.31\times10^{-6}$
		& $-0.055$
		& 0.674
		\\
		& $P_{\mathrm{tail}}$ & 0.027196 & 0.026017 & $+0.000071$ & $-0.000127$ to $0.000337$ & 0.175 & 0.179 \\
		\hline
		BET
		& $V$ (mL) & 1465.502 & 1503.693 & $-24.013$ & $-30.630$ to $-10.841$ & $-0.548$ & $<0.001$ \\
		& $C$ & 0.080750 & 0.079721 & $+0.001028$ & $0.000663$ to $0.001604$ & 0.518 & $<0.001$ \\
		& $R_{I}$ & 5.432968 & 5.269639 & $+0.132618$ & $0.119944$ to $0.148070$ & 0.994 & $<0.001$ \\
		& $B_{G}$ & 2.703394 & 2.259194 & $+0.442448$ & $0.379789$ to $0.553584$ & 0.962 & $<0.001$ \\
		& $P_{\mathrm{peripheral}}$ & 0.006176
		& 0.007047
		& $-0.000638$
		& $-0.000791$ to $-0.000501$
		& $-0.864$
		& $<0.001$
		\\
		& $P_{\mathrm{tail}}$ & 0.027196 & 0.024493 & $+0.002093$ & $0.001310$ to $0.003020$ & 0.548 & $<0.001$ \\
		\hline
	\end{tabular*}
	\par\smallskip
	\footnotesize
	Paired differences are reported as median
	(A\textsuperscript{3}PreCore minus comparator).
	Bootstrap 95\% confidence intervals were estimated
	using 10,000 paired resamples.
	Two-sided Wilcoxon signed-rank tests and paired
	rank-biserial effect sizes ($r_{\mathrm{rb}}$) are reported.
\end{table}

% ============================================================
\subsection{Geometric Characterization of Inter-Method Mask Disagreement}
\label{subsec:geometric_failures}
% ============================================================

The method-specific E-QC descriptors in the preceding section
characterized each candidate mask independently, but did not directly
quantify geometric disagreement between methods. We therefore examined
three complementary pairwise descriptors: physical centroid displacement,
PCA-derived principal-axis orientation difference, and
PCA-orientation-induced boundary displacement. Together, these measures
separate differences in mask position from differences in dominant spatial
orientation and from the physical boundary-scale consequence of that
orientation disagreement. Because the PCA-based quantities were derived
from the geometry of the candidate masks, they were interpreted as
descriptors of inter-method disagreement rather than as direct estimates
of the underlying registration-transform parameters.

\begin{figure}[h]
	\centering
	% -------------------- Panel A --------------------
	\begin{subfigure}[t]{0.48\linewidth}
		\centering
		\includegraphics[width=\linewidth]
		{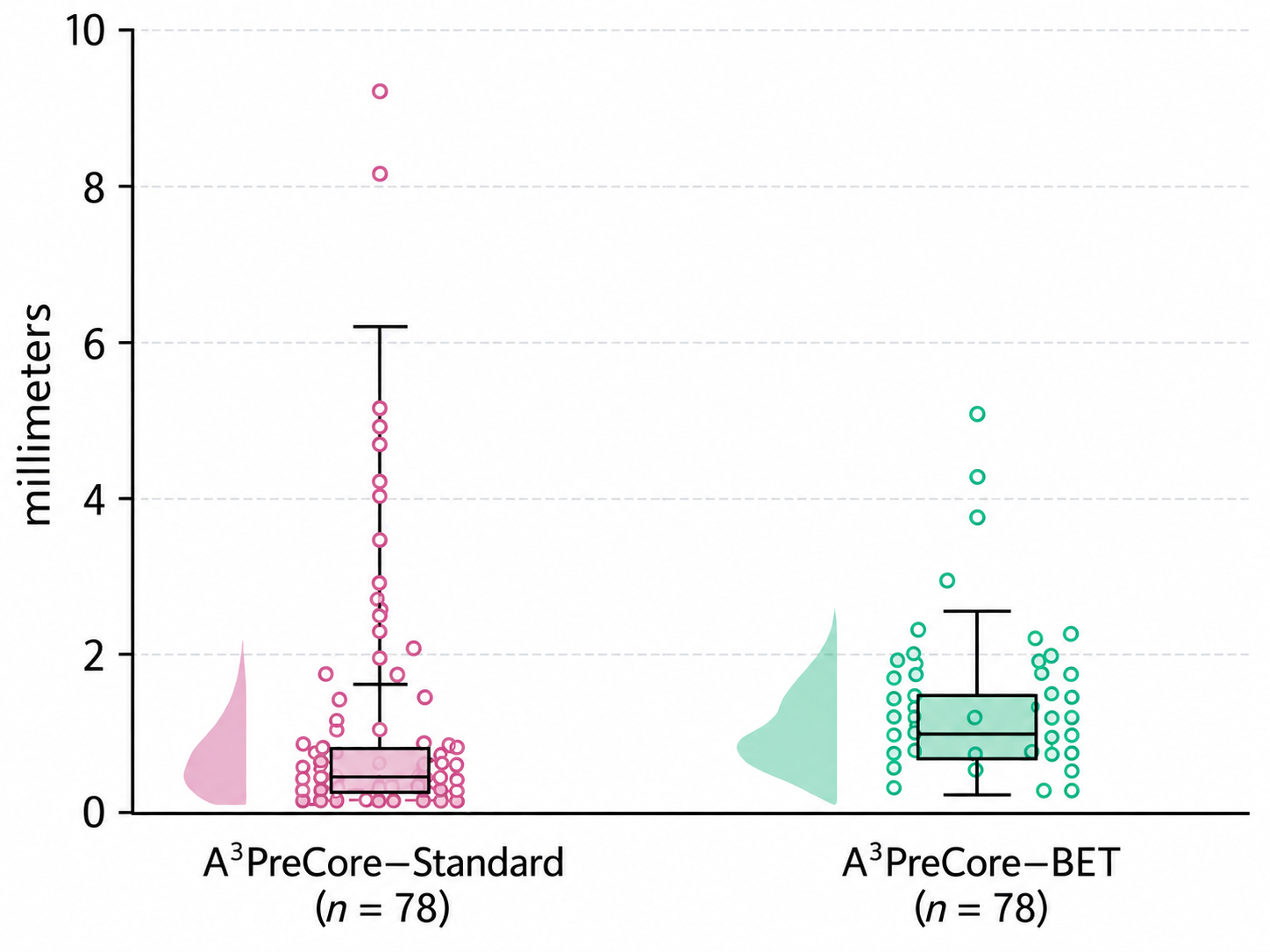}
		\caption{Centroid displacement.}
		\label{fig:geometry_centroid}
	\end{subfigure}
	\hfill
	% -------------------- Panel B --------------------
	\begin{subfigure}[t]{0.48\linewidth}
		\centering
		\includegraphics[width=\linewidth]
		{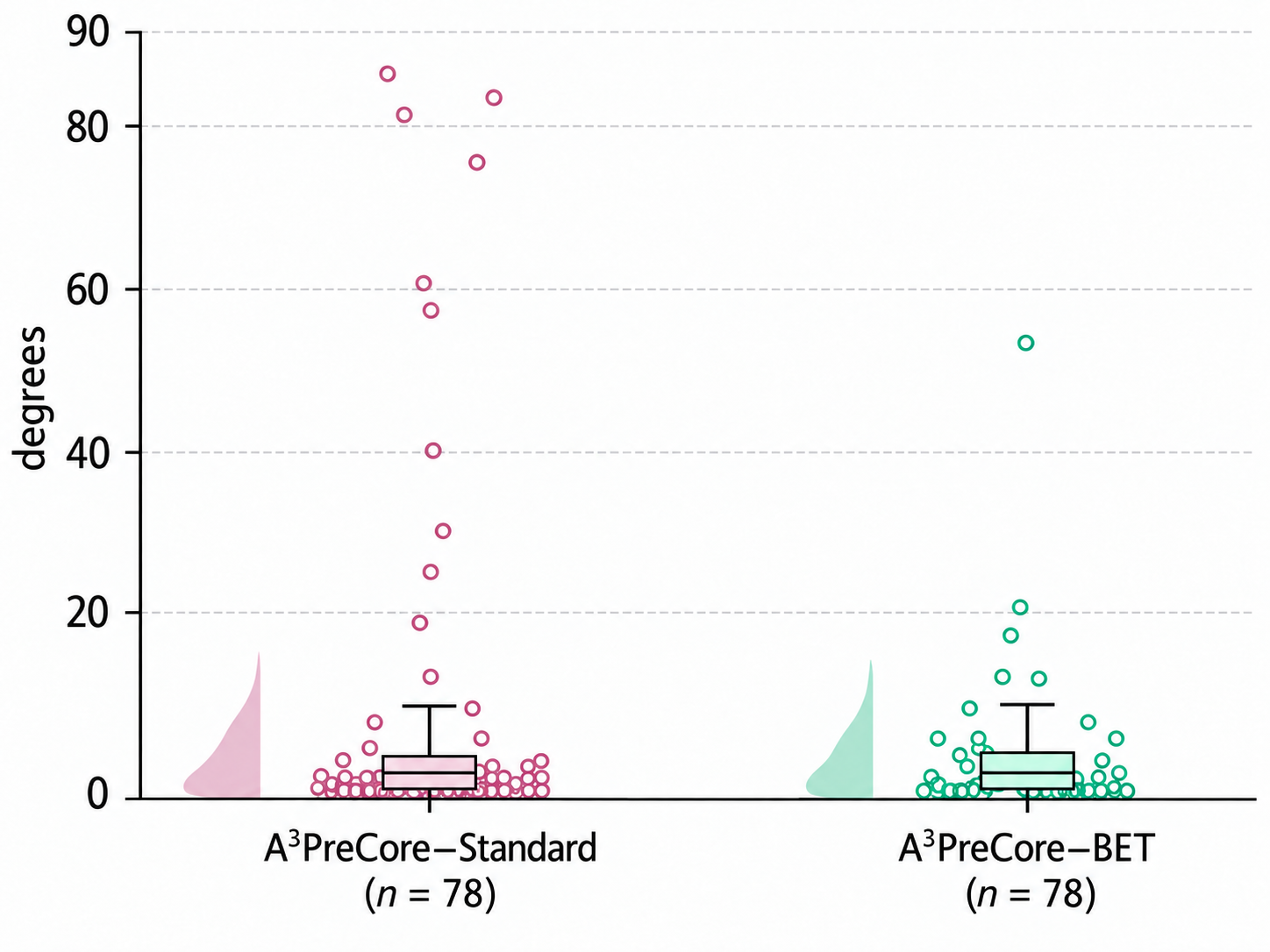}
		\caption{PCA-derived orientation difference.}
		\label{fig:geometry_orientation}
	\end{subfigure}
	
	\vspace{0.7em}
	
	% -------------------- Panel C --------------------
	\begin{subfigure}[t]{0.48\linewidth}
		\centering
		\includegraphics[width=\linewidth]
		{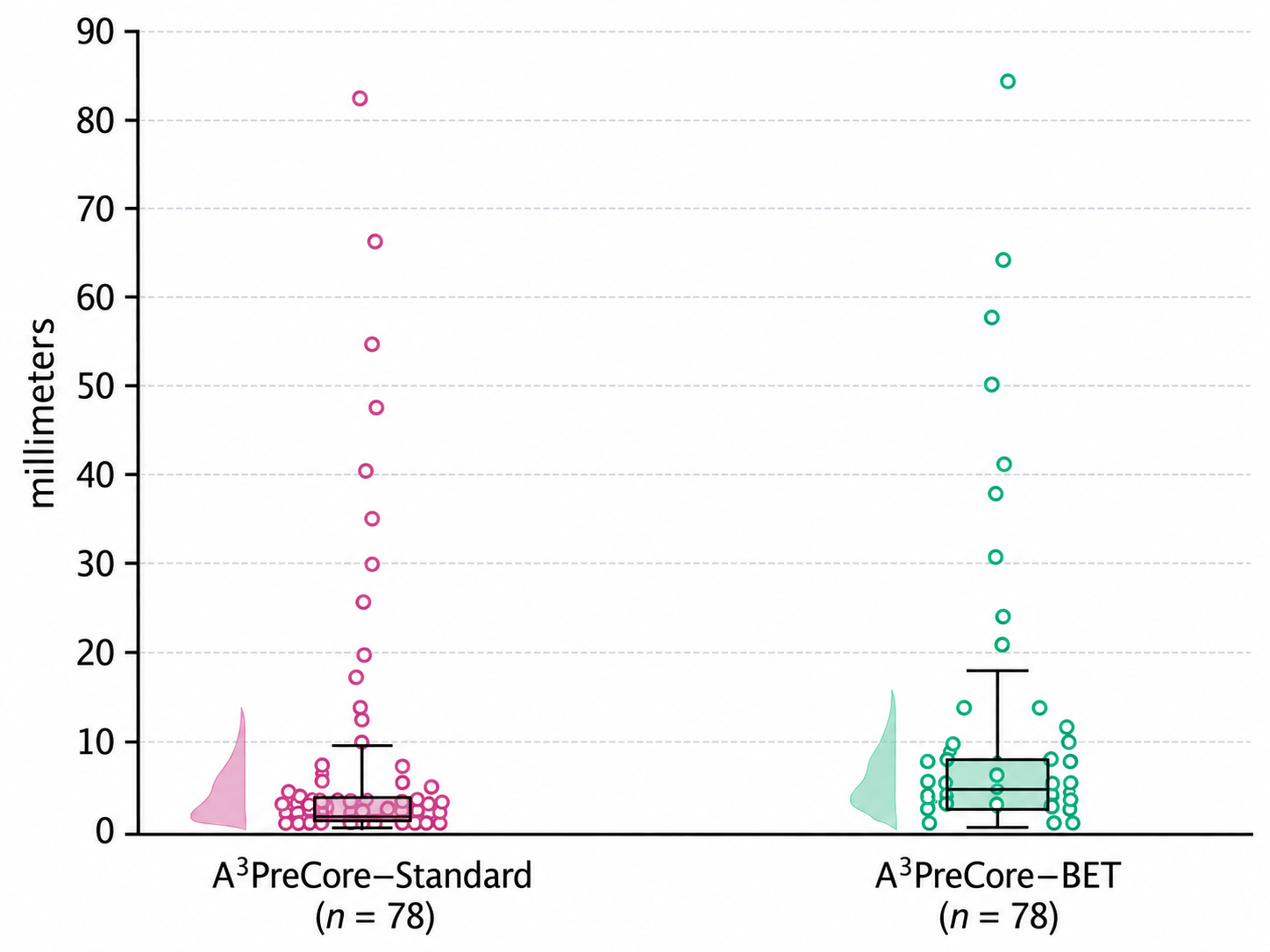}
		\caption{PCA-orientation-induced boundary displacement.}
		\label{fig:geometry_boundary}
	\end{subfigure}
	\hfill
	% -------------------- Panel D --------------------
	\begin{subfigure}[t]{0.48\linewidth}
		\centering
		\includegraphics[width=\linewidth]
		{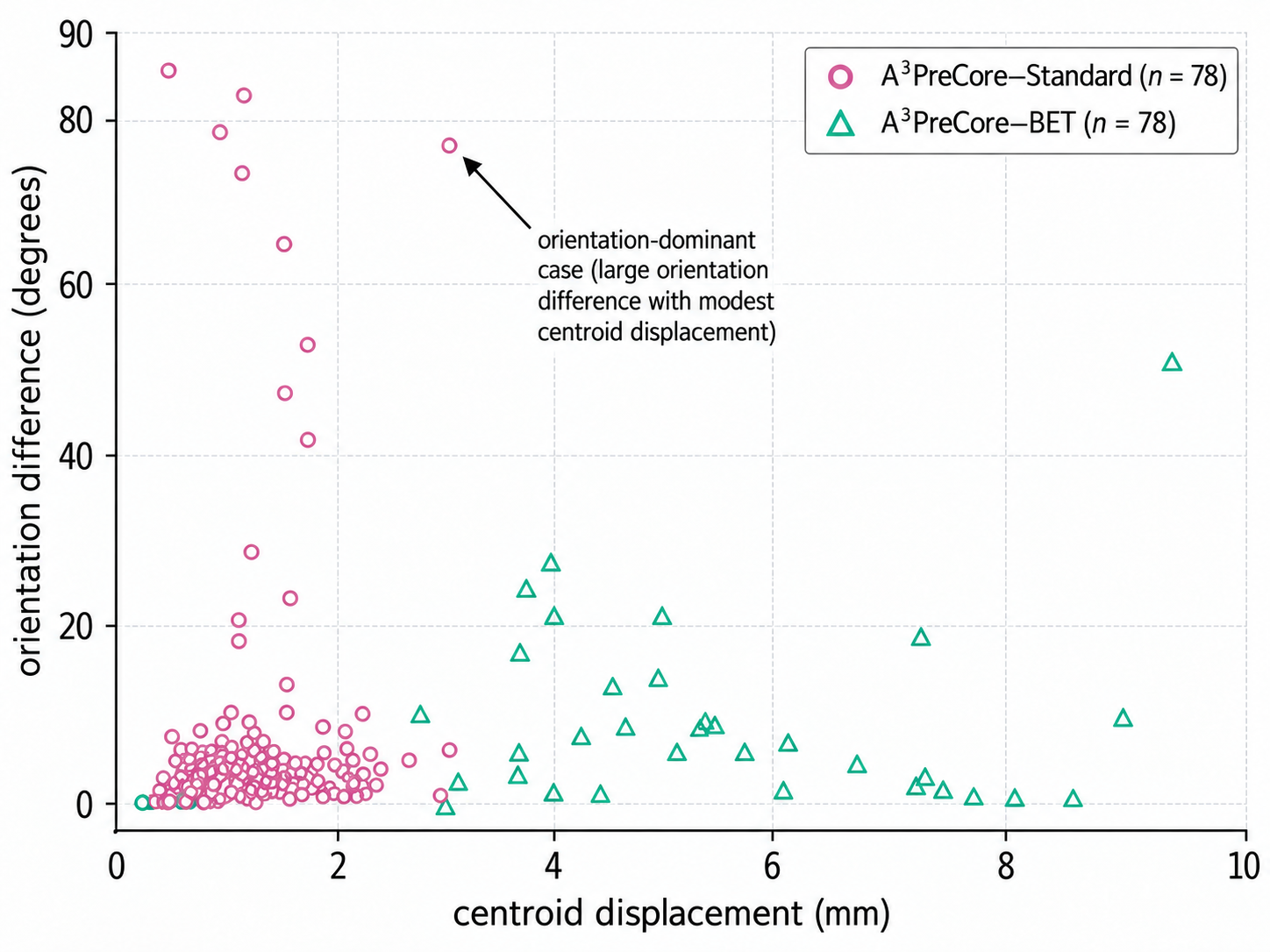}
		\caption{Centroid displacement versus orientation difference.}
		\label{fig:geometry_center_orientation}
	\end{subfigure}
	\caption{
		\textbf{Pairwise geometric characterization of inter-method mask
			disagreement.}
		\textbf{a}, Participant-level physical centroid displacement.
		\textbf{b}, PCA-derived principal-axis orientation difference.
		\textbf{c}, PCA-orientation-induced boundary displacement.
		\textbf{d}, Relationship between centroid displacement and orientation
		difference, illustrating that pronounced orientation disagreement may
		occur despite only modest centroid displacement.
		Magenta denotes A\textsuperscript{3}PreCore--\texttt{fsl\_anat}
		comparisons and green denotes A\textsuperscript{3}PreCore--BET
		comparisons. Distributions are shown for the common 78-participant
		cohort. These descriptors quantify inter-method geometric disagreement
		and are not interpreted as anatomical error relative to an independent
		ground-truth reference.
	}
	\label{fig:geometry_characterization}
\end{figure}

Across the common 78-participant cohort,
A\textsuperscript{3}PreCore and \texttt{fsl\_anat} showed close geometric
agreement for the majority of participants. Median centroid displacement
was 0.386~mm (IQR, 0.247--2.578~mm), median PCA-derived orientation
difference was $0.539^{\circ}$ (IQR,
$0.383^{\circ}$--$1.414^{\circ}$), and median
PCA-orientation-induced boundary displacement was 0.551~mm
(IQR, 0.383--1.412~mm)
(Table~\ref{tab:geometry_summary};
Figure~\ref{fig:geometry_characterization}a--c).
The participant-level distributions nevertheless contained a pronounced
upper tail, particularly for orientation difference and its associated
boundary displacement, indicating that small cohort-level medians coexist
with a limited subset of substantially larger geometric disagreements.

The A\textsuperscript{3}PreCore--BET comparison exhibited a different
distributional pattern. Median centroid displacement was 1.170~mm
(IQR, 0.698--1.586~mm), median PCA-derived orientation difference was
$3.442^{\circ}$ (IQR, $2.273^{\circ}$--$4.810^{\circ}$), and median
PCA-orientation-induced boundary displacement was 3.496~mm
(IQR, 2.292--4.951~mm). Thus, relative to BET,
A\textsuperscript{3}PreCore showed larger typical pairwise geometric
differences than relative to \texttt{fsl\_anat}. However, the distributions
for both comparisons were right-skewed and contained participant-level
outliers, emphasizing that median values alone do not fully characterize
inter-method geometric behaviour.

The centroid-displacement distributions further illustrate the difference
between typical and extreme behaviour
(Figure~\ref{fig:geometry_characterization}a).
Most A\textsuperscript{3}PreCore--\texttt{fsl\_anat} comparisons were
concentrated at small physical displacements, although several participants
formed a distinct upper tail. The BET comparison showed a higher central
displacement and a broader participant-level distribution, with individual
cases extending well beyond the main concentration of observations.

An even stronger separation between central tendency and participant-level
extremes was observed for PCA-derived orientation difference
(Figure~\ref{fig:geometry_characterization}b).
The majority of A\textsuperscript{3}PreCore--\texttt{fsl\_anat}
comparisons clustered near zero, but several observations exhibited large
orientation differences extending to the upper range of the distribution.
A\textsuperscript{3}PreCore--BET comparisons showed a higher typical
orientation difference, while also containing isolated high-orientation
cases. These findings indicate that the two comparisons differ not only in
their central tendency but also in the structure of their distributional
tails.

PCA-orientation-induced boundary displacement showed a corresponding
pattern
(Figure~\ref{fig:geometry_characterization}c).
Most participant-level values remained relatively small, whereas a limited
number of comparisons produced substantially larger physical boundary
displacements. This behaviour is consistent with the fact that a large
difference in principal-axis orientation can have a substantial physical
effect at the mask boundary even when the mask centroid remains relatively
stable.

This relationship is demonstrated directly in the joint centroid--orientation
plot
(Figure~\ref{fig:geometry_characterization}d).
Centroid displacement and PCA-derived orientation difference were not
coupled in a one-to-one manner. Several comparisons exhibited pronounced
orientation disagreement despite only modest centroid displacement,
including an orientation-dominant case highlighted in the figure.
Conversely, some comparisons with larger centroid displacement showed only
moderate orientation differences. Centroid-based assessment alone would
therefore fail to capture important classes of geometric disagreement.

Taken together, the pairwise geometric analysis revealed a multidimensional
failure structure that could not be represented adequately by any single
geometric measure. A\textsuperscript{3}PreCore and \texttt{fsl\_anat}
were closely aligned for most participants, but the
A\textsuperscript{3}PreCore--\texttt{fsl\_anat} distribution contained a
distinct subset of large orientation-related discrepancies.
The BET comparison showed larger typical differences in centroid position,
orientation, and orientation-induced boundary displacement, while also
exhibiting participant-specific extremes. Because BET is generated directly
in native FLAIR space, this comparison provides a complementary
registration-independent description of inter-method geometry rather than
an anatomical reference standard.

Overall, these results demonstrate that centroid displacement, principal-axis
orientation difference, and orientation-induced boundary displacement
capture complementary aspects of mask disagreement. Their joint evaluation
therefore provides substantially more information about participant-specific
geometric behaviour than centroid position or a global mask descriptor
alone.

\begin{table}[htbp]
	\caption{
		Pairwise geometric characterization of
		A\textsuperscript{3}PreCore relative to
		\texttt{fsl\_anat} and BET in the common
		78-participant cohort.
	}
	\label{tab:geometry_summary}
	
	\centering
	\scriptsize
	\setlength{\tabcolsep}{3.5pt}
	\renewcommand{\arraystretch}{1.15}
	
	\begin{tabular*}{0.98\linewidth}{
			@{\extracolsep{\fill}}
			lcccc
			@{}
		}
		\hline
		
		Comparison
		& Centroid displacement
		& PCA orientation
		& PCA-orientation-induced displacement
		& Orientation $>10^{\circ}$
		\\
		
		& (mm) & ($^{\circ}$) & (mm) & $n$\\
		
		\hline
		
		A\textsuperscript{3}PreCore--\texttt{fsl\_anat}
		& 0.386 (0.247--2.578)
		& 0.539 (0.383--1.414)
		& 0.551 (0.383--1.412)
		& 13/78
		\\
		
		A\textsuperscript{3}PreCore--BET
		& 1.170 (0.698--1.586)
		& 3.442 (2.273--4.810)
		& 3.496 (2.292--4.951)
		& 2/78
		\\
		
		\hline
	\end{tabular*}
	
	\par\smallskip
	\footnotesize
	Continuous values are median (interquartile range).
	Centroid displacement quantifies pairwise separation of mask centroids
	in physical space. PCA orientation denotes the sign-resolved
	principal-axis orientation difference, and PCA-orientation-induced
	displacement summarizes its physical boundary-scale consequence.
	These quantities characterize inter-method geometry and are not measures
	of anatomical error relative to an independent reference.
\end{table}

% ============================================================
\subsection{Blinded Expert Ranking of Brain-Mask Quality}
\label{subsec:blinded_expert_results}
% ============================================================

The method-blinded expert assessment comprised 120 masks from 40 randomly
selected participants, with one mask generated by each of the three methods
for every participant. The resulting participant-level rankings showed a
clear descriptive preference for A\textsuperscript{3}PreCore
(Table~\ref{tab:blinded_expert_ranking}).

A\textsuperscript{3}PreCore received rank~1 in 35 of 40 participants
(87.5\%), compared with 3 of 40 (7.5\%) for BET and 2 of 40 (5.0\%) for
\texttt{fsl\_anat}. A\textsuperscript{3}PreCore was ranked second in
3 participants (7.5\%) and third in only 2 participants (5.0\%).
In contrast, BET received rank~2 in 11 participants (27.5\%) and rank~3
in 26 participants (65.0\%), whereas \texttt{fsl\_anat} received rank~2
in 26 participants (65.0\%) and rank~3 in 12 participants (30.0\%).

Accordingly, A\textsuperscript{3}PreCore was the preferred mask in the
large majority of the blinded participant-level comparisons, while being
ranked least preferred in only a small minority of cases. This pattern
provides an independent qualitative complement to the quantitative findings
reported in the preceding sections: the method that avoided the gross
registration failures observed with \texttt{fsl\_anat} was also most
frequently preferred by blinded neuroscientist reviewers.

Qualitative reviewer feedback most commonly emphasized the perceived
precision and spatial plausibility of the A\textsuperscript{3}PreCore
brain-mask boundaries. In particular, reviewer preference was consistent
with the cleaner boundary behaviour observed in the spatial and qualitative
analyses, including reduced inferior or extracranial extension in comparison
with BET in the displayed cases. Because no independently delineated
native-FLAIR brain-mask reference was available, these rankings were
interpreted as method-blinded expert judgments of relative mask quality and
anatomical plausibility rather than as direct measurements of segmentation
accuracy.

\begin{table}[htbp]
	
	\caption{
		Method-blinded expert ranking of brain-mask quality in the random
		40-participant evaluation subset.
	}
	\label{tab:blinded_expert_ranking}
	
	\centering
	\small
	\renewcommand{\arraystretch}{1.15}
	\setlength{\tabcolsep}{6pt}
	
	\begin{tabular*}{0.98\linewidth}{
			@{\extracolsep{\fill}}
			lccc
			@{}
		}
		\hline
		
		Method
		& Rank 1, $n$ (\%)
		& Rank 2, $n$ (\%)
		& Rank 3, $n$ (\%)
		\\
		
		\hline
		
		A\textsuperscript{3}PreCore
		& 35 (87.5\%)
		& 3 (7.5\%)
		& 2 (5.0\%)
		\\
		
		BET
		& 3 (7.5\%)
		& 11 (27.5\%)
		& 26 (65.0\%)
		\\
		
		\texttt{fsl\_anat}
		& 2 (5.0\%)
		& 26 (65.0\%)
		& 12 (30.0\%)
		\\
		
		\hline
	\end{tabular*}
	
	\par\smallskip
	\footnotesize
	Rank~1 denotes the preferred mask and rank~3 the least preferred mask.
	The three masks for each participant were presented using neutral,
	method-blinded identifiers. Values summarize the final participant-level
	ordering derived from the two neuroscientist assessments. Values summarize the final participant-level ordering obtained from method-blinded neuroscientist assessments. Two primary reviewers evaluated
	each participant independently; when their rankings disagreed, the independent
	assessment of a third reviewer was used for adjudication, with the final
	ordering determined by majority agreement.
	
\end{table}

% ============================================================
\subsection{Representative Inter-Method Qualitative Comparisons}
\label{subsec:qualitative_comparisons}
% ============================================================

Representative tri-planar native-FLAIR comparisons were examined
to provide anatomical context for the quantitative overlap and
geometric analyses. The selected examples illustrate distinct
patterns of inter-method agreement and disagreement, including
near-complete overlap, gross geometric misregistration,
orientation-dominant disagreement, and asymmetric
method-exclusive mask extent.

All examples were selected to illustrate specific spatial
patterns rather than to estimate their prevalence within
the cohort. Difference maps distinguish A\textsuperscript{3}PreCore-
exclusive voxels, comparator-exclusive voxels, and shared
mask regions. Because an independently delineated anatomical
reference was not available for these comparisons, the
direction or extent of disagreement was not interpreted
as evidence of anatomical correctness.

\paragraph{A\textsuperscript{3}PreCore versus \texttt{fsl\_anat}.}

\begin{figure}[h]
	\centering
	\includegraphics[
	width=0.6\linewidth
	]{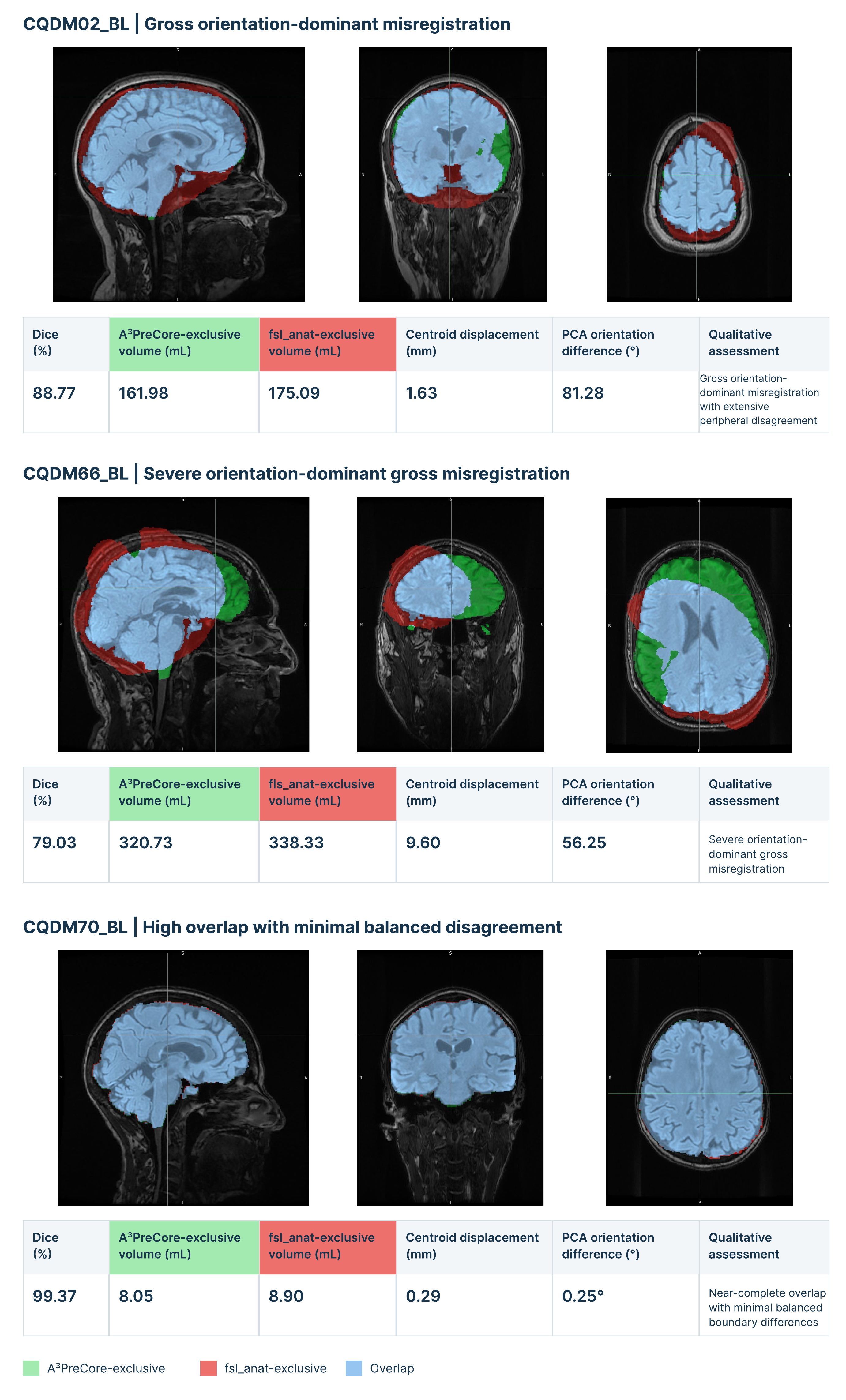}
	
	\caption{
		\textbf{Representative native-FLAIR comparisons between
			A\textsuperscript{3}PreCore and \texttt{fsl\_anat}.}
		Sagittal, coronal, and axial views illustrate three
		distinct patterns of inter-method disagreement.
		The top row (CQDM02\_BL) demonstrates pronounced
		orientation disagreement despite limited centroid
		displacement. The middle row (CQDM66\_BL) shows
		severe gross geometric disagreement involving
		substantial positional and orientational differences.
		The bottom row (CQDM70\_BL) illustrates near-complete
		overlap with minimal residual boundary differences.
		Green indicates A\textsuperscript{3}PreCore-exclusive
		regions, red indicates \texttt{fsl\_anat}-exclusive
		regions, and light blue indicates shared mask extent.
		Quantitative measurements are provided beneath each
		representative case. The difference maps characterize
		inter-method disagreement rather than anatomical
		ground-truth error.
	}
	\label{fig:qualitative_fslanat}
\end{figure}

Three representative comparisons demonstrated substantial
variation in the spatial relationship between
A\textsuperscript{3}PreCore and \texttt{fsl\_anat}
(Figure~\ref{fig:qualitative_fslanat}).

CQDM02\_BL illustrated pronounced orientation-dominant
disagreement. Despite a centroid displacement of only
1.63~mm, the PCA-derived orientation difference reached
$81.28^{\circ}$. The Dice coefficient was 0.8877, with
161.98~mL of A\textsuperscript{3}PreCore-exclusive volume
and 175.09~mL of \texttt{fsl\_anat}-exclusive volume.
The corresponding difference maps revealed extensive,
spatially coherent disagreement, demonstrating that a
relatively small centroid displacement can coexist with
substantial differences in mask orientation and extent.

A different pattern was observed in CQDM66\_BL, which
represented a severe gross-misregistration example.
The Dice coefficient decreased to 0.7903, with
320.73~mL of A\textsuperscript{3}PreCore-exclusive
volume and 338.33~mL of \texttt{fsl\_anat}-exclusive
volume. Unlike CQDM02\_BL, this case exhibited both
substantial centroid displacement (9.60~mm) and
pronounced PCA-derived orientation disagreement
($56.25^{\circ}$). Large, complementary exclusive
regions were apparent across the sagittal, coronal,
and axial views, indicating extensive geometric
disagreement rather than localized boundary variation.

Conversely, CQDM70\_BL demonstrated near-complete
spatial agreement. Its Dice coefficient reached
0.9937, with only 8.05~mL and 8.90~mL of
A\textsuperscript{3}PreCore- and
\texttt{fsl\_anat}-exclusive volume, respectively.
Centroid displacement was 0.29~mm, and the PCA-derived
orientation difference was $0.25^{\circ}$.
The remaining differences were small and predominantly
confined to localized mask boundaries.

Collectively, these examples illustrate three distinct
geometric patterns: pronounced orientation disagreement
with limited centroid displacement, combined positional
and orientational disagreement, and near-complete
inter-method correspondence. Their contrasting
characteristics complement the participant-level
geometric distributions presented in
Section~\ref{subsec:geometric_failures}.

\paragraph{A\textsuperscript{3}PreCore versus BET.}

\begin{figure}[h]
	\centering
	\includegraphics[
	width=0.6\linewidth
	]{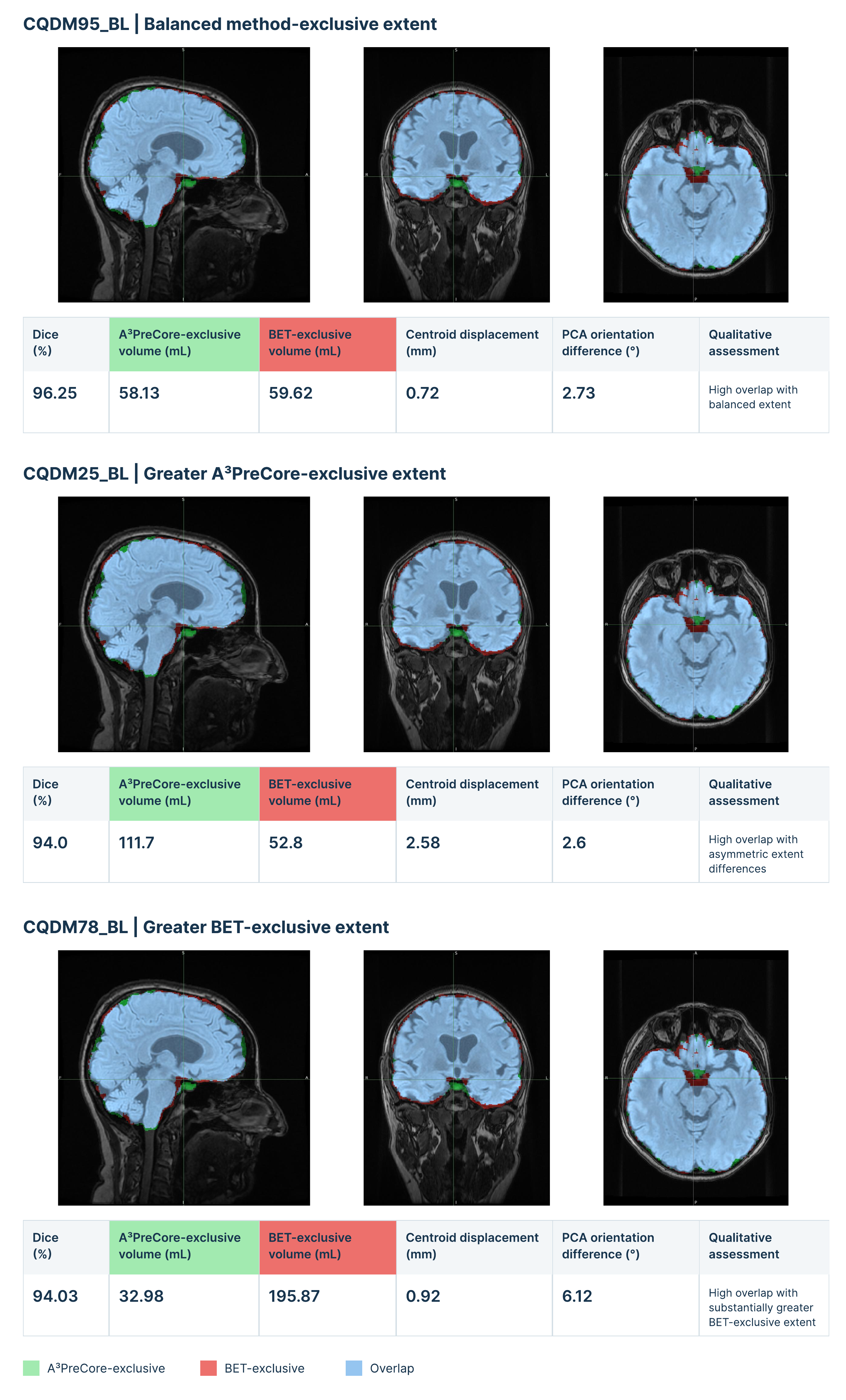}
	
	\caption{
		\textbf{Representative native-FLAIR comparisons between
			A\textsuperscript{3}PreCore and BET.}
		Sagittal, coronal, and axial views demonstrate
		distinct directional patterns of inter-method
		disagreement. The top row (CQDM95\_BL) shows
		extensive overlap with nearly balanced
		method-exclusive volumes. The middle row
		(CQDM25\_BL) illustrates greater
		A\textsuperscript{3}PreCore-exclusive extent,
		whereas the bottom row (CQDM78\_BL) demonstrates
		substantially greater BET-exclusive extent
		despite a comparable Dice coefficient.
		Green indicates A\textsuperscript{3}PreCore-exclusive
		regions, red indicates BET-exclusive regions,
		and light blue indicates shared mask extent.
		Quantitative measurements are reported beneath
		each case. The overlays describe relative
		mask extent and spatial disagreement without
		establishing anatomical superiority of either method.
	}
	\label{fig:qualitative_bet}
\end{figure}

The representative A\textsuperscript{3}PreCore--BET
comparisons demonstrated that similarly high global
overlap can conceal markedly different directional
patterns of mask disagreement
(Figure~\ref{fig:qualitative_bet}).

CQDM95\_BL exhibited extensive overlap and nearly
balanced method-exclusive volumes. Its Dice coefficient
was 0.9625, with 58.13~mL of
A\textsuperscript{3}PreCore-exclusive volume and
59.62~mL of BET-exclusive volume. Centroid displacement
and PCA-derived orientation difference were 0.72~mm
and $2.73^{\circ}$, respectively. The difference maps
showed relatively limited residual disagreement,
predominantly distributed along the mask boundaries.

CQDM25\_BL and CQDM78\_BL provided contrasting
examples of directional disagreement despite having
almost identical Dice coefficients.

For CQDM25\_BL, the Dice coefficient was 0.9400,
with 111.70~mL of A\textsuperscript{3}PreCore-exclusive
volume and 52.80~mL of BET-exclusive volume.
The corresponding centroid displacement was 2.58~mm,
and the PCA-derived orientation difference was
$2.60^{\circ}$. The difference maps showed a
greater overall A\textsuperscript{3}PreCore-exclusive
extent.

In contrast, CQDM78\_BL had a Dice coefficient of
0.9403 but demonstrated the opposite directional
pattern. BET-exclusive volume reached 195.87~mL,
whereas A\textsuperscript{3}PreCore-exclusive volume
was limited to 32.98~mL. Centroid displacement
was 0.92~mm, and the PCA-derived orientation
difference was $6.12^{\circ}$. The displayed overlays
revealed a substantially greater BET-exclusive
extent, including visible differences along
peripheral and inferior mask boundaries.

These examples demonstrate that global overlap
alone cannot characterize the direction or spatial
distribution of mask disagreement. In particular,
CQDM25\_BL and CQDM78\_BL exhibited comparable
Dice coefficients despite pronounced differences
in their respective method-exclusive volumes.
Directional volumetric measurements and spatial
difference maps therefore provide complementary
information that is not captured by a single
whole-mask overlap measure.

Taken together, the qualitative comparisons illustrate
the complementary information provided by overlap,
directional exclusive-volume measurements, and
pairwise geometric descriptors. The
A\textsuperscript{3}PreCore--\texttt{fsl\_anat}
examples demonstrate how near-complete agreement
and severe geometric disagreement can occur within
the same methodological comparison, including cases
in which substantial orientation disagreement is
not accompanied by proportionately large centroid
displacement.

The BET examples further demonstrate that comparable
Dice coefficients may correspond to substantially
different directional and spatial patterns of mask
disagreement. These observations reinforce the
importance of evaluating spatially resolved
inter-method differences alongside global overlap
and geometric measurements, without treating
either automatic method as an independent
anatomical reference.

Collectively, the results demonstrate that the improved registration
robustness of A\textsuperscript{3}PreCore was achieved without systematic
alteration of mask geometry in successfully registered cases.
The paired gross-misregistration analysis, spatial and boundary
measurements, multidimensional E-QC characterization, qualitative
comparisons, and blinded expert assessments provide complementary
evidence of this behaviour. A\textsuperscript{3}PreCore maintained
close spatial agreement with successfully registered
\texttt{fsl\_anat} masks while avoiding the severe geometric
discrepancies observed in confirmed registration-failure cases.
Comparisons with BET further revealed differences in peripheral
and inferior mask extent, highlighting the complementary information
provided by spatially resolved evaluation. Together, these findings
support the robustness of A\textsuperscript{3}PreCore for
T$_1$-to-FLAIR brain-mask transfer, while distinguishing
inter-method agreement from independently established
anatomical accuracy.

%% file: sections/Sec7.tex
% ============================================================
\section{Discussion}
\label{sec:discussion}
% ============================================================

% ============================================================
\subsection{Principal Findings}
\label{subsec:discussion_principal}
% ============================================================

This study evaluated A\textsuperscript{3}PreCore as a full-T$_1$-guided
strategy for transferring an established T$_1$-space anatomical brain mask
into native FLAIR space. The principal finding was that successful workflow
execution did not necessarily ensure plausible geometric placement. All
attempted BET, \texttt{fsl\_anat}, and A\textsuperscript{3}PreCore workflows
produced output masks. However, gross rigid-body misregistration was identified
in \(21/78\) \texttt{fsl\_anat} outputs and in none of the corresponding
A\textsuperscript{3}PreCore outputs. Within the present cohort,
A\textsuperscript{3}PreCore was therefore associated with substantially fewer
gross placement failures than the implemented \texttt{fsl\_anat}.

The controlled registration analysis supported this finding while showing that
the difference was concentrated in a specific subset of participants. When the
transformations from both registration strategies were reapplied to the same
full bias-corrected T$_1$-weighted image, the cohort-level NMI distributions
were closely matched. Nevertheless, every participant with
neuroscientist-confirmed gross \texttt{fsl\_anat} misregistration had higher NMI after
application of the A\textsuperscript{3}PreCore transformation. In participants without evident gross \texttt{fsl\_anat} displacement,
the two strategies produced similar NMI values. A\textsuperscript{3}PreCore
therefore appeared to prevent a methodologically important failure mode
rather than produce a uniform improvement in registrations that were already
successful.

The spatial-agreement analyses provided complementary evidence.
A\textsuperscript{3}PreCore remained highly concordant with successful
\texttt{fsl\_anat} and with masks generated directly from FLAIR using BET.
The lower-agreement tail in the \texttt{fsl\_anat} comparison was concentrated among
the grossly displaced outputs. Surface, regional, and slice-wise analyses
also showed that similar whole-mask volumes could coexist with substantial
differences in boundary position and orientation. The method-specific E-QC
descriptors further demonstrated that agreement in global mask extent or
FOV-relative position did not necessarily imply agreement in intensity,
boundary-gradient, peripheral, or inferior-tail characteristics, whereas
the pairwise geometric coordinates identified a smaller subset of pronounced
participant-specific discrepancies. These observations indicate that file
generation, native-grid compatibility, total mask volume, any individual
E-QC descriptor, and global overlap are insufficient on their own.

The findings should not be interpreted as estimates of absolute anatomical
accuracy. Independently delineated native-FLAIR brain masks were unavailable,
and the \texttt{fsl\_anat} failure classification incorporated quantitative disagreement
measures and tri-planar visual assessment. The evidence therefore supports a
more specific conclusion: in the present dataset,
A\textsuperscript{3}PreCore produced fewer gross geometric-placement failures
than the \texttt{fsl\_anat} while remaining broadly concordant with
both direct BET masks and successful \texttt{fsl\_anat} transfers.

% ============================================================
\subsection{\texorpdfstring{A\textsuperscript{3}PreCore}{A3PreCore} as a Full-T1-Guided Geometric-Transfer Strategy}
\label{subsec:discussion_geometric_strategy}
% ============================================================

A\textsuperscript{3}PreCore is an optimization-based geometric mask-transfer
method rather than a new brain-extraction or tissue-segmentation algorithm.
It does not redefine which T$_1$-weighted voxels constitute brain tissue.
Instead, it searches over a six-degree-of-freedom rigid-transformation space
comprising three translational and three rotational parameters to identify
the location and orientation that provide the best normalized-mutual-information
score under the configured search and optimization procedure. The resulting
subject-specific transformation is then applied to the existing T$_1$-space
anatomical mask. This distinction is important because transferred-mask quality
depends on two related but separable operations: definition of the source mask
and optimization of its placement in the target image. A plausible source mask
may produce an unusable native-FLAIR output when the estimated transformation
is incorrect, whereas an accurately estimated transformation cannot repair a
systematically inaccurate source-mask boundary.

Both registration branches propagated the same T$_1$-derived anatomical mask,
allowing placement differences to be assessed without introducing differences
in source-mask definition. The close agreement between
A\textsuperscript{3}PreCore and \texttt{fsl\_anat} in non-failure cases suggests that the
proposed workflow generally produced a closely concordant native-FLAIR mask
when \texttt{fsl\_anat} registration was successful. Its main advantage was observed in
the subset in which the \texttt{fsl\_anat} branch produced gross displacement. Any
systematic inclusion or exclusion already present in the T$_1$-space source
mask will, however, also be inherited by the A\textsuperscript{3}PreCore
output. The method should therefore be viewed as a complement to established
anatomical-processing pipelines rather than as a replacement for them.

One possible explanation for the difference between the registration
strategies is the spatial support presented to the multimodal similarity
objective. In the \texttt{fsl\_anat}, the moving image contains predominantly
skull-stripped brain tissue, whereas the fixed native FLAIR image may retain
skull, scalp, neck, head contour, inferior extracranial signal, and other
structures. The moving and fixed images therefore differ not only in tissue
contrast but also in the anatomical regions represented during transformation
estimation. This support mismatch may reduce the information available for
estimating global head position, particularly when the initial pose is unusual,
inferior coverage is extensive, or extracranial signal competes with
intracranial image structure.

The full bias-corrected T$_1$ image preserves global structures that are
spatially related to features visible in the raw FLAIR acquisition. Head
contour, skull and scalp boundaries, and inferior anatomy may provide
additional constraints on translation and rotation despite differences in
cross-modal contrast. Together with an entropy-based multimodal objective and
a broad angular search, this additional support may reduce convergence to
geometrically implausible local solutions
\cite{wells1996multi,maes1996multi,jenkinson2001global,jenkinson2002improved}.
The predominance of translation-dominant and combined failures in the
\texttt{fsl\_anat} branch is consistent with this interpretation.

The present comparison did not isolate the contribution of the full-T$_1$
representation alone. The workflows differed in three registration
components: the moving-image representation, similarity criterion, and
angular-search specification. \texttt{fsl\_anat} used a skull-stripped
T$_1$-weighted image, mutual information, and default angular-search settings,
whereas A\textsuperscript{3}PreCore used the full bias-corrected T$_1$ image,
normalized mutual information, and an explicit \(\pm90^{\circ}\) search around
each axis. The observed difference should therefore be attributed to the
complete A\textsuperscript{3}PreCore strategy as implemented rather than
exclusively to the presence of extracranial information.

The controlled NMI findings provide optimization-consistent support for this
interpretation but do not constitute independent anatomical validation. NMI
favoured A\textsuperscript{3}PreCore in all confirmed \texttt{fsl\_anat}
gross-misregistration cases and remained closely matched when \texttt{fsl\_anat}
registration was successful. However, a global intensity-based similarity
measure may be insensitive to small but relevant boundary differences, and the
participant-level NMI direction depended partly on the voxel support used for
evaluation. NMI should therefore be interpreted alongside geometric measures
and visual review rather than as a stand-alone measure of anatomical validity.

A\textsuperscript{3}PreCore is also restricted to rigid-body placement. This
choice is appropriate for correcting dominant differences in head pose between
same-participant T$_1$-weighted and FLAIR acquisitions while avoiding scaling,
shearing, or nonlinear deformation of the anatomical mask. It does not imply
that acquisition-related distortion or sequence-dependent geometric
differences are absent. Applications requiring correction beyond rigid pose
may benefit from affine, nonlinear, distortion-aware, or acquisition-specific
registration approaches.

% ============================================================
\subsection{Multidimensional E-QC and Interpretation Without Manual Ground Truth}
\label{subsec:discussion_eqc_interpretation}
% ============================================================

The E-QC findings demonstrate why brain-mask assessment should not be
reduced to a single global measure. The six method-specific descriptors
($V$, $C$, $R_I$, $B_G$, $P_{\mathrm{peripheral}}$, and
$P_{\mathrm{tail}}$) characterize complementary aspects of mask geometry,
image-derived support, and peripheral or inferior extent. A rigidly
displaced mask may retain a plausible total volume or FOV-relative centroid
position despite exhibiting geometrically implausible placement.
Conversely, two masks with high global overlap may differ in comparatively
small but spatially relevant boundary regions. Descriptor-level
characterization and global overlap therefore remain insufficient when
interpreted individually and should be considered alongside explicit
pairwise geometric measures.

The pairwise geometric coordinates provided information that was absent from
the method-specific descriptors alone. Centroid displacement characterized
differences in global mask position, whereas PCA-derived orientation
difference identified cases in which mask orientation changed without a
comparably large centroid shift. The corresponding
orientation-induced boundary-displacement measure expressed orientation
disagreement in physical-distance units. The presence of an
orientation-dominant case with modest centroid displacement showed that
centroid-based assessment alone would not identify every gross geometric
abnormality.

Boundary, regional, and slice-wise analyses further separated global
agreement from local conformity. Large surface displacement could occur with
little change in total mask volume, whereas comparatively small peripheral
differences could be obscured by high global Dice values. The inferior region
was the least stable in both prespecified comparisons and was particularly
informative in the A\textsuperscript{3}PreCore--BET analysis. Differences in
mask extent near cervical, spinal, or other extracranial structures may
contribute relatively few voxels to the complete mask while potentially
affecting measurements near the cerebellum, brainstem, and inferior temporal
regions.

The main value of E-QC in the present study was therefore not the production
of a single pass--fail score. Instead, the method-specific descriptors and
pairwise geometric coordinates distinguished different forms of disagreement,
identified participant-specific geometric abnormalities, revealed differences
concealed by volume and global overlap, and prioritized masks for targeted
tri-planar inspection. This multidimensional interpretation is particularly
useful when a method preserves broadly similar intracranial coverage while
differing in a comparatively small boundary region.

Automatic agreement must nevertheless be interpreted cautiously when no
independent anatomical reference is available. Expert-derived masks generated
according to a prespecified protocol provide the most direct available basis
for estimating anatomical segmentation accuracy, although such references
remain subject to annotation decisions and inter-rater variability. Because
independently delineated native-FLAIR masks were unavailable, Dice, Jaccard,
surface distance, directional exclusive volume, regional agreement, and
pairwise geometric descriptors quantified relationships between automatic
outputs rather than absolute correctness.

High agreement between two automatic methods is reassuring, particularly when
the methods use different sources of information, but it does not prove that
either output is anatomically correct. Both methods may share a systematic
boundary error. Conversely, low agreement confirms that the methods differ but
does not determine which boundary is preferable. This distinction is
particularly relevant to the A\textsuperscript{3}PreCore--BET comparison
because one method transfers a T$_1$-defined anatomical mask, whereas the
other estimates the brain boundary directly from FLAIR. Their disagreements
may reflect differences in modality, mask definition, dilation, local boundary
behaviour, or geometric placement.

E-QC consequently occupies an intermediate position between automatic
agreement and independent anatomical annotation. Its method-specific
descriptors quantify complementary mask and mask-to-image characteristics,
whereas its pairwise geometric coordinates describe relative position,
orientation, and boundary-level disagreement. These measures can flag
suspicious outputs, reveal participant-specific geometric abnormalities,
and support interpretation of systematic boundary differences, but they
cannot replace voxel-level reference masks. Similarly, the
three-neuroscientist review provided important confirmation of gross
displacement but did not constitute independent manual delineation.
Claims of absolute anatomical superiority will therefore require
reference-based validation.

% ============================================================
\subsection{Relationship to Existing Methods and Pipeline Implications}
\label{subsec:discussion_existing_implications}
% ============================================================

Established brain-extraction approaches include intensity- and
morphology-based methods, atlas and label-fusion techniques, hybrid anatomical
models, and learning-based systems
\cite{smith2002fast,Shattuck2002,segonne2004hybrid,iglesias2011robust,
eskildsen2012beast,kleesiek2016deep,isensee2019automated,hoopes2022synthstrip}.
Direct FLAIR-space extraction has the practical advantage of requiring only
the target image and avoiding cross-modal registration failure. In the present
cohort, BET generated technically complete masks for all available FLAIR
images and showed high overall agreement with A\textsuperscript{3}PreCore.

The boundary-sensitive and regional analyses nevertheless identified
systematic differences between BET and A\textsuperscript{3}PreCore,
particularly near peripheral and inferior mask regions. In representative
cases, greater BET-exclusive extent was concentrated near inferior boundaries
adjacent to the skull base, cervical region, and other extracranial
structures. High global overlap therefore reflected broadly similar
intracranial coverage but did not capture every local boundary difference
between the two methods. Because no manual native-FLAIR reference was
available, these findings describe relative boundary behaviour rather than
definitive anatomical superiority.

T$_1$-guided transfer provides a complementary advantage by reusing masks
derived from a structural modality with strong anatomical contrast and mature
processing tools. Its principal vulnerability lies in the transformation
between modalities. A\textsuperscript{3}PreCore addresses this stage without
replacing the source-mask algorithm or requiring a training dataset. It is
therefore compatible in principle with masks generated by FSL,
FreeSurfer-related workflows, atlas-based methods, or other validated
T$_1$-processing pipelines, provided that the source mask and full
bias-corrected T$_1$ image share a consistent coordinate system.

Learning-based extraction methods may capture complex local boundary features
that are unavailable to a rigid-transfer procedure. Their performance may,
however, depend on the training distribution, acquisition characteristics,
pathology representation, and deployment domain.
A\textsuperscript{3}PreCore is comparatively simple, training-free, and
geometrically interpretable. These properties may facilitate integration into
existing pipelines and make failure modes easier to inspect, although they do
not establish superiority over modern learning-based systems. The present BET
comparison should also not be interpreted as a comprehensive benchmark against
all contemporary direct FLAIR or modality-agnostic extraction approaches.

More flexible registration methods, including boundary-based, symmetric,
affine, and nonlinear strategies, may further improve cross-modal alignment
\cite{wells1996multi,maes1996multi,jenkinson2001global,jenkinson2002improved,
avants2008symmetric,greve2009accurate}.
The present method deliberately used rigid transformation to preserve source
anatomy while correcting global differences in pose. Whether greater
transformation flexibility provides additional benefit without introducing
unnecessary deformation, instability, or sensitivity to FLAIR pathology
remains to be established.

A broader implication is that preprocessing validation should be divided into
explicit stages. Input availability, workflow completion, output readability,
native-grid compatibility, registration similarity, descriptor-level mask
plausibility, geometric placement, and local boundary conformity should not
be collapsed into a single success indicator. A workflow may generate an
apparently valid output while retaining a gross transformation error, and a
mask may have high Dice agreement or a similar volume while containing
systematic regional boundary differences.

The findings support further evaluation of A\textsuperscript{3}PreCore as a
registration-based branch when paired T$_1$-weighted and FLAIR images are
available. Direct FLAIR extraction remains useful when a suitable T$_1$ image
or anatomical source mask is unavailable. Automated E-QC descriptors and
pairwise geometric coordinates could prioritize uncertain cases for
tri-planar review, whereas retaining the transformation matrix and registered
full-T$_1$ image would provide auditable intermediate outputs. These
safeguards would allow global placement and local boundary behaviour to be
assessed independently of file completion.

Reliable mask placement is potentially relevant to FLAIR pipelines in which
brain masks are used before lesion segmentation, regional burden estimation,
radiomic analysis, or image normalization. Gross displacement may exclude
valid brain tissue or include background structures, whereas subtler inferior
and peripheral differences may affect regional measurements despite having
limited influence on global volume. The current findings establish an
important preprocessing distinction between technical completion and
geometric plausibility. Whether these differences improve downstream lesion
or clinical measurements remains to be determined.

% ============================================================
\subsection{Limitations and Future Directions}
\label{subsec:discussion_limitations_future}
% ============================================================

Several limitations define the interpretation and generalizability of the
findings. First, independently delineated native-FLAIR brain masks were
unavailable. The study could therefore assess technical completion, gross
placement, automatic agreement, image-derived plausibility, relative geometry,
and visually assessed boundary behaviour, but not absolute voxel-level
anatomical accuracy. The three-neuroscientist review strengthened the
assessment of gross displacement but did not constitute independent manual
delineation. Future validation should include native-FLAIR reference masks
generated under a prespecified protocol by multiple blinded reviewers,
including both routine cases and an enriched sample of suspected failures.

Second, the comparison did not isolate the individual contributions of the
A\textsuperscript{3}PreCore strategy. Moving-image representation, similarity
criterion, and angular-search range differed simultaneously between \texttt{fsl\_anat} and A\textsuperscript{3}PreCore. It is therefore not possible to
determine whether the reduction in gross misregistration arose primarily from
the full-T$_1$ image, normalized mutual information, expanded angular search,
or interactions among these factors. Prespecified ablation experiments should
vary these components independently while holding the remaining registration
settings constant.

Third, the study was conducted in a single-site cohort of older
community-dwelling adults. Scanner characteristics, acquisition protocols,
participant age, pathology burden, head pose, and field-of-view coverage may
affect both registration and direct FLAIR extraction. The observed \texttt{fsl\_anat}
failure rate and BET disagreement pattern should therefore not be generalized
directly to other datasets. External validation should include multiple sites,
scanners, protocols, age groups, and clinically relevant populations.

Fourth, A\textsuperscript{3}PreCore is limited to rigid-body transformation.
Rigid registration preserves the geometry of the source mask. It is suitable
for correcting global pose, but it cannot address scale differences, shear,
nonlinear anatomical variation, or sequence-dependent distortion. Future
studies should compare the rigid strategy with affine, boundary-based,
distortion-aware, and nonlinear alternatives while assessing the risk of
introducing unnecessary or biologically inappropriate deformation.

Fifth, E-QC remains exploratory. Its descriptor definitions and parameter
choices are heuristic, several measures may be correlated, and the
PCA-derived coordinates describe relative mask geometry rather than the
underlying transformation parameters. Although the present findings show
that multidimensional characterization can identify discrepancies missed by
global measures, no validated decision thresholds for automated acceptance
or rejection were established. Calibration and validation should therefore
be performed using independent datasets and blinded reference assessments.

Sixth, interpretation of local boundary differences relied on the convergence
of boundary-sensitive measures and visual review rather than manual
voxel-level ground truth. The findings support systematic differences between
the methods, particularly in peripheral and inferior regions, but do not
establish that every A\textsuperscript{3}PreCore boundary represents the
anatomically optimal contour. Reference-based analyses are required to
determine whether these differences correspond to lower surface error and
improved regional accuracy.

Finally, the study evaluated brain-mask generation and placement rather than
white-matter-hyperintensity or lesion-segmentation performance. Improved mask
geometry does not necessarily lead to improved lesion sensitivity, regional
burden estimation, or clinical association. A downstream study should apply
the same lesion-segmentation algorithm after BET, \texttt{fsl\_anat}, and
A\textsuperscript{3}PreCore masking while holding all other settings constant.
Evaluation should include voxelwise and lesion-wise agreement with manual
lesion annotations, regional lesion burden, performance near peripheral and
inferior boundaries, and sensitivity to mask dilation.

A longer-term goal is to combine full-T$_1$ anatomical guidance, direct FLAIR
information, automated failure detection, and lesion segmentation within a
unified but modular pipeline. Maintaining separately auditable processing
stages would help determine whether downstream errors originate from
source-mask definition, cross-modal placement, local FLAIR boundary
estimation, QC classification, or lesion segmentation.

The potential relevance of this framework extends to studies in populations
for whom FLAIR-based lesion assessment is particularly important, including
older adults and individuals with neurological disorders such as cerebral
small-vessel disease, multiple sclerosis, and neurodegenerative conditions.
In these settings, variability in anatomy, lesion burden, acquisition
coverage, and image quality may increase the importance of reliable
participant-specific brain-mask placement. A robust and explicitly
quality-controlled T$_1$-to-FLAIR transfer strategy may therefore provide a
more stable preprocessing foundation for subsequent lesion quantification and
regional analyses. Future studies should determine whether these geometric
advantages generalize across clinical populations, scanners, acquisition
protocols, and downstream lesion-segmentation tasks.

%% file: sections/Sec8.tex
\section{Conclusion}
\label{sec:conclusion}

A\textsuperscript{3}PreCore provides a practical, training-free, optimization-based strategy for propagating an established T$_1$-derived anatomical brain mask into native FLAIR space. The method estimates a subject-specific six-degree-of-freedom rigid transformation between the fully bias-corrected T$_1$-weighted image and native FLAIR using normalized mutual information, jointly optimizing translation and rotation, and then applies the resulting transformation to the T$_1$-space anatomical mask.

In the present cohort, gross rigid-body placement failures observed
with the implemented \texttt{fsl\_anat} were not reproduced by
A\textsuperscript{3}PreCore. At the same time, the proposed method
remained closely concordant with successful \texttt{fsl\_anat} registrations and
showed high overall agreement with masks generated directly from FLAIR
using BET. These findings demonstrate that technical workflow completion,
native-grid compatibility, and plausible global mask characteristics
alone do not ensure reliable cross-modal mask placement. Assessment of
both translational and rotational geometry is therefore important for
identifying participant-specific registration failures.

The results also support a multidimensional approach to mask
quality-control assessment. Method-specific E-QC descriptors, together
with overlap, boundary, regional, and pairwise geometric measures,
revealed patterns of disagreement that were not apparent from global
mask volume or overlap alone. In particular, centroid displacement,
PCA-derived orientation difference, and orientation-induced boundary
displacement provided complementary information about inter-method
geometry and may help prioritize suspicious outputs for targeted
tri-planar review.

A\textsuperscript{3}PreCore should therefore be interpreted as an optimized geometric mask-transfer strategy rather than as a replacement for anatomical brain extraction. Its observed performance reflects the use of full-T$_1$ spatial support together with normalized-mutual-information-based six-degree-of-freedom rigid optimization of subject-specific translation and rotation. Independent native-FLAIR reference masks and multi-site validation will be required to establish absolute anatomical accuracy and generalizability. Within these limitations, the present findings support A\textsuperscript{3}PreCore as a robust, auditable approach to reducing severe cross-modal brain-mask placement failures in native FLAIR preprocessing.

%% file: sections/sec9.tex
% ============================================================
% Supplementary Material
% ============================================================

\section*{Supplementary Material}

% Supplementary numbering
\renewcommand{\thesection}{S\arabic{section}}
\renewcommand{\thesubsection}{\thesection.\arabic{subsection}}
\renewcommand{\thesubsubsection}{\thesubsection.\arabic{subsubsection}}

\renewcommand{\thefigure}{S\arabic{figure}}
\renewcommand{\thetable}{S\arabic{table}}

\setcounter{section}{0}
\setcounter{figure}{0}
\setcounter{table}{0}

% ============================================================
\section{Supplementary Results}
\label{sec:supp_results}
% ============================================================

% ============================================================
\subsection{Sensitivity Analysis of Controlled NMI}
\label{subsec:supp_nmi_sensitivity}
% ============================================================

\begin{figure}
	\centering
	\includegraphics[
	width=0.8\linewidth,
	keepaspectratio
	]{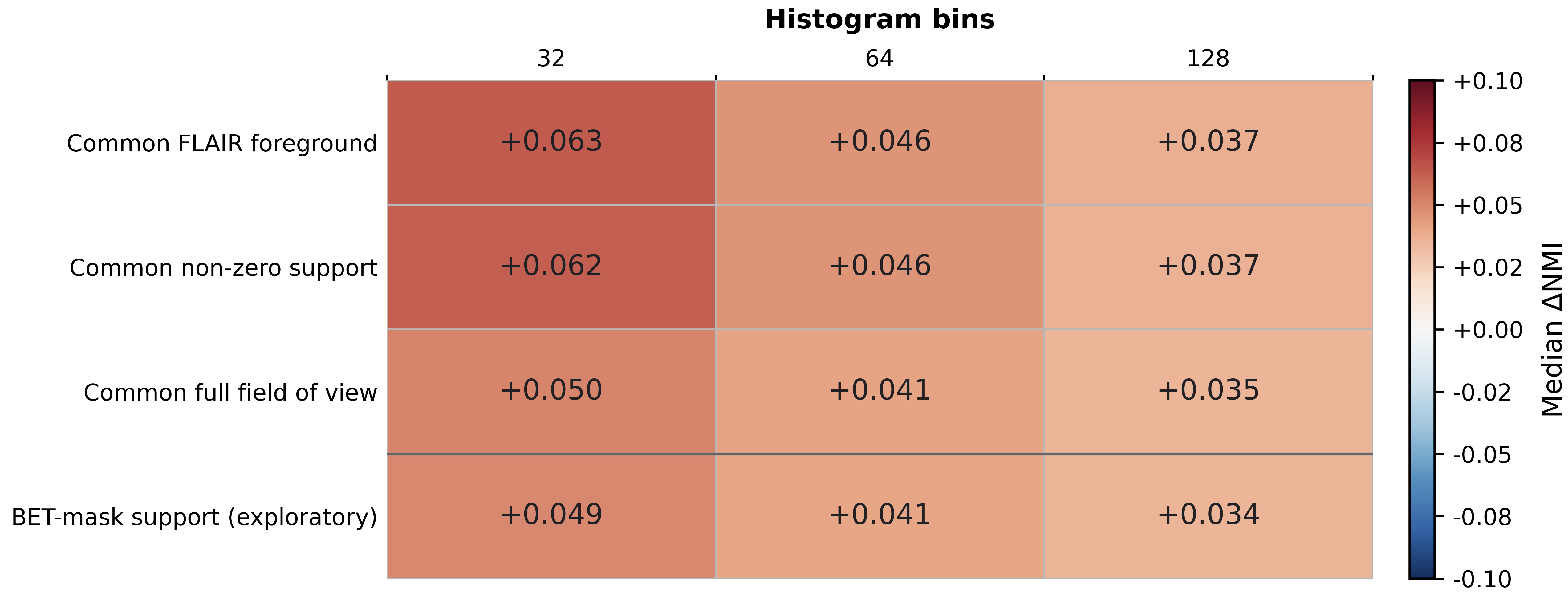}
	
	\caption{
		\textbf{Sensitivity of the paired controlled-NMI comparison to voxel
			support and histogram discretization.}
		Median participant-level differences,
		$\Delta\mathrm{NMI}
		=
		\mathrm{NMI}_{\mathrm{A3P}}
		-
		\mathrm{NMI}_{\mathrm{fsl\_anat}}$,
		are shown for common FLAIR foreground, common non-zero support,
		common full-field-of-view support, and BET-mask support using
		32, 64, and 128 histogram bins. Positive values indicate higher
		NMI for A\textsuperscript{3}PreCore. BET-mask support was evaluated
		as an exploratory support-restriction analysis.
	}
	\label{fig:nmi_sensitivity}
\end{figure}

To evaluate the dependence of the controlled NMI comparison on voxel
support and histogram discretization, the paired analysis was repeated
across alternative support definitions and histogram-bin settings.
The configured main analysis used common FLAIR-foreground support with
64 histogram bins. Sensitivity analyses were performed using 32, 64,
and 128 bins under four support definitions: common FLAIR foreground,
common non-zero support, common full field of view, and BET-mask support
(Table~\ref{tab:nmi_sensitivity};
Figure~\ref{fig:nmi_sensitivity}).

Across the three common-support definitions, the median paired
A\textsuperscript{3}PreCore-minus-\texttt{fsl\_anat} NMI difference
remained slightly positive at all histogram resolutions, although its
absolute magnitude was close to zero. Under the configured main analysis,
the median $\Delta\mathrm{NMI}$ was 0.000063
(IQR, $-0.000171$ to 0.032549), with higher NMI for
A\textsuperscript{3}PreCore in 44/78 participants and for
\texttt{fsl\_anat} in 34/78. Within common FLAIR-foreground support,
the median paired difference was 0.000144 at 32 bins and 0.000047 at
128 bins. Across the remaining common-support definitions, median
differences ranged from 0.000065 to 0.000256 for common non-zero support
and from 0.000098 to 0.000147 for the common full field of view.

Mean $\Delta\mathrm{NMI}$ values were consistently positive and
substantially larger than the corresponding medians across the
common-support configurations. For example, the configured main analysis
yielded a mean paired difference of 0.015939 but a median difference of
only 0.000063. This separation between mean and median is consistent with
a subset of participants exhibiting comparatively large positive NMI
differences rather than a uniform cohort-wide shift of similar magnitude.

The participant-level direction of the paired NMI difference was more
sensitive to evaluation support than to histogram resolution. When
evaluation was restricted to BET-mask support, the median paired
difference became slightly negative at all three histogram resolutions:
$-0.001203$, $-0.000981$, and $-0.000827$ for 32, 64, and 128 bins,
respectively. Under each BET-mask-support configuration,
\texttt{fsl\_anat} had the higher NMI in 50/78 participants and
A\textsuperscript{3}PreCore in 28/78, although the mean paired difference
remained positive. BET-mask support was therefore retained as an
exploratory support-restriction analysis rather than interpreted as a
replacement for the configured common-FLAIR-foreground analysis.

Importantly, all 21 participants classified as having gross
\texttt{fsl\_anat} misregistration retained positive paired NMI
differences across the evaluated common-support configurations.
By contrast, median differences among participants without gross
\texttt{fsl\_anat} misregistration remained close to zero. This pattern
is consistent with the principal controlled-registration analysis,
indicating that the largest NMI differences were concentrated in the
gross-misregistration subgroup rather than distributed uniformly across
the cohort.

The influence of individual participants on the configured main-analysis
median was additionally evaluated using leave-one-participant-out
analyses. The full-sample median $\Delta\mathrm{NMI}$ was 0.00006260.
Across the 78 leave-one-participant-out analyses, the median ranged from
0.00005219 to 0.00007301, with a maximum absolute deviation of
$1.04\times10^{-5}$ from the full-sample estimate. No single participant
therefore materially determined the cohort median. This analysis addresses
sensitivity to individual participants only and should not be interpreted
as evidence of invariance to voxel-support definition, histogram
resolution, intensity preprocessing, or NMI implementation.

\begin{table}[htbp]
	
	\caption{
		Sensitivity of the controlled NMI comparison to evaluation support and
		histogram resolution. Positive $\Delta\mathrm{NMI}$ values indicate higher
		NMI for A\textsuperscript{3}PreCore than for \texttt{fsl\_anat}.
		The configured main analysis is shown in bold.
	}
	\label{tab:nmi_sensitivity}
	
	\centering
	\scriptsize
	\setlength{\tabcolsep}{3.5pt}
	\renewcommand{\arraystretch}{1.15}
	
	\begin{tabular}{
			@{}
			>{\raggedright\arraybackslash}p{0.20\linewidth}
			>{\raggedright\arraybackslash}p{0.07\linewidth}
			>{\raggedright\arraybackslash}p{0.14\linewidth}
			>{\raggedright\arraybackslash}p{0.27\linewidth}
			>{\raggedright\arraybackslash}p{0.22\linewidth}
			@{}
		}
		\hline
		
		Evaluation support
		& Bins
		& Mean $\Delta$NMI
		& Median $\Delta$NMI (IQR)
		& A\textsuperscript{3}PreCore/\texttt{fsl\_anat} higher, $n$
		\\
		
		\hline
		
		Common FLAIR foreground
		& 32
		& 0.021211
		& 0.000144 ($-0.000105$--0.043475)
		& 46/32
		\\
		
		\textbf{Common FLAIR foreground}
		& \textbf{64}
		& \textbf{0.015939}
		& \textbf{0.000063 ($-0.000171$--0.032549)}
		& \textbf{44/34}
		\\
		
		Common FLAIR foreground
		& 128
		& 0.012836
		& 0.000047 ($-0.000127$--0.025678)
		& 46/32
		\\
		
		Common full field of view
		& 32
		& 0.015441
		& 0.000098 ($-0.000121$--0.028743)
		& 47/31
		\\
		
		Common full field of view
		& 64
		& 0.012883
		& 0.000128 ($-0.000075$--0.023143)
		& 52/26
		\\
		
		Common full field of view
		& 128
		& 0.011072
		& 0.000147 ($-0.000010$--0.019317)
		& 56/22
		\\
		
		Common non-zero support
		& 32
		& 0.021289
		& 0.000256 ($-0.000060$--0.043714)
		& 52/26
		\\
		
		Common non-zero support
		& 64
		& 0.015850
		& 0.000070 ($-0.000132$--0.032316)
		& 48/30
		\\
		
		Common non-zero support
		& 128
		& 0.012831
		& 0.000065 ($-0.000109$--0.025599)
		& 48/30
		\\
		
		\hline
		
		BET-mask support
		& 32
		& 0.014050
		& $-0.001203$ ($-0.002803$--0.041012)
		& 28/50
		\\
		
		BET-mask support
		& 64
		& 0.011578
		& $-0.000981$ ($-0.002302$--0.033901)
		& 28/50
		\\
		
		BET-mask support
		& 128
		& 0.009823
		& $-0.000827$ ($-0.001943$--0.028810)
		& 28/50
		\\
		
		\hline
	\end{tabular}
	
	\par\smallskip
	\footnotesize
	$\Delta\mathrm{NMI}$ was calculated participant-wise as
	$\mathrm{NMI}_{\mathrm{A3P}}-\mathrm{NMI}_{\mathrm{fsl\_anat}}$.
	All configurations summarize the same 78 paired participants.
	The configured main analysis used common FLAIR-foreground support and
	64 histogram bins. BET-mask support was evaluated as an exploratory
	support restriction. NMI is a registration-related multimodal-similarity
	measure and does not independently establish anatomical correctness.
	
\end{table}

% ============================================================
\subsection{Regional and Slice-Wise Spatial Agreement}
\label{subsec:supp_regional_agreement}
% ============================================================

Regional and slice-wise analyses were performed to characterize the
spatial distribution of inter-method disagreement beyond whole-mask
overlap. Regional Dice coefficients were calculated after dividing the
mask extent into inferior, middle, and superior thirds along the
inferior--superior anatomical axis. Slice-wise Dice coefficients were
summarized across all evaluable slices along the same axis.

Agreement varied systematically by anatomical region
(Table~\ref{tab:regional_slice_disagreement}). For both comparisons,
the inferior third showed lower Dice agreement than the middle and
superior thirds. This pattern was most pronounced for the
A\textsuperscript{3}PreCore--BET comparison, for which the median
inferior-third Dice was 0.870 compared with 0.962 in the middle third
and 0.942 in the superior third. The corresponding values for the
A\textsuperscript{3}PreCore--\texttt{fsl\_anat} comparison were
0.961, 0.991, and 0.989, respectively.

The slice-wise summaries showed a similar pattern. Median slice Dice
remained high for both comparisons, whereas the lower-tail
10th-percentile slice Dice was substantially lower, particularly for
A\textsuperscript{3}PreCore--BET. These findings indicate that high
overall or median slice agreement can coexist with more pronounced
disagreement in a restricted subset of slices, consistent with the
regional concentration of differences near inferior mask boundaries.

\begin{table}[htbp]
	
	\caption{
		Regional and slice-wise spatial agreement between
		A\textsuperscript{3}PreCore and the two comparators.
	}
	\label{tab:regional_slice_disagreement}
	
	\centering
	\small
	\renewcommand{\arraystretch}{1.15}
	\setlength{\tabcolsep}{6pt}
	
	\begin{tabular}{
			@{}
			>{\raggedright\arraybackslash}p{0.25\linewidth}
			>{\raggedright\arraybackslash}p{0.33\linewidth}
			>{\raggedright\arraybackslash}p{0.30\linewidth}
			@{}
		}
		\hline
		
		Metric
		& A\textsuperscript{3}PreCore--\texttt{fsl\_anat}
		& A\textsuperscript{3}PreCore--BET
		\\
		
		\hline
		
		Inferior-third Dice
		& 0.961 (0.838--0.980)
		& 0.870 (0.800--0.898)
		\\
		
		Middle-third Dice
		& 0.991 (0.939--0.994)
		& 0.962 (0.957--0.965)
		\\
		
		Superior-third Dice
		& 0.989 (0.935--0.993)
		& 0.942 (0.931--0.947)
		\\
		
		Mean slice Dice
		& 0.975 (0.869--0.983)
		& 0.883 (0.872--0.897)
		\\
		
		Median slice Dice
		& 0.990 (0.940--0.994)
		& 0.956 (0.952--0.959)
		\\
		
		10th-percentile slice Dice
		& 0.962 (0.795--0.977)
		& 0.776 (0.704--0.828)
		\\
		
		\hline
	\end{tabular}
	
	\par\smallskip
	\footnotesize
	Values are median (interquartile range) across participants.
	The mask extent was divided into inferior, middle, and superior thirds
	along the inferior--superior anatomical axis. Slice-wise values summarize
	Dice coefficients across all evaluable slices along the same axis.
	The 10th-percentile slice Dice was retained as the lower-tail summary
	because the minimum slice Dice was frequently zero in small terminal slices
	and was therefore less stable as a cohort-level descriptor.
	
\end{table}

% ============================================================
\subsection{Whole-Mask Volume Disagreement}
\label{subsec:supp_volume_disagreement}
% ============================================================

Whole-mask volume differences were examined to determine the extent to
which global mask size reflected spatial disagreement between methods.
Because substantial registration-related displacement or boundary
reconfiguration can occur with little change in total mask volume,
volume-based measures were interpreted jointly with overlap and
boundary-sensitive descriptors.

A\textsuperscript{3}PreCore and \texttt{fsl\_anat} showed minimal
whole-mask volume disagreement despite the presence of substantial
participant-specific spatial discrepancies. The median signed volume
difference was $-0.370$~mL, and the median absolute difference was
0.464~mL (0.030\%). No participant exceeded an absolute volume
difference of 2\%, despite 21/78 participants exceeding a normalized
symmetric difference of 0.10 and 21/78 exceeding an HD95 of 5~mm
(Table~\ref{tab:volume_disagreement}).

The A\textsuperscript{3}PreCore--BET comparison showed larger differences
in total mask volume. The median signed difference was $-24.013$~mL,
the median absolute difference was 30.277~mL, and the median absolute
percentage difference was 2.136\%. Forty-one of 78 participants exceeded
an absolute volume difference of 2\%. Nevertheless, marked spatial
disagreement was not reducible to volume difference alone: 9/78 participants
exceeded a normalized symmetric difference of 0.10, 12/78 exceeded an
HD95 of 5~mm, and 2/78 exceeded an HD95 of 10~mm.

These findings reinforce the distinction between global mask size and
spatial agreement. In particular, the
A\textsuperscript{3}PreCore--\texttt{fsl\_anat} comparison demonstrates
that large geometric or boundary discrepancies can occur despite
near-identical total mask volumes when disagreement primarily reflects
translation, orientation, or spatial redistribution rather than net
mask size.

\begin{table}[htbp]
	
	\caption{
		Whole-mask volume disagreement and frequencies of marked spatial
		disagreement between A\textsuperscript{3}PreCore and the two comparators.
	}
	\label{tab:volume_disagreement}
	
	\centering
	\small
	\renewcommand{\arraystretch}{1.15}
	\setlength{\tabcolsep}{6pt}
	
	\begin{tabular}{
			@{}
			>{\raggedright\arraybackslash}p{0.25\linewidth}
			>{\raggedright\arraybackslash}p{0.33\linewidth}
			>{\raggedright\arraybackslash}p{0.30\linewidth}
			@{}
		}
		\hline
		
		Metric
		& A\textsuperscript{3}PreCore--\texttt{fsl\_anat}
		& A\textsuperscript{3}PreCore--BET
		\\
		
		\hline
		
		Signed volume difference (mL)
		& $-0.370$ ($-0.647$--0.136)
		& $-24.013$ ($-50.048$--4.521)
		\\
		
		Absolute volume difference (mL)
		& 0.464 (0.217--0.839)
		& 30.277 (17.690--56.709)
		\\
		
		Absolute volume difference (\%)
		& 0.030 (0.015--0.061)
		& 2.136 (1.114--3.910)
		\\
		
		A\textsuperscript{3}PreCore volume smaller/larger, $n$
		& 53/25
		& 56/22
		\\
		
		Normalized symmetric difference $>0.10$, $n$
		& 21
		& 9
		\\
		
		HD95 $>5$ mm, $n$
		& 21
		& 12
		\\
		
		HD95 $>10$ mm, $n$
		& 11
		& 2
		\\
		
		Absolute volume difference $>2\%$, $n$
		& 0
		& 41
		\\
		
		\hline
	\end{tabular}
	
	\par\smallskip
	\footnotesize
	Continuous values are median (interquartile range); counts are numbers
	of participants out of 78. Signed volume difference was calculated as
	A\textsuperscript{3}PreCore minus comparator volume.
	HD95 denotes the 95th-percentile Hausdorff distance.
	Marked spatial or boundary disagreement can coexist with minimal
	whole-mask volume difference when disagreement primarily reflects
	orientation, translation, or boundary placement rather than net mask size.
	
\end{table}